\documentclass[twoside,11pt]{article}

\usepackage[shortlabels]{enumitem}
\usepackage{amsmath}
\usepackage{amsfonts}
\usepackage{Definitions}
\usepackage{color}
\usepackage[colorlinks,
            linkcolor=blue,
            anchorcolor=blue,
            citecolor=green
            ]{hyperref}
\usepackage{xcolor}
\usepackage{xspace}
\usepackage{graphicx}
\usepackage{subcaption}
\usepackage{booktabs}
\usepackage{multirow}
\usepackage{multicol}

\usepackage{microtype}

\usepackage{wrapfig}
\usepackage{nicefrac}       %
\usepackage[round]{natbib}  %
\usepackage{fullpage}
\usepackage{authblk}

\usepackage{algorithm}
\usepackage{algpseudocode}

\usepackage{enumitem}
\usepackage[titletoc,title]{appendix}
\usepackage{multirow}
\usepackage{xspace}

\usepackage{Definitions}
\usepackage{tikz}
\usetikzlibrary{positioning,arrows.meta,fit,calc}
\usepackage[normalem]{ulem}
\usepackage{xcolor}
\usepackage{cancel}

\usepackage{Definitions}

\providecommand{\keywords}[1]
{
  \textbf{{Keywords---}} #1
}

\usepackage{lastpage}

\usepackage[capitalize,noabbrev]{cleveref}

\begin{document}

\title{Spectral Ghost in Representation Learning:\\ 
A Unified View from Component Analysis to Self-Supervised Learning}

\author{
    \normalsize
   {Bo Dai}$^1$,  
   {Na Li}$^2$,  {Dale Schuurmans}$^3$\\
   $^1$Georgia Tech\quad $^2$Harvard University\quad $^3$University of Alberta
}

\date{}

\maketitle

\begin{abstract}%
    Self-supervised learning~(SSL) extracts representations from massive unlabeled data that transfer to a variety of downstream tasks. Its empirical success has produced a dizzying array of objectives and procedures, yet a clear understanding of what representation is desired, and why these seemingly different methods work, is still missing.
    In this paper, we develop a unified spectral framework for representation learning. We first show that a representation is sufficient for arbitrary downstream prediction with a linear head if it spans the spectral decomposition of the pointwise mutual information, which can be extracted purely from unlabeled data. 
    This view organizes representation learning along two axes, the parametrization of the spectral representation (linear, energy-based, denoising, nonlinear, latent-variable, and multi-modal) and the learning objective for fitting it, under which a broad range of existing methods, including SimCLR, MoCo, BYOL, Barlow Twins, VICReg, DINO, MAE, JEPA, BERT, GPT, and CLIP, as well as classic PCA, CCA, Laplacian eigenmaps, LLE, and SNE, are recovered as particular combinations. 
    The framework also clarifies optimization behaviors: the batch-size requirement is determined by whether the objective is unbiased, a consistent surrogate, or biased, rather than by contrastive vs.\ non-contrastive losses. 
    Finally, the framework inspires new scalable algorithms and extends spectral representations to causal inference and reinforcement learning.
\end{abstract}


\newpage
\tableofcontents

\newpage
\section{Introduction}\label{sec:intro}

Representation learning has demonstrated tremendous success across a wide variety of machine learning tasks---ranging from classical regression and classification~\citep{chen2020simple,grill2020bootstrap}, to object detection~\citep{he2020momentum}, causal inference~\citep{wang2022spectral, sun2024spectral}, and reinforcement learning~\citep{ren2022spectral,zhang2022making}---and across multiple modalities---such as text, images, videos, audio and time series~\citep{meng2021coco, giorgi2020declutr, goyal2021self,schiappa2023self,liu2022audio,wickstrom2022mixing}. 
Such methods have improved empirical performance by unleashing the power of unlabeled data for practical applications in a \emph{self-supervised} way. Specifically, these algorithms formulate a training task based on unlabeled data---\eg, next-token prediction~\citep{radford2018improving, radford2019language,brown2020language}, masked token prediction~\citep{mikolov2013efficient, devlin2018bert, he2022masked}, noise contrastive classification~\citep{chen2020simple,oord2018representation}, or denoising~\citep{vincent2008extracting}---and learn a data representation by completing such a predefined task. The diverse application of representation learning has attracted increasing attention in the machine learning community, resulting in a variety of learning objectives, such as contrastive~\citep{oord2018representation,radford2021learning} vs.\ non-contrastive~\citep{zbontar2021barlow, garrido2022duality} losses, stationary~\citep{chen2020simple,he2020momentum} vs.\ non-stationary negative sampling~\citep{grill2020bootstrap}, and so on. 

Given the large number of distinct paradigms proposed for representation learning algorithms with empirical successes, there have been a few attempts to clarify the connections~\citep{balestriero2022contrastive,balestriero2023cookbook}. Particularly,~\citet{balestriero2022contrastive} relate contrastive and non-contrastive methods to global and local spectral embedding methods, and~\citet{johnson2023contrastive} interpret several contrastive objectives as approximating the positive-pair kernel $\frac{\PP\rbr{x, x'}}{\PP\rbr{x}\PP\rbr{x'}}$, and consider its kernel PCA as the representation. However, a clear understanding of desired properties of the representation is still absent. 
Meanwhile, even though they follow the same \emph{self-supervised} spirit, a common framework for characterizing the properties of the existing SSL algorithms, and analyzing the relationships among these methods is long overdue.  
These gaps create an unnecessarily high barrier to representation learning research and algorithm design, both from the theoretical and empirical perspectives.

Concretely, despite the empirical successes achieved by representation from SSL, there are essential research questions that have yet to be resolved, \ie, 
\begin{center}
    \begin{itemize}
    \item \emph{What representation is sufficient and effective for a variety of downstream tasks?} 
    \item \emph{How can such a representation be learned in an efficient and scalable way?}
    \end{itemize}
\end{center}
which have never been answered explicitly. Here, by ``sufficient" we mean that the downstream tasks can be completed by composing functions only upon these representations, instead of from the original data. This requires the representations to capture enough information while ignoring unnecessary details. By ``effective", we mean that the composed function upon representation for downstream tasks should be light-weight, rather than large deep models. By ``efficient" and ``scalable", we mean the optimization for the representation learning should be with less computational and memory cost, and can be easily scaled up. 
The unlaid theoretical foundation casts a shadow over the empirical successes of SSL representation learning.

On the other hand, due to disparate origins and the variety of algorithmic realizations, current representation learning algorithms use distinct components and exhibit dramatically different behaviors. Without a unified understanding, it is not clear whether the existing methods achieve the desired properties, and difficult to analyze these algorithms in the same framework for fair comparison. 

The absence of a unified framework for representation learning also leads to unnecessary barriers for pragmatic researchers. One must exhaustively probe these algorithms to fully grasp the dizzying set of choices in each ad hoc recipe in an isolated way, which not only makes the application of such algorithms extremely complicated, but also potentially compromises empirical performance due to suboptimal design of algorithm components. Such unnecessary complications also limit the use of representation learning methods in broader applications.

These urgent questions are in dire need of investigation in SSL to answer the foundational question with a unified view. 

\paragraph{Contribution.}  In this paper, we reveal the sufficiency of representation from spectral view, under which we develop a unified framework that empowers theoretical researchers to analyze existing algorithms, develop novel algorithms, and to extend the range of viable applications for representation learning. Particularly, our contribution lies in the following aspects.

\begin{itemize}
    \item We provide the \emph{spectral view} for characterizing the desired properties to the sufficient representation in~\secref{sec:sufficient_spectral}, for both the symmetric target $\PP\rbr{x'|x}$ and the asymmetric target $\PP\rbr{y|x}$. Moreover, we also reveal the sufficient representation purely upon unlabeled data, which paves the way for efficient spectral representation learning. 
    \item The spectral view induces \emph{a unified framework} for representation learning along two axes: the \emph{parametrization} of the spectral representation, \ie, linear, energy-based, denoising, nonlinear, latent-variable, and multi-modal, in~\secref{sec:spec_repr}--\ref{sec:mm_spectral}, and the \emph{learning objective} for fitting it in~\secref{subsec:spectral_learning}, with the batch-size analysis in~\secref{sec:objective_discussion}. Existing SSL methods are recovered as particular combinations, as summarized in~\figref{fig:overview} and~\tabref{tab:overview}, and the classic component analysis and manifold learning methods are revisited in~\secref{sec:spectral_history}. 
    \item The framework clarifies the optimization behavior of SSL methods: the batch-size requirement is determined by whether the objective is unbiased, a consistent surrogate, or biased w.r.t.\ the mini-batch, rather than by contrastive vs.\ non-contrastive losses: the contrastive ranking-based NCE in SimCLR and CLIP is consistent with any batch size and, under a well-specified model, only gains statistical efficiency from more negative samples, while the non-contrastive Barlow-Twins and VICReg are biased and inconsistent with small batch sizes. 
    \item The unified framework also inspires a variety of novel scalable representation learning algorithms in~\secref{subsec:spectral_learning},~\ref{subsec:more_ebm_spectral},~\ref{subsec:denoising_spectral_learning},~\ref{subsec:more_nonlinear_learning}, and~\ref{subsec:more_lv_spectral} for different spectral representations. 
    \item Finally, we exploit spectral representations for causal inference, control, reinforcement learning, and animal behavior analysis in~\secref{sec:application_spectral}, which extends their applicability and effectiveness.
\end{itemize}

\begin{figure}[t]
\centering
\resizebox{\linewidth}{!}{%
\begin{tikzpicture}[
  font=\footnotesize,
  core/.style={draw,thick,rounded corners,fill=gray!12,align=center,inner sep=5pt},
  head/.style={draw,thick,rounded corners,fill=blue!10,align=center,text width=3.0cm,minimum height=0.95cm},
  meth/.style={draw,rounded corners,align=center,text width=3.0cm,minimum height=3.7cm,anchor=north},
  band/.style={draw,thick,rounded corners,fill=orange!12,align=center,inner sep=5pt,minimum width=19.9cm},
  arr/.style={-{Stealth},thick}]
\node[core] (core) at (8.375,0) {\textbf{Sufficiency of spectral representation} (\S\ref{sec:sufficient_spectral})\\[2pt]
  $\displaystyle \frac{\PP(x'|x)}{\PP(x')}=\inner{\phi(x)}{\phi(x')}$, \quad core set $z$:\ $x',y$ depend on $x$ only through $z$};
\foreach \i/\name/\sec/\ms in {
  0/Linear/sec:spec_repr/{Spectral contrastive\\Barlow-Twins\\VICReg\\BYOL, MINC\\[4pt]\textcolor{gray}{\emph{Classic} (\S\ref{sec:spectral_history}):}\\\textcolor{gray}{PCA, CCA, MDS, LE,}\\\textcolor{gray}{LLE, Isomap, SNE}},
  1/Energy-based/sec:energy_spectral/{SimCLR\\word2vec\\MoCo},
  2/Denoising/sec:denoising_spectral/{Diffusion SSL\\DAE\\BERT\\GPT},
  3/Nonlinear/sec:nonlinear_spectral/{MAE\\SimSiam\\I-JEPA, V-JEPA\\[4pt]\textcolor{gray}{\emph{Classic}: t-SNE}},
  4/Latent-variable/sec:lv_spectral/{DeepCluster\\SeLa\\DINO, SwAV},
  5/Multi-modal/sec:mm_spectral/{CLIP, SigLIP\\SogCLR\\AmorLIP, NeuCLIP}}{
  \node[head] (h\i) at (3.35*\i,-2.1) {\textbf{\name}\\(\S\ref{\sec})};
  \node[meth] (m\i) at ($(h\i.south)+(0,-0.15)$) {\ms};
  \draw[arr] (3.35*\i,-1.2) -- (h\i.north);
}
\draw[thick] (core.south) -- (8.375,-1.2);
\draw[thick] (0,-1.2) -- (16.75,-1.2);
\node[band] (obj) at (8.375,-7.55) {\textbf{Learning objectives} (\S\ref{subsec:spectral_learning}) \quad\&\quad \textbf{batch-size analysis} (\S\ref{subsec:objective_batch})\\[2pt]
  square loss \ $\cdot$\ binary / ranking NCE \ $\cdot$\ density ratio \ $\cdot$\ projected square loss \ $\cdot$\ penalized orthogonality / unbiased VICReg\\
  (variational) power iteration \ $\cdot$\ denoising / MLM loss \ $\cdot$\ (approximate) MLE, EM};
\foreach \i in {0,...,5}{\draw[{Stealth}-,thick,orange!70!black] (m\i.south) -- (m\i.south |- obj.north);}
\node[band,fill=green!10] (app) at (8.375,-9.05) {\textbf{Downstream predictors} (\S\ref{subsec:downstream_predictor}) \ $\cdot$\ \textbf{Applications} (\S\ref{sec:application_spectral}): causal inference, control, reinforcement learning, animal behavior};
\draw[arr] (obj.south) -- (app.north);
\end{tikzpicture}}
\caption{Overview of the spectral framework. The sufficiency of the spectral representation (top) induces different \emph{parametrizations} (middle), each fitted by a \emph{learning objective} (bottom); existing SSL methods and classic component analysis methods are recovered as particular combinations, as listed in~\tabref{tab:overview}.}
\label{fig:overview}
\end{figure}

\begin{table}[h]
\centering
\caption{Existing representation learning methods in the spectral framework.}
\label{tab:overview}
\small
\resizebox{\linewidth}{!}{\begin{tabular}{llll}
\toprule
Method & Parametrization & Learning objective & Section \\
\midrule
Spectral contrastive & Linear & Square loss & \ref{subsec:spectral_ssl} \\
Barlow-Twins, VICReg & Linear & Penalized orthogonality & \ref{subsec:spectral_ssl} \\
BYOL, MINC & Linear & Variational power iteration & \ref{subsec:spectral_ssl} \\
SimCLR, word2vec & Energy-based & Ranking / binary NCE & \ref{subsec:ebm_ssl} \\
MoCo & Energy-based & Power iteration with ranking NCE & \ref{subsec:ebm_ssl} \\
Diffusion SSL & Denoising (lifted energy-based) & Diffusion (square) loss & \ref{subsec:denoising_spectral_learning} \\
DAE & Denoising (asymmetric energy-based) & Denoising (square) loss & \ref{subsec:denoising_spectral_learning} \\
BERT~(MLM) & Denoising (asymmetric energy-based) & MLM (single-ratio discrete diffusion) & \ref{subsec:denoising_spectral_learning} \\
GPT & Denoising (asymmetric energy-based) & Exact MLE & \ref{subsec:denoising_spectral_learning} \\
MAE & Nonlinear (asymmetric) & Projected square loss & \ref{subsec:nonlinear_spectral_learning} \\
SimSiam, I-JEPA, V-JEPA & Nonlinear (asymmetric) & Variational power iteration & \ref{subsec:nonlinear_spectral_learning} \\
DeepCluster, SeLa & Latent-variable & EM for MLE & \ref{subsec:lv_spectral_learning} \\
DINO, SwAV & Latent-variable & Variational power iteration & \ref{subsec:lv_spectral_learning} \\
CLIP, SigLIP & Multi-modal energy-based & Ranking / binary NCE & \ref{subsec:mm_ebm_ssl} \\
SogCLR, AmorLIP, NeuCLIP & Multi-modal energy-based & Approximate MLE & \ref{subsec:mm_ebm_ssl} \\
\bottomrule
\end{tabular}}
\end{table}
\section{Sufficiency of Spectral Representation}\label{sec:sufficient_spectral}

\subsection{Sufficient Representation from Unlabeled Data}\label{subsec:symmetric_spectral}

To reveal the desired properties for the sufficient representation, we start with prediction tasks for estimating $\EE\sbr{y|x}$ from data, in which one needs to learn a function $f\rbr{\cdot}: \Xcal\rightarrow \Ycal$, where $\Ycal$ can be $\RR^k$ for regression, $1$-of-$k$ for classification, and more complex structured space.
We denote 
\begin{equation}\label{eq:svd}
    \PP\rbr{y|x} = \inner{\phi\rbr{x}}{\mu\rbr{y}},
\end{equation}
as the singular value decomposition~(SVD) of the conditional operator $\PP\rbr{y|x}$, where $\phi\rbr{\cdot}: \Xcal\rightarrow \RR^d$ and $\mu\rbr{\cdot}:\Ycal\rightarrow \RR^d$. We emphasize the SVD of $\PP\rbr{y|x}$ exists as long as the conditional operator is compact, \eg, Hilbert-Schmidt, but can potentially be infinite dimension $d$. For notation simplicity, we denote $\inner{\phi\rbr{x}}{\mu\rbr{y}} = \phi\rbr{x}^\top \mu\rbr{y}$ for finite $d$-dimensional representation, and  $\inner{\phi\rbr{x}}{\mu\rbr{y}} = \int \phi_\omega\rbr{x} \mu_\omega\rbr{y} p\rbr{\omega} d\omega$ for infinite-dimensional representation with $p\rbr{\omega}$ as some measure over indexing variable $\omega$. 
Then, we have 
\begin{equation}\label{eq:cond_mean}
    \EE\sbr{y|x} = \int y \PP\rbr{y|x}dy = \inner{\phi\rbr{x}}{\underbrace{\int y\mu\rbr{y}dy}_w}
\end{equation}
which means the optimal mean square solution can be \emph{sufficiently} represented by $\phi\rbr{x}$ in a \emph{linear form}, therefore, satisfying our requirement.  

However, the $\phi\rbr{x}$ is task-dependent as a factorization of $\PP\rbr{y|x}$, therefore, it is only sufficient for a particular regression task upon $\PP\rbr{y|x}$. For a different task induced by a different $\tilde\PP\rbr{y|x} = \inner{\tilde\phi\rbr{x}}{\tilde\mu\rbr{y}}$, we may need different $\tilde\phi\rbr{x}$, whose corresponding subspace is different from $\phi\rbr{x}$.

A straightforward idea is seeking the \emph{complete space} composed by all the subspaces $\phi\rbr{x}$ for all $\PP\rbr{y|x}$, denoted as $\psi\rbr{x}$.
With the complete $\psi\rbr{x}$ provided, it is enough to predict arbitrary dependent $y$ in any down stream tasks. Specifically, for arbitrary task with a corresponding $\PP\rbr{y|x}$, there exists one linear subspace of the span of $\psi\rbr{x}$ that covers the singular subspace of $\PP\rbr{y|x}$, and therefore, is sufficient for representing $\EE\sbr{y|x}$ linearly. 
It is intractable to enumerate all the possible $\PP\rbr{y|x}$ for all random variable $y$ to recover the complete space for sufficient representation $\psi\rbr{x}$. We consider a $\nu\rbr{z}$, which has the same rank with $\psi\rbr{x}$, and forms $\PP\rbr{z|x} \defeq \inner{\psi\rbr{x}}{\nu\rbr{z}}$, as the \emph{core set}, such that any other variable, including $x'$ and the downstream $y$, depends on $x$ only through $z$. Although $\nu\rbr{z}$ and the core set is still conceptual and unknown, we consider
\begin{align}\label{eq:sufficient_repr}
    \PP\rbr{x'| x} &= \int \PP\rbr{x'|z} \PP\rbr{z|x}dz = \PP\rbr{x'}\inner{\psi\rbr{x'}}{\underbrace{\rbr{\int \frac{\nu\rbr{z}\nu\rbr{z}^\top }{\PP\rbr{z}}dz}}_{A}\psi\rbr{x}},\\
    &= \PP\rbr{x'}\inner{\psi\rbr{x'}}{A\psi\rbr{x}},
\end{align}
with the second equation comes from Bayes' rule, \ie, 
    $\PP\rbr{x'| z} = \frac{\PP\rbr{z|x'}\PP\rbr{x'}}{\PP\rbr{z}}$.
This implies we can extract the sufficient representation $\psi\rbr{x}$ from~\eqref{eq:sufficient_repr} purely from unlabeled data, rather than exploiting prediction tasks with labeled data. 

In~\eqref{eq:sufficient_repr}, as $\nu\rbr{z}$ sharing the same rank to $\psi\rbr{x}$, \ie, the components of $\nu\rbr{z}$ are linearly independent, $A\in \RR^{d\times d}$ is a full-rank positive-definite matrix, which can be factorized as $A = B^\top B$ with $B\in\RR^{d\times d}$, which leads to 
\begin{equation}\label{eq:reparam_repr}
    \PP\rbr{x'|x} =  \PP\rbr{x'}\inner{\underbrace{B\psi\rbr{x'}}_{\varphi\rbr{x'}}}{ B\psi\rbr{x}} = \PP\rbr{x'}\inner{\varphi\rbr{x'}}{\varphi\rbr{x}} \Rightarrow \frac{\PP\rbr{x, x'}}{\PP\rbr{x}\PP\rbr{x'}} = \inner{\varphi\rbr{x'}}{\varphi\rbr{x}}.    
\end{equation}
Since $A$ is full-rank, so as $B$. Therefore, the linear space constructed by $\psi\rbr{x}$ and $\varphi\rbr{x}$ are the same. Therefore, we can also represent $\EE\sbr{y|x}$ linearly with $\varphi\rbr{x}$, \ie, 
$\EE\sbr{y|x} = \inner{\varphi\rbr{x}}{w}$, through the spectral decomposition of Pointwise Mutual Information (PMI), implied by~\eqref{eq:reparam_repr}. Similar linear sufficiency of contrastive representations has been established under specialized assumptions, \eg, multi-view redundancy~\citep{tosh2021contrastive} and topic models~\citep{tosh2021topic}, which are special cases of the spectral view here. 

As we have clarified the requirement for ``sufficient and effective" representation from the spectral perspective, the immediate followup question is whether the current representations learned from the existing SSL methods satisfy the condition, which has been answered affirmatively in following sections. 

\subsection{Task-Dependent and Asymmetric Spectral Representation}\label{subsec:asymmetric_spectral}

We have completed the sufficiency in representation through spectral view for $\Xcal$, which seeks for complete space from the extra neighborhood information $\PP\rbr{x'|x}$. Conceptually, the extracted spectral representation $\varphi\rbr{x}$ from $\PP\rbr{x'|x}$ will be sufficient for arbitrary down stream tasks. However, the general sufficient representation includes all the spectral representations from arbitrary $\PP\rbr{y|x}$, which might be not necessary for some scenarios where the downstream tasks are predefined, \eg, in natural language processing, the downstream tasks can be recast as next-token prediction~\citep{radford2019language}, and in multi-modality problems, the downstream tasks can be recast as retrieval~\citep{radford2021learning}, one can extract the corresponding task-dependent spectral representation, instead of the generic sufficient spectral representation for statistical, memory and computational efficiency. 

Particularly, in multi-modality setting, the paired data $\cbr{\rbr{x, y}}_{i=1}^n\sim \PP\rbr{y|x}$ are given, then, we can directly extract sufficient representation $\phi\rbr{x}$ and $\mu\rbr{y}$ from spectral decomposition of $\PP\rbr{y|x}$ in~\eqref{eq:svd}, or equivalently of the PMI $\frac{\PP\rbr{x, y}}{\PP\rbr{x}\PP\rbr{y}}$ with $\PP\rbr{y}$ absorbed into $\mu\rbr{y}$, towards to the target tasks. Given the spectral representation, we can sufficient represent $\EE\sbr{y|x}$, recover $\PP\rbr{y|x}$, and so on, to complete the particular downstream tasks.

In other words, the spectral representation can be extracted from two types of target operators: the \emph{symmetric} $\PP\rbr{x'|x}$ in~\eqref{eq:reparam_repr}, where $x'$ is a neighbor or augmentation of $x$ in the same space, and the \emph{asymmetric} $\PP\rbr{y|x}$ in~\eqref{eq:svd}, where $y$ comes from a different space, \eg, another modality. Most of the following sections are presented with the symmetric target, and the asymmetric target is elaborated in~\secref{sec:denoising_spectral}, where the asymmetry comes from the corruption, and in~\secref{sec:mm_spectral}, where the asymmetry comes from different modalities.

\subsection{Sufficiency for Regression and Classification}\label{subsec:regression_classification}

The motivation for linear spectral sufficient representation comes from linearly representing all the optimal regressors to downstream least square regression, as shown in~\eqref{eq:cond_mean}.

The same idea can be generalized for classification. Recall the decision rule, which minimizes the conditional risk, \ie, 
\begin{equation}
    R\rbr{y=i|x} = \sum_{j=1}^k\omega\rbr{y=i, y=j}\PP\rbr{y=j|x} \Rightarrow R\rbr{\cdot|x}  = \inner{\varphi\rbr{x}}{\underbrace{\sum_{j=1}^k\omega\rbr{\cdot, y=j} \mu\rbr{y=j}}_{W\defeq \Omega\mu\rbr{\cdot}\in \RR^{d\times k}}}.
\end{equation}
It implies that the conditional risk can be linearly represented by $\varphi\rbr{x}$, and the classification task can be completed by
\begin{equation}
    \argmin_{y\in \cbr{1, \ldots, k}}\,\, \inner{W_y}{\varphi\rbr{x}}.
\end{equation}
Particularly, if we set
$\omega\rbr{i, j} = 
\begin{cases}
    0, &\quad \text{if}\quad i = j\\
    1, &\quad \text{if}\quad i\neq j
\end{cases}$, the conditional risk reduces to $R\rbr{y|x} = 1 - \PP\rbr{y|x}$, which recovers the Bayes decision rule, \ie, 
\begin{equation}
    \argmax_{y\in \cbr{1, \ldots, k}}\,\, \PP\rbr{y|x} = \argmin_{y\in \cbr{1, \ldots, k}}\,\, \inner{W_y}{\varphi\rbr{x}}.
\end{equation}

As we can linearly represent the posterior $\PP\rbr{y|x}$ by $\varphi\rbr{x}$, the related linear statistics, \eg, mean, moments, and characteristic function, can obviously be represented linearly as well, while other statistics, \eg, median, can be represented nonlinearly. Therefore, the function family linearly/nonlinearly-composed by sufficient representation can be used for any loss function whose solution is some statistics w.r.t. $\PP\rbr{y|x}$, \eg, logistic/softmax classification, robust $L1$-loss for regression, and so on.

\section{Linear Spectral Representation}\label{sec:spec_repr}

In this section, and the following sections, we will justify the existing methods through the unified spectral framework, which naturally categorizes the SSL methods through their parametrizations: the direct linear factorization~\eqref{eq:reparam_repr} in this section, the energy-based spectral representation in~\secref{sec:energy_spectral}, the denoising spectral representation in~\secref{sec:denoising_spectral}, the nonlinear and latent-variable spectral representations in~\secref{sec:nonlinear_spectral} and~\ref{sec:lv_spectral}, and the multi-modal spectral representation in~\secref{sec:mm_spectral}. For each parametrization, we first recover the existing methods, and then introduce the corresponding new algorithms.

\subsection{Spectral Representation View for Existing SSL}\label{subsec:spectral_ssl}

The spectral representation view naturally connects a variety existing SSL algorithms, including contrastive and non-contrastive, which were originally derived from different perspectives with different optimization procedures, but approaching the same target representations. 
\begin{itemize}[leftmargin=*]
    \item {\bf Square Contrastive Representation~\citep{haochen2021provable}.} Recall the spectral representation is characterized by~\eqref{eq:reparam_repr}. However, directly matching the LHS and RHS of~\eqref{eq:reparam_repr} may not lead to tractable objective. By multiplying $\sqrt{\PP\rbr{x}\PP\rbr{x'}}$ to both sides of~\eqref{eq:reparam_repr}, we obtain 
    \begin{equation}\label{eq:spectral_match}
        {\frac{\PP\rbr{x', x}}{\PP\rbr{x}\PP\rbr{x'}}} = \inner{\varphi\rbr{x'}}{\varphi\rbr{x}} \Rightarrow \underbrace{{\frac{\PP\rbr{x', x}}{\sqrt{\PP\rbr{x}\PP\rbr{x'}}}}}_{T\rbr{x, x'}} = \sqrt{\PP\rbr{x}\PP\rbr{x'}}\inner{\varphi\rbr{x'}}{\varphi\rbr{x}}. 
    \end{equation}
    We define the normalized transition operator
    \begin{align}
        \label{eq:T}
        T\rbr{x,x'}&=\frac{\PP\rbr{x', x}}{\sqrt{\PP\rbr{x}\PP\rbr{x'}}},
    \end{align}
    which plays a key role in further developments below.
    Concretely, we can now apply
    mean square matching to $T\rbr{x,x'}$, leading to the spectral contrastive loss for spectral representation learning, \ie, 
    \begin{align}   
        \min_{\varphi}\,\, &\ell\rbr{\varphi}\defeq \int \nbr{T(x, x') - \sqrt{\PP\rbr{x}\PP\rbr{x'}}\inner{\varphi\rbr{x'}}{\varphi\rbr{x}}}^2 dx dx' \\
        &= \underbrace{\int T(x, x')^2 dxdx'}_{\text{constant}} - 2\EE_{\PP\rbr{x, x'}}\sbr{\varphi(x)^\top \varphi\rbr{x'}} + \EE_{\PP\rbr{x}\PP\rbr{x'}}\sbr{\rbr{\varphi(x)^\top \varphi\rbr{x'}}^2}, \label{eq:spectral_contrastive}
    \end{align}
    which has been investigated in~\citep{haochen2021provable,ren2022spectral}. Obviously, the obtained spectral contrastive loss is compatible with stochastic gradient-based optimizer, therefore, can be solved efficiently.

    \item {\bf Barlow-Twins Representation~\citep{zbontar2021barlow}.}
    We consider the linear projected spectral representation, \ie, $\xi\rbr{x} = Z^\top \varphi\rbr{x}$. Obviously, with any full-rank linear projection $Z$, $\xi\rbr{x}$ spans the same subspace as $\varphi\rbr{x}$. Particularly, we denote
    \[
        A \defeq \EE\sbr{\varphi\rbr{x}\varphi\rbr{x}^\top} = \int \varphi\rbr{x}\varphi\rbr{x}^\top \PP\rbr{x}dx = U\Lambda U^\top,
    \]
    where the last equation comes from eigen-decomposition of symmetric $A$ with $U^\top U = I \in \RR^{d\times d}$. If we set $Z = U\Lambda^{-\frac{1}{2}}$, $\xi\rbr{x}$ satisfies the orthonormal condition $\EE\sbr{\xi\rbr{x}\xi\rbr{x}^\top} = Z^\top A Z = \Ib$, and we have the condition to characterize the spectral representation as
    \begin{align}\label{eq:normalized_spec}
        &\int T\rbr{x, x'}\sqrt{ \PP\rbr{x}}\xi\rbr{x} \sqrt{ \PP\rbr{x'}}\xi\rbr{x'}^\top dx dx' \nonumber \\
        =&   \inner{\int { \PP\rbr{x}}\xi\rbr{x}\varphi\rbr{x}dx}{\int \varphi\rbr{x'} {\PP\rbr{x'}}\xi\rbr{x'}^\top dx' }  = Z^\top A^\top A Z = \Lambda, 
    \end{align}
    where the last equation comes from the fact of the definition of $Z$, and $\Lambda$ collects the eigenvalues of the pointwise mutual information. These eigenvalues are at most $1$: for any $f$ with $\EE\sbr{f\rbr{x}^2} = 1$, by Cauchy-Schwarz inequality, 
    \begin{equation}\label{eq:pmi_eigen_bound}
        \EE_{\PP\rbr{x, x'}}\sbr{f\rbr{x}f\rbr{x'}} \le \sqrt{\EE_{\PP\rbr{x}}\sbr{f\rbr{x}^2}\EE_{\PP\rbr{x'}}\sbr{f\rbr{x'}^2}} = 1,
    \end{equation}
    since $x$ and $x'$ share the marginal $\PP\rbr{x}$, and the maximum $1$ is attained by the constant function $f\equiv 1$, which is trivial and removed by centering. 
    It implies the learning objective, 
    \begin{equation}\label{eq:barow_twins_I}
        \min_{\xi}\,\, \nbr{\Ib - \EE_{\PP\rbr{x, x'}}\sbr{\xi\rbr{x} \xi\rbr{x'}^\top}}_F^2, \quad \st\,\, \EE\sbr{\xi\rbr{x}\xi\rbr{x}^\top} = \Ib.
    \end{equation}
    Without the orthonormal constraint, the eigenvalues are normalized away, \eg, $\xi_k = e_k/\sqrt{\gamma_k}$ for any eigenfunctions $e_k$ with eigenvalues $\gamma_k$ leads to $\EE_{\PP\rbr{x, x'}}\sbr{\xi\rbr{x} \xi\rbr{x'}^\top} = \Ib$, so the top eigen-space cannot be identified. With the constraint, the eigenvalues of $\EE_{\PP\rbr{x, x'}}\sbr{\xi\rbr{x} \xi\rbr{x'}^\top}$ are at most those of the top-$d$ eigen-space by the Poincar\'e separation theorem, and are at most $1$ by~\eqref{eq:pmi_eigen_bound}, therefore,~\eqref{eq:barow_twins_I} is minimized by the top-$d$ eigen-space.
    Follow the notation from~\citet{zbontar2021barlow}, 
    \[
        C = \EE_{\PP\rbr{x, x'}}\sbr{\xi\rbr{x} \xi\rbr{x'}^\top} = \Lambda^{-\frac{1}{2}}U^\top \EE_{\PP\rbr{x, x'}}\sbr{\varphi\rbr{x} \varphi\rbr{x'}^\top} U \Lambda^{-\frac{1}{2}} \in \RR^{d\times d},
    \]
    where the $\Lambda\in\RR^{d\times d}$ is a diagonal matrix and can be estimated through
    \begin{align}\label{eq:orth_cond}
         U^\top\EE\sbr{\varphi\rbr{x}\varphi\rbr{x}^\top}U = \Lambda.
    \end{align}
    With $U$ and $\Lambda$ given by~\eqref{eq:orth_cond}, $\xi\rbr{x} = \Lambda^{-\frac{1}{2}}U^\top\varphi\rbr{x}$ is the whitening of $\varphi\rbr{x}$, which satisfies the orthonormal constraint in~\eqref{eq:barow_twins_I} by construction. Therefore, the constraint is enforced by the reparametrization, and the reformulation below is unconstrained in $\varphi$. Barlow-Twins approximates the whitening by batch normalization, which only standardizes each dimension.
    Then, we can reformulate~\eqref{eq:barow_twins_I} equivalently as
    \begin{equation}
        \min_{\varphi}\,\, \sum_{i=1}^d\rbr{1 - C_{ii}}^2 + \sum_{i\neq j} C_{ij}^2,
    \end{equation}
    whose minimizer spans the same top-$d$ eigen-space as that of
    \begin{equation}\label{eq:barow_twins_II}
        \min_{\varphi}\,\, \sum_{i=1}^d\rbr{1 - C_{ii}}^2 + \lambda \sum_{i\neq j} C_{ij}^2,
    \end{equation}
    for any $\lambda\ge 0$, possibly with a different basis within the eigen-space, as shown in~\appref{app:bt_lambda},
    recovering the exact Barlow-Twins objective~\citep{zbontar2021barlow}. 
    
    \paragraph{Remark (Optimization Difficulty):} Although we have revealed that the Barlow-Twin formulation is also seeking a spectral representation, the objective in Barlow-Twins~\eqref{eq:barow_twins_II} is actually not compatible with a stochastic gradient algorithm: the gradient calculation to~\eqref{eq:barow_twins_II} is biased, and thus, may lead to inferior performance due to the bias induced by approximating the expectation. Concretely, as shown in~\appref{app:bt_bias}, with the batch normalization, the expected mini-batch loss with batch size $B$ equals the population loss plus $\frac{1}{B}R\rbr{\varphi} + \Ocal\rbr{\frac{1}{B^2}}$, where $R\rbr{\varphi}$ depends on the fourth moments of $\varphi$, so the minimizer of the mini-batch loss differs from the population one. The bias comes from two sources: the square of the estimated cross-correlation, which can be removed by U-statistics, and the division by the estimated variance in the batch normalization, \ie, the whitening, which cannot be removed as long as the batch statistics are used. This observation implies that the Barlow-Twins method also requires a large batch size to reduce the sample bias, otherwise the solution is inconsistent, even though it is not a contrastive objective, which was the original motivation of Barlow-Twins representation.

    \item{\bf Variance-Invariance-Covariance Regularization~(VICReg)~\citep{bardes2021vicreg}.} With the same notations, we consider another linear setup $Z = U\Lambda^{-\frac{1}{2}}$, then, we have $\xi\rbr{x} = Z^\top \varphi\rbr{x}$ satisfying orthonormal condition, \ie, 
    \begin{equation}\label{eq:orthonomal_cond}
        \EE\sbr{\xi\rbr{x}\xi\rbr{x}^\top} = Z^\top\EE\sbr{\varphi\rbr{x}\varphi\rbr{x}^\top}Z = \Ib. 
    \end{equation}
    Then, we have the variational characteristic of spectral decomposition w.r.t. $\xi\rbr{x}$, \ie, 
    \begin{equation}\label{eq:svd_variational}
        \max_{\xi}\,\, \int T\rbr{x, x'}\sqrt{ \PP\rbr{x}}\xi\rbr{x}^\top \sqrt{ \PP\rbr{x'}}\xi\rbr{x'} dx dx' = \EE_{\PP\rbr{x, x'}}\sbr{\xi\rbr{x}^\top \xi\rbr{x'}}.
    \end{equation}
    Combine~\eqref{eq:orthonomal_cond} and~\eqref{eq:svd_variational} with different penalties through the penalty method, we obtain 
    \begin{align}\label{eq:vicreg_I}
        \min_{\xi}\,\, & - \EE_{\PP\rbr{x, x'}}\sbr{\xi\rbr{x}^\top \xi\rbr{x'}} 
        +\eta \rbr{\frac{1}{d}\rbr{\sum_{i\neq j} \sbr{\cov\rbr{\xi\rbr{x}}}_{ij}^2 + \sum_{i\neq j}\sbr{\cov\rbr{\xi\rbr{x'}}}_{ij}^2}}\\
        & + \frac{\lambda}{d}\sum_{i=1}^d \max\rbr{0, 1 - \sqrt{\sbr{\cov\rbr{\xi\rbr{x}}}_{ii}}} + \frac{\lambda}{d}\sum_{i=1}^d \max\rbr{0, 1 - \sqrt{\sbr{\cov\rbr{\xi\rbr{x'}}}_{ii}}}, \nonumber 
    \end{align}
    which is almost the same as the original VICReg objective proposed in~\citep{bardes2021vicreg}, except replacing $ - \EE_{\PP\rbr{x, x'}}\sbr{\xi\rbr{x}^\top \xi\rbr{x'}}$ with $\EE_{\PP\rbr{x, x'}}\sbr{\nbr{\xi\rbr{x} - \xi\rbr{x'}}^2}$. In fact, recall the fact that under the orthonormal condition~\eqref{eq:orthonomal_cond}, 
    \[
        \EE_{ \PP\rbr{x, x'}}\sbr{\nbr{\xi\rbr{x} - \xi\rbr{x'}}^2} = -2 \EE_{\PP\rbr{x, x'}}\sbr{\xi\rbr{x}^\top \xi\rbr{x'}} + \underbrace{\EE\sbr{\xi\rbr{x}^\top\xi\rbr{x}}}_{d} + \underbrace{\EE\sbr{\xi\rbr{x'}^\top\xi\rbr{x'}}}_{d},
    \]
    we recover the VICReg. 
    Similar argument also applies to a variant of VICReg~\citep{balestriero2022contrastive} with square penalty w.r.t. covariance, \ie, 
    \begin{equation}\label{eq:vicreg_II}
    \min_{\xi}\,\, - 2\EE_{\PP\rbr{x, x'}}\sbr{\xi\rbr{x}^\top \xi\rbr{x'}} + \lambda \nbr{ \EE_{\PP(x)}\sbr{\xi\rbr{x}\xi\rbr{x}^\top}  - \Ib}_F^2. 
    \end{equation}

    \paragraph{Remark (Optimization Difficulty):} Similar to Barlow-Twins, the optimization objectives of VICReg~\eqref{eq:vicreg_I} and~\eqref{eq:vicreg_II} also contain terms that are not compatible with stochastic gradient descent algorithm, which may lead to inferior performances due to the bias induced by expectation approximation. Concretely, in~\eqref{eq:vicreg_I}, the square root and the hinge are nonlinear functions of the estimated variance, which induce the bias inherently. In~\eqref{eq:vicreg_II}, the penalty is a quadratic function of the expectation $\EE\sbr{\xi\rbr{x}\xi\rbr{x}^\top}$, and the plug-in estimator with the batch covariance $\widehat\Sigma = \frac{1}{B}\sum_{i=1}^B\xi\rbr{x_i}\xi\rbr{x_i}^\top$ leads to
    \begin{equation}\label{eq:vicreg_bias}
        \EE_B\sbr{\nbr{\widehat\Sigma - \Ib}_F^2} = \nbr{\EE\sbr{\xi\rbr{x}\xi\rbr{x}^\top} - \Ib}_F^2 + \frac{1}{B}\sum_{k, l}\operatorname{Var}\rbr{\xi_k\rbr{x}\xi_l\rbr{x}},
    \end{equation}
    where the extra term depends on the fourth moments of $\xi$, the same as the square part of the bias in Barlow-Twins. This observation implies the VICReg also requires large batch size to reduce the sample bias, even it is not contrastive objective, therefore, inspiring the investigation of alternative penalties to avoid trivial solutions. Different from Barlow-Twins, however,~\eqref{eq:vicreg_II} involves no division by the batch statistics, and its bias can be removed, which leads to a new algorithm in~\secref{subsec:spectral_learning}.

    \item{\bf BYOL~\citep{grill2020bootstrap} and MINC  Representation~\citep{guo2025representation}.} BYOL~\citep{grill2020bootstrap} and its variants~\citep{tian2021understanding, wang2021towards,richemond2023edge} can be recast as variants in implementing power iteration for spectral representation as explained in~\citep{guo2025representation}. We follow the notation and set $Z = U$, then, we  have 
    \begin{equation}\label{eq:orth_cond_I}
        \EE\sbr{\xi\rbr{x}\xi\rbr{x}^\top} = Z^\top\EE\sbr{\varphi\rbr{x}\varphi\rbr{x}^\top}Z = \Lambda, \quad \text{with}\quad  \xi\rbr{x} = Z^\top \varphi\rbr{x} 
    \end{equation}
    the eigenvalue property that
    \begin{align}
        &\int T\rbr{x, x'}\sqrt{\PP\rbr{x'}}\xi\rbr{x'}^\top dx' \\
        =&  \inner{\sqrt{\PP\rbr{x}}\varphi\rbr{x}}{\underbrace{\int \varphi\rbr{x'} \xi\rbr{x'}^\top \PP\rbr{x'}dx'}_{AZ = U\Lambda}}\\
        =& \sqrt{\PP\rbr{x}}\varphi\rbr{x}^\top U\Lambda = \sqrt{\PP\rbr{x}}\xi\rbr{x}^\top \Lambda,
    \end{align}
    which implies
    \begin{equation}\label{eq:eigen_fixpoint}
        \int T\rbr{x, x'}\sqrt{\PP\rbr{x'}}\xi\rbr{x'} dx' = \Lambda\sqrt{\PP\rbr{x}}\xi\rbr{x}. 
    \end{equation}
    The fixed point equation~\eqref{eq:eigen_fixpoint} of $\xi\rbr{x}$ induces the power iteration while ensuring the orthogonality~\eqref{eq:orth_cond_I} for spectral representation learning by following conceptual rule:
    \begin{align}
        \Lambda_{t+1} =& \EE\sbr{\xi_t\rbr{x}\xi_t\rbr{x}^\top}, \label{eq:lambda}\\
        \Lambda_{t+1}\sqrt{\PP\rbr{x}}\zeta_{t+1}\rbr{x} = &\int T\rbr{x, x'} \sqrt{\PP\rbr{x'}}\xi_{t}\rbr{x'}dx', \label{eq:var_iter}\\
        \sqrt{\PP\rbr{x}}\xi_{t+1}\rbr{x} = & \textbf{orth}\rbr{\sqrt{\PP\rbr{x}}\zeta_{t+1}\rbr{x}}, \label{eq:orth_op}
    \end{align}
    where $\textbf{orth}(\cdot)$ denotes a conceptual orthogonal operation. The exact integrals in the first two updates and the orthogonalization in the third update are intractable. However, with different variational updates, we arrive at different practical algorithms, including  MINC~\citep{guo2025representation} and BYOL~\citep{grill2020bootstrap}.

    Concretely, in MINC, exponential moving average (EMA) is adopted for approximating~\eqref{eq:lambda}, and orthogonal variational version of generalized Hebbian rule~\citep{sanger1989optimal, kim2005iterative, xie2015scale} for~\eqref{eq:var_iter} and~\eqref{eq:orth_op}, which leads to the update
    \begin{align}
        \Lambda_{t+1} &= \beta \Lambda_t + \rbr{1 - \beta}\EE\sbr{\xi_t\rbr{x}\xi_t\rbr{x}^\top},\\
        \xi_{t+1}\rbr{x} &= \xi_t\rbr{x} + \alpha_t\rbr{
        \EE_{\PP\rbr{x,x'}}\sbr{\xi_t\rbr{x'}^\top\nabla_\xi\xi_t\rbr{x}} - \EE\sbr{\xi_t\rbr{x}^\top {\color{blue}\texttt{LT}}\sbr{ \Lambda_{t+1}} \nabla_\xi\xi_t\rbr{x}}}, \label{eq:minc_gha}
    \end{align}
    where the operator ${\color{blue}\texttt{LT}}\rbr{\cdot}$ stands for making matrix lower triangular by setting all elements above the diagonal of its matrix argument to zero. As explained in~\citep{guo2025representation}, the update~\eqref{eq:minc_gha} is obtained by combining Gram-Schmidt process for orthogonalization to the gradient update of
    \begin{align}\label{eq:minc_obj}
        \min_{\xi}\,\, & \int \sbr{\nbr{\int \frac{\PP\rbr{x', x}}{\sqrt{\PP\rbr{x}}} \xi_t\rbr{x'}dx' -  \Lambda_{t+1} {\sqrt{\PP\rbr{x}}} \xi\rbr{x} }_{\Lambda_{t+1}^{-1}}^2}dx\\
        \propto& -2\EE_{\PP\rbr{x', x}}\sbr{\xi_t(x')^\top \xi\rbr{x}} + \EE\sbr{\xi\rbr{x}^\top \Lambda_{t+1}\xi\rbr{x}}.
    \end{align}

    Alternatively, BYOL exploits different way to implement the power iteration. Particularly, the $\Lambda$ plays as the eigen-value matrix in stationary solution, therefore, it should satisfy the variational characteristic of eigenvalues as shown in~\eqref{eq:eigen_fixpoint}. This property induces the learning rule of $\Lambda$ in BYOL. At $t$-th iteration, we have
    \begin{align}\label{eq:byol_lambda}
        &\min_{{\color{blue}\Lambda}}\,\, \int \sbr{\nbr{\int \frac{\PP\rbr{x', x}}{\sqrt{\PP\rbr{x}}} \xi_t\rbr{x'}dx' -  {\color{blue}\Lambda} {\sqrt{\PP\rbr{x}}} \xi_t\rbr{x}}_2^2}dx\\
        &\propto -2\EE_{\PP\rbr{x', x}}\sbr{\xi_t(x')^\top {\color{blue}\Lambda}\xi_t\rbr{x}} + \EE\sbr{\xi_t\rbr{x}^\top {\color{blue}\Lambda}^\top{\color{blue}\Lambda}\xi_t\rbr{x}}
        \propto \EE_{\PP\rbr{x', x}}\sbr{\nbr{{\color{blue}\Lambda} \xi_t\rbr{x} - \xi_t\rbr{x'}}_2^2}. \nonumber
    \end{align}
    Meanwhile, the fixed-point condition~\eqref{eq:eigen_fixpoint} also induces variational update rule for $\xi$, which is used for BYOL, 
    \begin{align}\label{eq:byol_xi}
        &\min_{{\color{blue}\xi}}\,\, \int \sbr{\nbr{\int \frac{\PP\rbr{x', x}}{\sqrt{\PP\rbr{x}}} \xi_t\rbr{x'}dx' -  {\Lambda} {\sqrt{\PP\rbr{x}}} {\color{blue}\xi\rbr{x}}}_2^2}dx\\
        &\propto -2\EE_{\PP\rbr{x', x}}\sbr{\xi_t(x')^\top {\Lambda}{\color{blue}\xi\rbr{x}}} + \EE\sbr{{\color{blue}\xi\rbr{x}}^\top {\Lambda}^\top{\Lambda}{\color{blue}\xi\rbr{x}}}
        \propto \EE_{\PP\rbr{x', x}}\sbr{\nbr{{\Lambda}{\color{blue}\xi\rbr{x}} - \xi_t\rbr{x'}}_2^2}. \nonumber
    \end{align}
    Combine~\eqref{eq:byol_lambda} and~\eqref{eq:byol_xi}, we obtain BYOL~\citep{grill2020bootstrap} with a linear predictor and without normalization, which is the variant analyzed in~\citet{tian2021understanding}. With exact updates, the iterates converge to the top-$d$ eigenspace without collapse or explicit orthogonalization, as shown in~\appref{app:byol}; the nonlinear predictor and the normalization will be discussed in~\secref{subsec:nonlinear_spectral_learning}. 

    Similarly, we can also derive other update rule for $\Lambda$ through the properties of eigen-values or Riemannian gradient, as shown in~\citet{richemond2023edge}, leading to a family of BYOL variants. Moreover,~\citet{guo2025representation} generalizes the technique to derive a series of non-contrastive representation learning, beyond square loss to $f$-mutual information loss. 

    \paragraph{Remark (Alternative Fix-Point Iteration):} Besides developing practical fix-point update from~\eqref{eq:eigen_fixpoint}, we can also exploit the equivalent condition 
    \begin{equation}\label{eq:eigen_fixpoint_I}
        T\rbr{x, x'} = \sqrt{\PP\rbr{x'}\PP\rbr{x}}\inner{\xi\rbr{x'}}{ \xi\rbr{x}},
    \end{equation}
    which can easily obtained by plugging the definition of $\xi$ into~\eqref{eq:spectral_match}. It induces the update via variational implicit method. Specifically, in $t$-th iteration, we have 
    \begin{align}\label{eq:eigen_l2}
        \min_{\xi}\,\,  &\int \nbr{T(x, x') - \sqrt{\PP\rbr{x}\PP\rbr{x'}}\inner{\xi_t\rbr{x'}}{\xi\rbr{x}}}^2 dx dx' \\
        &= \underbrace{\int T(x, x')^2 dxdx'}_{\text{constant}} - 2\EE_{\PP\rbr{x, x'}}\sbr{\xi_t(x')^\top \xi\rbr{x}} + \EE_{\PP\rbr{x}\PP\rbr{x'}}\sbr{\rbr{\xi_t(x')^\top \xi\rbr{x}}^2} \\
        &\propto  - 2\EE_{\PP\rbr{x, x'}}\sbr{\xi_t(x')^\top \xi\rbr{x}} + \EE_{\PP\rbr{x}}\sbr{\xi\rbr{x}^\top\underbrace{\EE_{\PP\rbr{x'}}\sbr{\xi_t(x')\xi_t(x')^\top }}_{\Lambda_t}\xi\rbr{x}},
    \end{align}
    which induces the same update rule from~\eqref{eq:minc_obj}, justifying the correctness of the derivation. Although this condition does not imply new algorithm in direct spectral factorization, we will see that this condition is more convenient for algorithm design in energy-based spectral representation. 

    \paragraph{Remark (Efficient Non-Contrastive Learning):} We observe that MINC and BYOL lead to the same spectral representation to square contrastive representation~\citep{haochen2021provable} but by non-contrastive learning. 
    Meanwhile, comparing to the non-contrastive VICReg~\citep{bardes2021vicreg} and Barlow-Twins~\citep{zbontar2021barlow}, MINC and BYOL are compatible with stochastic gradient: one can easily obtain an unbiased stochastic gradient estimator, so they are efficient with any batchsize.

\end{itemize}

\subsection{Learning Objectives for Spectral Representation}\label{subsec:spectral_learning}

Besides the existing methods for the learning of spectral representation, we here exploit different properties and views of~\eqref{eq:reparam_repr} to design several alternative learning objectives. We emphasize that these objectives are generic: they are widely applicable to energy-based, denoising, nonlinear, latent-variable, and multi-modal spectral representations, and several existing methods in the following sections are recovered as their instantiations.

\begin{itemize}
    \item {\bf Unnormalized Distribution Fitting.} Recall the fact that the factorization induces an unnormalized distribution, we can exploit classification-based NCE~\citep{gutmann2010noise} or ranking-based NCE~\citep{ma2018noise} for unnoramlized distribution fitting to implement spectral representation $\varphi\rbr{\cdot}$ learning.

    Specifically, for a given $x$, NCE augments the samples from $\PP\rbr{x'|x}$ with a group of noise samples $\PP\rbr{x'}$, and provides positive labels to samples from $\PP\rbr{x'|x}$, while negative labels to samples from noise data. Then, we have
    \begin{equation}\label{eq:nce_binary_posterior}
        \PP\rbr{y=1|x', x} = \frac{\PP\rbr{x'|x}}{\PP\rbr{x'|x} + \PP\rbr{x'}} = \frac{1}{1+\frac{\PP\rbr{x'}}{\PP\rbr{x'|x}}}.
    \end{equation}
    Notice that~\eqref{eq:nce_binary_posterior} is the solution of binary logistic regression~\citep{gutmann2012noise}, we can plug the parametrization~\eqref{eq:reparam_repr} to~\eqref{eq:nce_binary_posterior}, and apply logistic regression for spectral representation learning, which leads to
    \begin{align}\label{eq:nce_binary}
        \max_{\varphi}\,\, \EE_{\PP\rbr{x, x'}}\sbr{ \log \frac{\inner{\varphi\rbr{x'}}{\varphi\rbr{x}}}{1 + \inner{\varphi\rbr{x'}}{\varphi\rbr{x}}}} -  \EE_{\PP\rbr{x}\PP\rbr{x'}}\sbr{ \log \rbr{1 + \inner{\varphi\rbr{x'}}{\varphi\rbr{x}}}}.
    \end{align}

    Similarly, the ranking-based NCE~\citep{ma2018noise}, which generalizes binary logistic regression for multi-class classification for unnormalized distribution fitting, can also be used for spectral representation learning. For each $x$, we collect positive sample $x'\sim \PP\rbr{x'|x}$ and $k-1$ negative sample $x'\sim \PP\rbr{x'}$. Then, the posterior that $k$-th sample is positive denotes as
    \begin{equation}\label{eq:nce_multi_posterior}
        \PP(k|\cbr{x'_i}_{i=1}^k, x) = \frac{\PP\rbr{x'_k|x}\prod_{i\neq k} \PP\rbr{x_i'}}{\sum_j \PP\rbr{x'_j|x}\prod_{i\neq j} \PP\rbr{x_i'}} = \frac{\frac{\PP\rbr{x'_k|x}}{\PP\rbr{x'_k}}}{\sum_{j}\frac{\PP\rbr{x'_j|x}}{\PP\rbr{x'_j}}},
    \end{equation}
    which is the solution of multi-class logistic regression. We plug the parametrization~\eqref{eq:reparam_repr} to~\eqref{eq:nce_multi_posterior} and apply multi-class logistic regression for spectral representation learning, which leads to
    \begin{equation}\label{eq:nce_multiclass}
        \max_{\varphi}\,\,\EE_{x,\, x'_1\sim\PP\rbr{x'|x},\, x'_{2:k}\sim\PP\rbr{x'}}\sbr{\log \varphi\rbr{x}^\top \varphi\rbr{x'_1} - \log \sum_{j=1}^{k}\varphi\rbr{x}^\top \varphi\rbr{x'_j}},
    \end{equation}
    where the positive sample is indexed by $1$ without loss of generality, and the denominator sums over all the $k$ candidates, including the positive one.

    \item {\bf Density Ratio Fitting.} We observe from~\eqref{eq:spectral_match} that $\inner{\varphi\rbr{x}}{\varphi\rbr{x'}}$ play as the density ratio of ${\frac{\PP\rbr{x', x}}{\PP\rbr{x}\PP\rbr{x'}}}$. Therefore, we can exploit the density ratio estimation techniques~\citep{sugiyama2012density,nguyen2010estimating,nachum2020reinforcement} to learn the spectral representation~\citep{lu2024f}. 
    
    Specifically, consider the $f$-divergence where $f$ is convex, 
    \begin{align}
        D_f\rbr{\PP\rbr{x', x}|| \PP\rbr{x}\PP\rbr{x'}} &= \EE_{\PP\rbr{x}\PP\rbr{x'}}\sbr{f\rbr{\frac{\PP\rbr{x', x}}{\PP\rbr{x}\PP\rbr{x'}}}} \\
        &= \sup_{g}\rbr{\EE_{\PP\rbr{x', x}}\sbr{g\rbr{x, x'}} - \EE_{\PP\rbr{x}\PP\rbr{x'}}\sbr{f^*\rbr{g\rbr{x, x'}}}}, \label{eq:density_ratio_repr}
    \end{align}
    where $f^*\rbr{\cdot}$ denotes the conjugate function of $f\rbr{\cdot}$, 
    and the second equation comes from the convexity of $f(\cdot)$ and interchangeability principle~\citep{dai2017learning}.
    Meanwhile, by the conjugate property of $f\rbr{\cdot}$, the optimal solution to~\eqref{eq:density_ratio_repr} satisfies 
    \begin{equation}
        g\rbr{x,x'} =
        f'\rbr{\frac{\PP\rbr{x', x}}{\PP\rbr{x}\PP\rbr{x'}}} \Rightarrow \inner{\varphi\rbr{x}}{\varphi\rbr{x'}} = \rbr{f'}^{-1}\rbr{g\rbr{x, x'}},
    \end{equation}
    where $f'\rbr{\cdot}$ denotes the gradient of $f\rbr{\cdot}$. 
    This implies we can consider the parametrization 
    \begin{equation}\label{eq:ratio_param}
        g\rbr{x, x'} = f'\rbr{\inner{\varphi\rbr{x}}{\varphi\rbr{x'}}},
    \end{equation}
    and plug into~\eqref{eq:density_ratio_repr} for spectral representation learning~\citep{lu2024f}.
    
    For example, in $KL$-divergence, we have $f\rbr{u} =  u\log u$, $f^*\rbr{t} = \exp\rbr{t-1}$, and $f'\rbr{u} = \log u+1$. Therefore, we have 
    \[
        g\rbr{x, x'} = \log\rbr{\inner{\varphi\rbr{x}}{\varphi\rbr{x'}}} +1,
    \]
    with 
    \[
        \max_{\varphi}\,\, \EE_{\PP\rbr{x', x}}\sbr{\log\rbr{\inner{\varphi\rbr{x}}{\varphi\rbr{x'}}}} - \EE_{\PP\rbr{x}\PP\rbr{x'}}\sbr{{\inner{\varphi\rbr{x}}{\varphi\rbr{x'}}}}.
    \]
    In $\chi^2$-divergence, we have $f\rbr{u} = (u-1)^2$, $f^*\rbr{t} = \frac{t^2}{4} + t$, and $f'\rbr{u} = 2\rbr{u-1}$. 
    Therefore, we have
    \[
        g\rbr{x, x'} = 2 \inner{\varphi\rbr{x}}{\varphi\rbr{x'}} - 2,
    \]
    with 
    \[
        \max_{\varphi}\,\, 2\EE_{\PP\rbr{x', x}}\sbr{{\inner{\varphi\rbr{x}}{\varphi\rbr{x'}}}} - \EE_{\PP\rbr{x}\PP\rbr{x'}}\sbr{\rbr {\inner{\varphi\rbr{x}}{\varphi\rbr{x'}} }^2 },
    \]
    which is exactly the spectral decomposition~\eqref{eq:spectral_contrastive}.

    \item {\bf Projected Representation Learning.} Instead of fitting the density ratio, we can fit the conditional expectation of a projection of $x'$. Concretely, for any target statistic $h\rbr{x'}\in\RR^m$, \eg, raw pixels or random features of $x'$, the spectral decomposition~\eqref{eq:reparam_repr} implies
    \begin{equation}\label{eq:projected_mean}
        \EE\sbr{h\rbr{x'}|x} = \int h\rbr{x'}\PP\rbr{x'}\varphi\rbr{x'}^\top dx'\, \varphi\rbr{x} = W_h\varphi\rbr{x}, \quad W_h\in\RR^{m\times d},
    \end{equation}
    which can be fitted by the square loss
    \begin{equation}\label{eq:projected_repr}
        \min_{\varphi, W}\,\, \EE_{\PP\rbr{x, x'}}\sbr{\nbr{W\varphi\rbr{x} - h\rbr{x'}}_2^2}.
    \end{equation}
    The objective is linear in the expectation, therefore, compatible with stochastic gradient with any batch size. However, the square loss only constrains $W\varphi\rbr{x} = \EE\sbr{h\rbr{x'}|x}$, which raises two issues. First, $\varphi\rbr{x}$ is identifiable from $\EE\sbr{h\rbr{x'}|x}$ only if $W_h$ is of full column rank $d$, which requires $m\ge d$ and a sufficiently rich $h$. Second, the learned features may be redundant, \ie, linearly dependent. To avoid the redundancy, we can impose the orthonormal constraint $W^\top W = \Ib$ on the parameter $W$, rather than on the expectation $\EE\sbr{\varphi\rbr{x}\varphi\rbr{x}^\top}$ as in Barlow-Twins and VICReg, which does not introduce bias into the stochastic gradient. With this constraint,~\eqref{eq:projected_repr} reduces to the reduced-rank regression, whose solution $\varphi\rbr{x} = W^\top\EE\sbr{h\rbr{x'}|x}$ spans the top-$d$ singular subspace of the conditional expectation operator $x\mapsto\EE\sbr{h\rbr{x'}|x}$, therefore, is non-redundant as long as the operator has rank at least $d$, as shown in~\appref{app:projected}. 
    
    The choice of $h$ connects several methods: its asymmetric counterpart with the masked input $\xtil$ and the pixel target $h\rbr{x} = x$ recovers MAE in~\secref{subsec:nonlinear_spectral_learning}; $h\rbr{x'} = \varphi_t\rbr{x'}$, \ie, the representation from the previous iteration, recovers the power iteration in BYOL and MINC in~\secref{subsec:spectral_ssl}, and with a projector upon $\varphi\rbr{x}$, SimSiam and JEPA in~\secref{subsec:nonlinear_spectral_learning}; and random features of $x'$ lead to a new algorithm.

    \item {\bf Unbiased VICReg.} The penalty in~\eqref{eq:vicreg_II} is a polynomial of the expectation, \ie, 
    \begin{equation}
        \nbr{\EE\sbr{\xi\rbr{x}\xi\rbr{x}^\top} - \Ib}_F^2 = \EE_{x, \bar x}\sbr{\rbr{\xi\rbr{x}^\top\xi\rbr{\bar x}}^2} - 2\EE\sbr{\nbr{\xi\rbr{x}}^2} + d,
    \end{equation}
    where $x$ and $\bar x$ are independent samples from $\PP\rbr{x}$. Therefore, instead of the biased plug-in estimator~\eqref{eq:vicreg_bias}, we can exploit the U-statistic, \ie, using only distinct pairs in the batch $\cbr{\rbr{x_i, x'_i}}_{i=1}^B$, which leads to the unbiased estimator of~\eqref{eq:vicreg_II},
    \begin{equation}\label{eq:vicreg_u}
        -\frac{2}{B}\sum_{i=1}^B\xi\rbr{x_i}^\top\xi\rbr{x'_i} + \lambda\rbr{\frac{1}{B\rbr{B-1}}\sum_{i\neq j}\rbr{\xi\rbr{x_i}^\top\xi\rbr{x_j}}^2 - \frac{2}{B}\sum_{i=1}^B\nbr{\xi\rbr{x_i}}^2 + d}.
    \end{equation}
    Its expectation equals~\eqref{eq:vicreg_II} exactly, and so does its gradient, for any batch size $B\ge 2$. This is the same as the estimation of the square contrastive loss~\eqref{eq:spectral_contrastive} with distinct negative pairs. To the best of our knowledge, the existing implementations of VICReg and its variants exploit the plug-in estimator, and the U-statistic estimator~\eqref{eq:vicreg_u} has not been exploited in the literature.

    \item {\bf Generalized Variational Power Iteration.} In BYOL and MINC, we implement the variational fixed-point iteration through (weighted) square loss. We can exploit alternative losses for implementing the update. We consider the condition~\eqref{eq:sufficient_repr} or~\eqref{eq:reparam_repr} for the fixed-point iteration. Concretely, at $t$-th iteration, we are updating $\psi$ and $A$ by matching
    \begin{equation}
        \PP\rbr{x'|x} = \PP\rbr{x'}\inner{\psi_t\rbr{x'}}{A\psi_{t+1}\rbr{x}}.
    \end{equation}
    To obtain a tractable objective, we can exploit either NCE or variational $KL$-divergence we proposed above, which leads to 
    \begin{align}\label{eq:nce_power}
        \max_{\psi, A}\,\,\EE_{x,\, x'_1\sim\PP\rbr{x'|x},\, x'_{2:k}\sim\PP\rbr{x'}}\sbr{\log \rbr{A\psi\rbr{x}}^\top \psi_t\rbr{x'_1} - \log \sum_{j=1}^{k}\rbr{A\psi\rbr{x}}^\top \psi_t\rbr{x'_j}},
    \end{align}
    or
    \begin{align}\label{eq:density_ratio_power}
        \max_{\psi, A}\,\, \EE_{\PP\rbr{x', x}}\sbr{\log\rbr{\rbr{A\psi\rbr{x}}^\top \psi_t\rbr{x'}}} - \EE_{\PP\rbr{x}\PP\rbr{x'}}\sbr{\rbr{A\psi\rbr{x}}^\top \psi_t\rbr{x'}}.
    \end{align}
    We emphasize that besides the $KL$-divergence for matching, other variational $f$-divergence in~\eqref{eq:density_ratio_repr} can also be applied for power iteration, which we omit here. Note that the $\log$ in~\eqref{eq:nce_power} and~\eqref{eq:density_ratio_power} requires $\rbr{A\psi\rbr{x}}^\top\psi_t\rbr{x'} > 0$.
\end{itemize}

These objectives share the same target spectral representation but differ in their optimization behaviors, \eg, the dependence on the batch size, which we discuss in~\secref{sec:objective_discussion}.

\paragraph{Remark (Spectral Decomposition and Mutual Information):} 
If we select $f\rbr{u} = u\log u$, we recover the $KL$-divergence between $\PP\rbr{x, x'}$ and $\PP\rbr{x}\PP\rbr{x'}$, which is known as mutual information, \ie, 
\[
D_{KL}(\PP\rbr{x, x'}|| \PP\rbr{x}\PP\rbr{x'}) = I\rbr{X, X'}.
\]
There have been several works casting the NCE estimators~\eqref{eq:nce_binary} and~\eqref{eq:nce_multiclass} as from the Donsker-Varadhan lower bound of mutual information~\citep{hjelm2018learning,tschannen2019mutual}. Here we establish the connection between spectral decomposition and mutual information estimation, which echoes the observation in~\citet{tschannen2019mutual} that the success of these methods relies on the structure of the decomposition $\inner{\varphi\rbr{x}}{\varphi\rbr{x'}}$, rather than the mutual information itself.

\paragraph{Remark (Different from SimCLR~\citep{chen2020simple}):} We emphasize although SimCLR~\citep{chen2020simple} is also derived from ranking-based NCE, it is different from~\eqref{eq:nce_binary} and~\eqref{eq:nce_multiclass} in the sense that no $\exp\rbr{\cdot}$ operation appears in~\eqref{eq:nce_binary} and~\eqref{eq:nce_multiclass} (which thus require positive inner products $\inner{\varphi\rbr{x}}{\varphi\rbr{x'}} > 0$). We will discuss the $\exp\rbr{\cdot}$ effect in energy-based spectral representation in~\secref{sec:energy_spectral}, which naturally ensures positivity and induces significantly different properties.

\section{Energy-based Spectral Representation}\label{sec:energy_spectral}
In~\eqref{eq:reparam_repr}, we consider the linear factorization of $\PP\rbr{x'|x}$. However, one can also consider the energy-based representation~\citep{zhang2023energy, shribak2024diffusion} for spectral representation in an implicit way. 

Specifically, we consider the EBM parametrization 
\begin{equation}\label{eq:factor_ebm}
    \PP\rbr{x'|x} = \PP\rbr{x'}\exp\rbr{\upsilon\rbr{x'}^\top \upsilon\rbr{x}}.
\end{equation}
which can be rewritten with the quadratic potential function, \ie, 
\begin{equation}\label{eq:quadratic_ebm}
\textstyle
     \PP(x'|x) =  \PP\rbr{x'} \exp\rbr{{\nbr{\upsilon\rbr{x'}}^2}/{2}}\exp\rbr{-{\nbr{\upsilon\rbr{x'} - \upsilon\rbr{x}}^2}/{2}}\exp\rbr{{\nbr{\upsilon\rbr{x}}^2}/{2}}. 
\end{equation} 
The term $\exp\rbr{-\frac{\nbr{\upsilon\rbr{x'} - \upsilon\rbr{x}}^2}{2}}$ is the Gaussian kernel, for which we apply the random Fourier feature~\citep{rahimi2007random,dai2014scalable} and obtain the spectral decomposition of~\eqref{eq:factor_ebm},  
\begin{equation}\label{eq:spectral_ebm}
     \PP(x'|x) = \PP\rbr{x'} \inner{\varphi_\omega\rbr{x'}}{\varphi_\omega\rbr{x}}_{\Ncal\rbr{\omega}},
\end{equation}
where
\begin{align}\label{eq:ebm_repr}
    \textstyle
    \varphi_\omega\rbr{x} &= \sqrt{2}\cos\rbr{\omega^\top \upsilon\rbr{x} + b}\exp\rbr{{\nbr{\upsilon\rbr{x}}^2}/{2}},
\end{align}
with $\omega\sim \Ncal\rbr{0, \Ib}$ and $b\sim \text{Unif}\sbr{0, 2\pi}$, and $\inner{\cdot}{\cdot}_{\Ncal\rbr{\omega}}$ denotes the expectation over both $\omega$ and $b$. This follows from $\EE_{\omega, b}\sbr{2\cos\rbr{\omega^\top a + b}\cos\rbr{\omega^\top c + b}} = \EE_\omega\sbr{\cos\rbr{\omega^\top\rbr{a - c}}} = \exp\rbr{-\nbr{a - c}^2/2}$.
This bridges the factorized EBMs~\eqref{eq:factor_ebm} to infinite-dimensional linear factorization. 

With the energy-based representation~\eqref{eq:spectral_ebm}, we can represent the 
\[
\EE\sbr{y|x} = \inner{\varphi_\omega\rbr{x}}{W_\omega}_{\Ncal\rbr{\omega}} = \exp\rbr{{\nbr{\upsilon\rbr{x}}^2}/{2}} \int \alpha\rbr{x'} \underbrace{\exp\rbr{-{\nbr{\upsilon\rbr{x'} - \upsilon\rbr{x}}^2}/{2}}}_{k\rbr{\upsilon\rbr{x'}, \upsilon\rbr{x}}}\PP\rbr{x'}dx'.
\]
which is linear w.r.t. $\varphi_\omega\rbr{x}$, but nonlinear w.r.t. $\upsilon\rbr{x}$. In practice, we can either exploit Monte-Carlo approximation for first or second equation to obtain a tractable parametrization for $\EE\sbr{y|x}$ with parameters $W$ or $\alpha$.

\paragraph{Remark (Generalized Energy-based Spectral Representation):}
The model~\eqref{eq:factor_ebm} implicitly assumes the partition function is 1, \ie, $\int \PP\rbr{x'}\exp\rbr{\upsilon\rbr{x'}^\top \upsilon\rbr{x}}dx' = 1$, which may not hold in practice, therefore, inducing extra approximation error.  

To include the effect of the partition function, we consider
\begin{equation}\label{eq:normalized_ebm}
    \PP\rbr{x'|x} = \PP\rbr{x'}\exp\rbr{\upsilon\rbr{x'}^\top \upsilon\rbr{x} - \log Z\rbr{x}}, \quad Z\rbr{x} = \int \PP\rbr{x'}\exp\rbr{\upsilon\rbr{x'}^\top \upsilon\rbr{x}}dx'.
\end{equation}
It is clear that $Z\rbr{x}$ can be linearly represented by $\varphi_\omega\rbr{x}$, 
\begin{proof}
    We apply the Fourier decomposition to $Z\rbr{x}$, \ie, 
    \begin{align}
        Z\rbr{x} &= \EE_{\PP\rbr{x'}}\sbr{\exp\rbr{\upsilon\rbr{x'}^\top \upsilon\rbr{x}}} = \EE_{\PP\rbr{x'}}\sbr{\inner{\varphi_\omega\rbr{x}}{\varphi_\omega\rbr{x'}}_{\Ncal\rbr{\omega}}} \nonumber \\
        &=\inner{\varphi_\omega\rbr{x}}{\EE_{\PP\rbr{x'}}\sbr{\varphi_\omega\rbr{x'}}}_{\Ncal\rbr{\omega}}  =\inner{\varphi_\omega\rbr{x}}{u}_{\Ncal\rbr{\omega}}. \label{eq:z_repr}
    \end{align}
\end{proof}
With this parametrization of $\PP\rbr{x'|x}$, we have 
\begin{align}\label{eq:nonlinear_predictor}
    \EE[y|x] = \int y\PP\rbr{y|x}dy = \inner{\frac{\varphi_\omega\rbr{x}}{\inner{\varphi_\omega\rbr{x}}{u}}_{\Ncal\rbr{\omega}}}{w}_{\Ncal\rbr{\omega}}.
\end{align}

Although the obtained $\varphi_\omega(\cdot)$ is no longer able to represent the $\EE[y|x]$ linearly, it is still sufficient for prediction with a nonlinear form. Recall $w$ and $u$ share the same space of $\varphi_\omega\rbr{x}$, we consider 
\begin{align}
    u = \int \PP\rbr{x'}\varphi_\omega\rbr{x'}dx' \approx \sum_{i=1}^m \alpha_i \varphi_\omega\rbr{x'_i} \Rightarrow \inner{\varphi_\omega\rbr{x}}{u}_{\Ncal\rbr{\omega}} \approx \sum_{i=1}^m \alpha_i \exp\rbr{\upsilon\rbr{x'_i}^\top \upsilon\rbr{x}}.
\end{align}
Similarly, we have $w \approx \sum_{i=1}^m \beta_i \varphi_\omega\rbr{x'_i}$. Therefore, we have 
\begin{align}\label{eq:ebm_regressor}
    \EE[y|x] \approx \sum_{i=1}^m\frac{\beta_i \exp\rbr{\upsilon\rbr{x'_i}^\top \upsilon\rbr{x}}}{\sum_{j=1}^m \alpha_j \exp\rbr{\upsilon\rbr{x'_j}^\top \upsilon\rbr{x}}},
\end{align}
which recovers the attention mechanism, implying we can exploit attention for downstream tasks. 

On the other hand, we derive the classifier for $\PP\rbr{y|x}$. We consider the energy-based parametrization of the core set $z$ in~\secref{subsec:symmetric_spectral}, as in~\eqref{eq:normalized_ebm},
\begin{equation*}
    \PP\rbr{z|x} = \frac{1}{Z\rbr{x}}\PP\rbr{z}\exp\rbr{\upsilon(x)^\top \nu\rbr{z}},
\end{equation*}
and $\PP\rbr{y|x} = \int \PP\rbr{y|z}\PP\rbr{z|x}dz$. Similar to the diffusion spectral representation in~\secref{sec:denoising_spectral}, there are two views.
\begin{itemize}[leftmargin=*]
    \item {\bf Attention.} By the random feature property, $\exp\rbr{\upsilon(x)^\top \nu\rbr{z}} = \inner{\varphi_\omega\rbr{x}}{\psi_\omega\rbr{z}}_{\Ncal\rbr{\omega}}$ with $\psi_\omega\rbr{z} = \sqrt{2}\cos\rbr{\omega^\top \nu\rbr{z} + b}\exp\rbr{\nbr{\nu\rbr{z}}^2/2}$, therefore,
    \begin{equation}\label{eq:ebm_classifier_linear}
        \PP\rbr{y|x} = \frac{1}{Z\rbr{x}}\inner{\varphi_\omega\rbr{x}}{\underbrace{\int\PP\rbr{y|z}\PP\rbr{z} \psi_\omega\rbr{z}dz}_{\zeta_\omega\rbr{y}}}_{\Ncal\rbr{\omega}},
    \end{equation}
    which is exactly linear in $\varphi_\omega\rbr{x}$. Following the same Monte-Carlo approximation as~\eqref{eq:ebm_regressor}, the classifier is in the attention form,
    \begin{equation*}
        \PP\rbr{y|x}\approx \sum_{i=1}^m\frac{\beta_i\rbr{y}\exp\rbr{\upsilon\rbr{x'_i}^\top \upsilon\rbr{x}}}{\sum_{j=1}^m \alpha_j\exp\rbr{\upsilon\rbr{x'_j}^\top \upsilon\rbr{x}}},
    \end{equation*}
    with the weights $\beta_i\rbr{y}$ depending on the class. 

    \item {\bf Lift.} By Bayes' rule, $\PP\rbr{y|z}\PP\rbr{z} = \PP\rbr{y}\PP\rbr{z|y}$, we have
    \begin{align}\label{eq:ebm_classifier_cgf}
        \PP\rbr{y|x} &= \frac{\PP\rbr{y}}{Z\rbr{x}}\EE_{\PP\rbr{z|y}}\sbr{\exp\rbr{\upsilon\rbr{x}^\top \nu\rbr{z}}} = \frac{\PP\rbr{y}}{Z\rbr{x}}\exp\rbr{K_y\rbr{\upsilon\rbr{x}}}, \nonumber\\
        K_y\rbr{\upsilon} &= \sum_{n\ge 1}\frac{1}{n!}\kappa_n\rbr{y}\sbr{\upsilon^{\otimes n}},
    \end{align}
    where $K_y$ is the cumulant generating function of $\nu\rbr{z}$ under the posterior $\PP\rbr{z|y}$, which is bilinear in the lifted representation, \ie, $K_y\rbr{\upsilon} = \inner{\text{Lift}\rbr{\upsilon}}{\kappa\rbr{y}}$. With a projector $\texttt{nn}_\theta$ approximating the lift, we obtain the softmax classifier
    \begin{equation}\label{eq:ebm_classifier_lift}
        \PP\rbr{y|x} \propto \exp\rbr{\texttt{nn}_\theta\rbr{\upsilon\rbr{x}}^\top \eta\rbr{y}},
    \end{equation}
    where the prior $\PP\rbr{y}$ is absorbed into the inner product by appending a constant coordinate to $\texttt{nn}_\theta\rbr{\upsilon\rbr{x}}$, whose corresponding coordinate of $\eta\rbr{y}$ represents $\log\PP\rbr{y}$. The first-order truncation leads to the linear softmax classifier $\PP\rbr{y|x}\propto\exp\rbr{\upsilon\rbr{x}^\top \eta\rbr{y}}$ with $\eta\rbr{y} = \EE\sbr{\nu\rbr{z}|y}$ and the same constant coordinate, which is exact when $\PP\rbr{z|y}$ is a point mass, \ie, the label determines the core set, and is an approximation in general. Note that an MLP classification head with a final linear layer is exactly of the form~\eqref{eq:ebm_classifier_lift}, with the final layer as $\eta\rbr{y}$. For discrete $z$, the softmax latent model with $\nu\rbr{z} = W_z$ is a special case.
\end{itemize}
This also explains why a nonlinear probe upon $\upsilon\rbr{x}$ may outperform the linear probe: the projector recovers the higher-order cumulants dropped by the first-order truncation.

\subsection{Energy-based Spectral Representation Learning}\label{subsec:ebm_ssl}

With this connection between energy-based parametrization and spectral representation, we can unify a variety of existing SSL algorithms into spectral representation framework, which originally developed from different perspective. 

\begin{itemize}[leftmargin=*]

    \item {\bf Simple Framework for Contrastive Learning of Representation~(SimCLR)~\citep{chen2020simple}.}
    With the structure of $\PP\rbr{x'|x}$ in~\eqref{eq:normalized_ebm}, one can apply the ranking-based noise contrastive estimation~(NCE)~\citep{gutmann2010noise,ma2018noise} for $\upsilon\rbr{x}$, which induces the spectral representation $\varphi\rbr{x}$ through~\eqref{eq:ebm_repr}. 
    
    Specifically, following the notation in~\secref{subsec:spectral_learning}, we plug the energy-based parametrization~\eqref{eq:factor_ebm} or the normalized version~\eqref{eq:normalized_ebm} into the posterior formulation~\eqref{eq:nce_multi_posterior}, which leads to 
    \begin{equation}\label{eq:ebm_posterior}
        p_\theta\rbr{k|\cbr{x'_i}_{i=1}^k, x} = \frac{\exp\rbr{\upsilon\rbr{x'_k}^\top \upsilon\rbr{x}}}{\sum_{j=1}^{k}\exp\rbr{\upsilon\rbr{x'_j}^\top \upsilon\rbr{x}}},
    \end{equation}
    We apply multi-class logistic regression upon~\eqref{eq:ebm_posterior}, \ie, minimizing the $KL$-divergence between $\PP\rbr{k|\cbr{x'}_{i=1}^k, x} $ and $p_\theta\rbr{k|\cbr{x'}_{i=1}^k, x}$, leading to 
    \begin{equation}\label{eq:rank_nce_ebm}
        \min_{\upsilon}\,\, -\EE_{x,\, x'_1\sim\PP\rbr{x'|x},\, x'_{2:k}\sim\PP\rbr{x'}}
        \sbr{\upsilon\rbr{x'_1}^\top \upsilon\rbr{x} - \log\sum_{j=1}^{k}\exp\rbr{\upsilon\rbr{x'_j}^\top \upsilon\rbr{x}}}.
    \end{equation}
    If we restrict $\nbr{\upsilon\rbr{x}}_{2} = 1$ by considering the parametrization $\frac{\upsilon\rbr{x}}{\nbr{\upsilon\rbr{x}}_{2}}$,
    we obtain exactly the SimCLR~\citep{chen2020simple}. 
    
    Besides revealing the relationship between SimCLR and spectral representation, more importantly, we identify the correct parametrization of regressor or classifier for downstream tasks in terms of energy-based spectral representation $\upsilon\rbr{x}$, as we discussed in~\eqref{eq:ebm_regressor} and~\eqref{eq:ebm_classifier_lift}.

    \item {\bf Word2Vec~\citep{mikolov2013efficient, levy2014neural, goldberg2014word2vec}.} When we apply binary NCE with nonparametric model to text, we obtain word2vec. Concretely, we have the parametrization for word-context pair $(w, c)$ as
    \begin{equation}\label{eq:word2vec}
        \frac{\PP\rbr{w, c}}{\PP\rbr{w}\PP\rbr{c}} = \exp\rbr{\phi(w)^\top \mu(c)}.
    \end{equation}
    Apply binary NCE to~\eqref{eq:word2vec} with linear parametrization, \ie, $\phi\rbr{w} = Vw$ and $\mu\rbr{c} = Uc$ for one-hot $w$ and $c$ with the word and context embeddings $V$ and $U$, for observed $\rbr{w, c}\sim \Dcal$, we obtain word2vec,
    \begin{equation}
        \max_{V, U}\,\,  \EE_{\PP\rbr{w, c}}\sbr{\log \sigma\rbr{w^\top V^\top U c}} + \EE_{\PP\rbr{w}\PP\rbr{c}}\sbr{\log \sigma\rbr{-w^\top V^\top U c}},
    \end{equation}
    where $\sigma\rbr{t} = \frac{1}{1 + \exp\rbr{-t}}$. Moreover,~\citet{levy2014neural} show that, with $k$ negative samples, \ie, weighting the second term by $k$, the optimum satisfies $\phi\rbr{w}^\top\mu\rbr{c} = \log\frac{\PP\rbr{w, c}}{\PP\rbr{w}\PP\rbr{c}} - \log k$, \ie, word2vec implicitly factorizes the shifted logarithm of the PMI matrix, which connects it to the classic word embeddings by the explicit SVD of the logarithm of the PMI matrix.

    \item {\bf GloVe~\citep{pennington2014glove}.} With the same parametrization~\eqref{eq:word2vec}, instead of NCE, GloVe fits the logarithm of the co-occurrence counts $X_{wc}$ counted from the corpus by weighted least squares,
    \begin{equation}\label{eq:glove}
        \min_{V, U, b, \tilde b}\,\, \sum_{w, c} f\rbr{X_{wc}}\rbr{w^\top V^\top U c + b_w + \tilde b_c - \log X_{wc}}^2,
    \end{equation}
    with the weighting function $f\rbr{X} = \min\rbr{\rbr{X/X_{\max}}^{\alpha}, 1}$, which vanishes on unobserved pairs. Since $X_{wc} = N\PP\rbr{w, c}$ with $N$ the total count, $\log X_{wc} = \log\frac{\PP\rbr{w, c}}{\PP\rbr{w}\PP\rbr{c}} + \log\PP\rbr{w} + \log\PP\rbr{c} + \log N$, where the last three terms can be absorbed by the biases $b_w$ and $\tilde b_c$. Therefore, $\phi\rbr{w}^\top\mu\rbr{c}$ also fits the logarithm of the PMI, as in~\eqref{eq:word2vec}, while $f$ determines the weighting of the entries in the low-rank approximation.

    \item {\bf Momentum Contrast Representation Learning~(MoCo)~\citep{he2020momentum}.} We can also recover MoCo~\citep{he2020momentum} as a variational form of power iteration for energy-based spectral representation. As we discussed in~\secref{subsec:spectral_ssl}, the power iteration can be equivalently derived based on the fixed-point iteration of~\eqref{eq:eigen_fixpoint_I}. We exploit the idea in energy-based spectral representation. Concretely, we start with the spectral decomposition~\eqref{eq:spectral_ebm} of the energy-based model of~\eqref{eq:normalized_ebm}, \ie, 
    \begin{align}\label{eq:energy_fourier}
         \PP(x'|x) = \frac{1}{Z\rbr{x}}\PP\rbr{x'} \inner{\varphi_\omega\rbr{x'}}{\varphi_\omega\rbr{x}}_{\Ncal\rbr{\omega}}. 
    \end{align}
    We emphasize although we derive the algorithm based on~\eqref{eq:normalized_ebm}, the derivation is also applicable to~\eqref{eq:factor_ebm}, which leads to the same updates. 

    Following the fixed-point iteration idea, at $t$-th iteration, we propose to match $\PP(x'|x)$ and $\frac{1}{Z\rbr{x}}\PP\rbr{x'} \inner{\varphi_{\omega, t}\rbr{x'}}{\varphi_{\omega}\rbr{x}}_{\Ncal\rbr{\omega}}$, with $\varphi_{\omega, t}\rbr{x'}$ fixed as previous iteration solution and one $\varphi_{\omega}(x)$ updated. To obtain a tractable update, recall the fact that the RHS of~\eqref{eq:energy_fourier} is EBM, we exploit ranking-based NCE for model learning to avoid the partition function calculation, instead of $L_2$ distance in~\eqref{eq:eigen_l2}, leading to  
    \begin{equation}\label{eq:moco_nce_power}
         \max_{\varphi}\,\,\EE_{x,\, x'_1\sim\PP\rbr{x'|x},\, x'_{2:k}\sim\PP\rbr{x'}}\sbr{\log \inner{\varphi_{\omega, t}\rbr{x'_1}}{\varphi_{\omega}\rbr{x}}_{\Ncal\rbr{\omega}} - \log \sum_{j=1}^{k}\inner{\varphi_{\omega, t}\rbr{x'_j}}{\varphi_{\omega}\rbr{x}}_{\Ncal\rbr{\omega}}}.
    \end{equation}
    Based on the random feature property, \ie,
    \[
        \inner{\varphi_{\omega, t}\rbr{x'}}{\varphi_\omega\rbr{x}}_{\Ncal\rbr{\omega}} = \exp\rbr{\upsilon_t\rbr{x'}^\top \upsilon\rbr{x}},
    \]
    we achieve the equivalent objective for $\upsilon$, 
    \begin{equation}\label{eq:moco_obj}
        \upsilon_{t+1}=\argmax_{\upsilon}\,\, \EE_{x,\, x'_1\sim\PP\rbr{x'|x},\, x'_{2:k}\sim\PP\rbr{x'}}
        \sbr{\upsilon_t\rbr{x'_1}^\top \upsilon\rbr{x} - \log\sum_{j=1}^{k}\exp\rbr{\upsilon_t\rbr{x'_j}^\top \upsilon\rbr{x}}},
    \end{equation}
    which recovers the MoCo learning objective~\citep{he2020momentum}: $\upsilon\rbr{x}$ is the query encoder updated by the gradient, and $\upsilon_t$ is the key encoder for the positive and negative samples without gradient. In MoCo, $\upsilon_t$ is implemented by the exponential moving average of $\upsilon$, \ie, the momentum encoder, rather than exactly the previous iterate, and the negative samples are maintained in a queue encoded by the momentum encoder, which provides a large $k$ with a small batch size.

\end{itemize}

\subsection{More Energy-based Spectral Representation Learning}\label{subsec:more_ebm_spectral}

Besides the existing algorithms for energy-based spectral representation we explained above, we consider the instantiations of the objectives in~\secref{subsec:spectral_learning} for energy-based spectral representation. As we revealed, SimCLR is the instantiation of NCE~\eqref{eq:nce_multiclass} applied for energy-based parametrization~\eqref{eq:factor_ebm} and~\eqref{eq:normalized_ebm}, and MoCo is the instantiation of power iteration~\eqref{eq:nce_power} applied for energy-based parametrization. Here we introduce the instantiation of density ratio estimation for energy-based spectral representation. However, we emphasize that there are more variants with different divergences can be instantiated by plugging energy-based representation into corresponding objectives in~\secref{subsec:spectral_learning}.  
\begin{itemize}[leftmargin=*]
    \item {\bf Density Ratio Fitting.} We plug the energy-based representation~\eqref{eq:factor_ebm}  into~\eqref{eq:density_ratio_repr} with $KL$-divergence, leading to 
    \begin{equation}
        \max_{\upsilon}\,\, \EE_{\PP\rbr{x', x}}\sbr{{\inner{\upsilon\rbr{x}}{\upsilon\rbr{x'}}}} - \EE_{\PP\rbr{x}\PP\rbr{x'}}\sbr{\exp\rbr{\inner{\upsilon\rbr{x}}{\upsilon\rbr{x'}}}},
    \end{equation}
    which shares some similarity to ranking-based NCE~\eqref{eq:nce_multiclass}, but with the major difference lies in the negative term is expectation of $\exp$, rather than $\log$-sum-$\exp$, making the optimization much easier. 

    We also emphasize that the density ratio estimation is only applicable for~\eqref{eq:factor_ebm}. If the EBM with normalization factor introduced~\eqref{eq:normalized_ebm}, the density ratio fitting will lead to intractable optimization with the explicit calculation of $Z\rbr{x}$.
    
\end{itemize}
\section{Denoising Spectral Representation}\label{sec:denoising_spectral}

\begin{figure}[t]
\centering
\includegraphics[width=\textwidth]{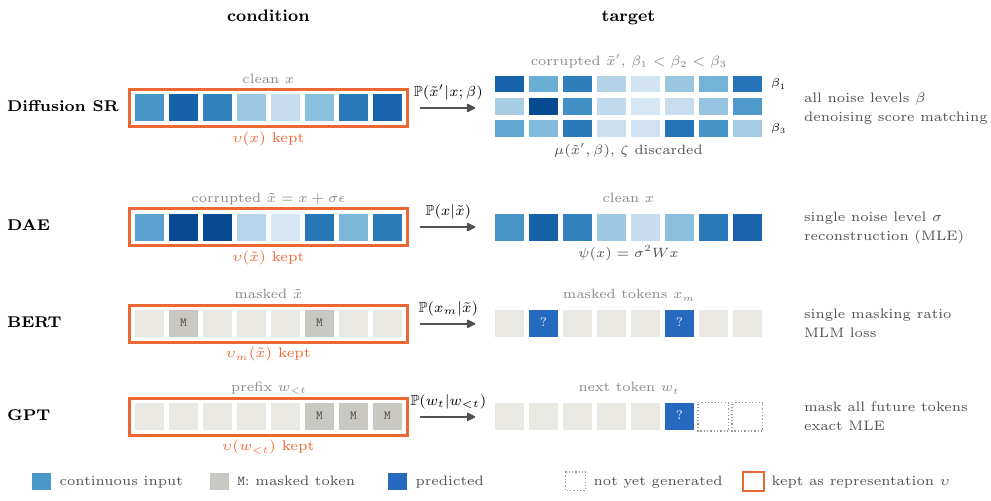}
\caption{\textbf{Condition and target of the denoising spectral representations.} All the methods share the asymmetric energy-based parametrization $\exp\rbr{\upsilon\rbr{\text{condition}}^\top\psi\rbr{\text{target}}}$ and keep the condition-side features $\upsilon$ as the representation. The diffusion spectral representation conditions on the clean $x$ and targets the corrupted $\xtil'$ at all noise levels, while the DAE conditions on the corrupted $\xtil$ and targets the clean $x$ at a single noise level, so that the two are mirror images. BERT replaces the Gaussian noise of the DAE by masking at a single ratio, and GPT masks all the future tokens.}
\label{fig:denoising}
\end{figure}

In this section, we consider the spectral representation learned through corruption, which covers a family of existing methods, including diffusion models, denoising auto-encoders, and masked and next-token prediction. Given a corruption $\PP\rbr{\xtil|x}$, two asymmetric operators in~\secref{subsec:asymmetric_spectral} arise: $\PP\rbr{\xtil'|x}$, with the clean condition and the corrupted target, and $\PP\rbr{x|\xtil}$, with the corrupted condition and the clean target. Both are learned by regressing onto the clean variable, through Tweedie's identity or maximum likelihood. All the methods in this section share the asymmetric energy-based parametrization $\exp\rbr{\upsilon\rbr{\text{condition}}^\top\psi\rbr{\text{target}}}$, with different features on the two sides, where the asymmetry comes from the corruption, as illustrated in~\figref{fig:denoising}; the multi-modal spectral representation in~\secref{sec:mm_spectral} shares the same parametrization, with the asymmetry coming from different modalities. As we will see, the corruption naturally leads to lifted and asymmetric potentials beyond the symmetric energy-based form in~\secref{sec:energy_spectral}, which inspires the nonlinear spectral representation in~\secref{sec:nonlinear_spectral}.

\subsection{Corruption and Spectral Representation}\label{subsec:corruption_spectral}

Given $x$, we consider the samples $x'\sim\PP(x'|x)$ perturbed by a corruption $\PP(\xtil'|x'; \beta)$ with noise level $\beta$, \eg, the Gaussian noise. 
We first show that the corruption preserves the linear spectral representation. For any linear spectral representation $\PP\rbr{x'|x} = \PP\rbr{x'}\inner{\varphi\rbr{x}}{\varphi\rbr{x'}}$, \eg, $\varphi_\omega$ in~\eqref{eq:spectral_ebm}, and any corruption $\PP\rbr{\xtil'|x'; \beta}$, we have
\begin{align}\label{eq:corrupted_linear}
    \PP(\xtil'|x; \beta) = \int \PP(\xtil' | x'; \beta) \PP(x'|x)dx' = \inner{\varphi\rbr{x}}{\underbrace{\int  \PP(\xtil' | x'; \beta) \PP(x')\varphi\rbr{x'} dx'}_{\xi\rbr{\xtil', \beta}}},
\end{align}
which is exactly linear in the same $\varphi\rbr{x}$ for every noise level $\beta$. Therefore, $\EE\sbr{h\rbr{\xtil'}|x}$ can be linearly represented by $\varphi\rbr{x}$ for any $h$, \ie, the spectral representation remains sufficient for the corrupted target, which becomes the asymmetric operator in~\secref{subsec:asymmetric_spectral}. 

However, the linear form~\eqref{eq:corrupted_linear} does not lead to a tractable score w.r.t.\ $\xtil'$. Applying Bayes' rule $\PP(\xtil'|x'; \beta)\PP(x') = \PP(\xtil'; \beta)\PP(x'|\xtil'; \beta)$, the corrupted pointwise mutual information is the posterior average of the original one, 
\begin{equation}\label{eq:corrupted_pmi}
    \frac{\PP(\xtil'|x; \beta)}{\PP(\xtil'; \beta)} = \EE_{\PP(x'|\xtil'; \beta)}\sbr{\frac{\PP\rbr{x'|x}}{\PP\rbr{x'}}}.
\end{equation}
For the energy-based spectral representation~\eqref{eq:factor_ebm}, this leads to
\begin{align}\label{eq:corrupted_cgf}
    \frac{\PP(\xtil'|x; \beta)}{\PP(\xtil'; \beta)} &= \EE_{\PP(x'|\xtil'; \beta)}\sbr{\exp\rbr{\upsilon\rbr{x}^\top\upsilon\rbr{x'}}} = \exp\rbr{K_{\xtil'}\rbr{\upsilon\rbr{x}}}, \nonumber \\
    K_{\xtil'}\rbr{\upsilon\rbr{x}} &= \sum_{n\ge 1}\frac{1}{n!}\kappa_n\rbr{\xtil'}\sbr{\upsilon\rbr{x}^{\otimes n}},
\end{align}
where $K_{\xtil'}\rbr{\cdot}$ is the cumulant generating function of $\upsilon\rbr{x'}$ under the posterior $\PP(x'|\xtil'; \beta)$, and $\kappa_n\rbr{\xtil'}$ is its $n$-th cumulant. Although $K_{\xtil'}$ is nonlinear in $\upsilon\rbr{x}$, it is bilinear in the lifted representation, \ie, $K_{\xtil'}\rbr{\upsilon\rbr{x}} = \inner{\text{Lift}\rbr{\upsilon\rbr{x}}}{\kappa\rbr{\xtil'}}$ with $\text{Lift}\rbr{\upsilon} = \sbr{\upsilon, \upsilon^{\otimes 2}, \upsilon^{\otimes 3}, \ldots}$ and $\kappa\rbr{\xtil'}$ stacking the scaled cumulants. This motivates the nonlinear spectral parametrization with a projector $\texttt{nn}_\theta$ learning a finite-dimensional approximation of the lift, 
\begin{equation}\label{eq:corrupted_nonlinear}
    \PP(\xtil'|x; \beta) \propto \PP(\xtil'; \beta)\exp\rbr{\texttt{nn}_\theta\rbr{\upsilon\rbr{x}}^\top \mu(\xtil', \beta)}, 
\end{equation}
where the base measure $\PP(\xtil'; \beta)$ is absorbed into the inner product by appending a constant coordinate to $\texttt{nn}_\theta\rbr{\upsilon\rbr{x}}$ in practice. 

To learn~\eqref{eq:corrupted_nonlinear}, we exploit Tweedie's identity. 
\begin{proposition}[Tweedie's Identity~\citep{efron2011tweedie}]\label{prop:tweedie}
For arbitrary corruption $\PP\rbr{\xtil'|x'; \beta}$ and $\beta$ in $\PP\rbr{\xtil'|x; \beta}$, we have 
\begin{equation}\label{eq:generali_tweedie}
    \nabla_{\xtil'} \log \PP(\xtil'|x; \beta) = \EE_{\PP(x'|\xtil', x; \beta)}\sbr{\nabla_{\xtil'} \log \PP(\xtil' | x'; \beta)}.
\end{equation}
\end{proposition}
This can be easily verified by simple calculation, \ie, 
\begin{align}
    &\nabla_{\xtil'}\log \PP(\xtil'|x; \beta) = \frac{\nabla_{\xtil'} \PP(\xtil'|x; \beta) }{\PP(\xtil'|x; \beta) } = \frac{\nabla_{\xtil'} \int \PP(\xtil' | x'; \beta) \PP(x'|x)dx' }{\PP(\xtil'|x; \beta) } \nonumber\\
    &= \int \frac{\nabla_{\xtil'} \log \PP(\xtil' | x'; \beta)\PP(\xtil' | x'; \beta) \PP(x'|x) }{\PP(\xtil'|x; \beta)} dx' = \EE_{\PP(x'|\xtil', x; \beta)}\sbr{\nabla_{\xtil'} \log \PP(\xtil' | x'; \beta)}.
\end{align}

\subsection{Denoising Spectral Representation Learning}\label{subsec:denoising_spectral_learning}

\begin{itemize}[leftmargin=*]
    \item {\bf Diffusion Spectral Representation~\citep{shribak2024diffusion}.} With the Gaussian corruption $\PP(\xtil'|x'; \beta) = \Ncal\rbr{\sqrt{1 - \beta}x', \beta\sigma^2 I}$, the diffusion spectral representation is the special case of~\eqref{eq:corrupted_nonlinear} with the identity projector, \ie, truncating $K_{\xtil'}$ to the first order,
    \begin{equation}\label{eq:corrupted_ebm}
        \PP(\xtil'|x; \beta) \propto \PP(\xtil'; \beta)\exp\rbr{\upsilon\rbr{x}^\top \mu(\xtil', \beta)}, 
    \end{equation}
    with $\mu\rbr{\xtil', \beta}\approx\EE_{\PP(x'|\xtil'; \beta)}\sbr{\upsilon\rbr{x'}}$. The first-order truncation is exact as $\beta\rightarrow 0$, where the posterior concentrates, and the second-order truncation, \ie, $\texttt{nn}_\theta\rbr{\upsilon} = \sbr{\upsilon, \text{vec}\rbr{\upsilon\upsilon^\top}}$, is exact when $\upsilon\rbr{x'}$ is Gaussian under the posterior. 

    For Gaussian perturbation with~\eqref{eq:corrupted_nonlinear}, Tweedie's identity~\eqref{eq:generali_tweedie} is applied as
    \begin{align}\label{eq:gauss_tweedie}
         \texttt{nn}_\theta\rbr{\upsilon\rbr{x}}^\top \nabla_{\xtil'}\mu\rbr{\xtil', \beta} &= \EE_{\PP(x'|\xtil', x; \beta)}\sbr{\frac{\sqrt{1 - \beta} x' - \xtil'}{\beta\sigma^2}} \nonumber
         \\
         \Rightarrow \xtil' + \beta\sigma^2 \texttt{nn}_\theta\rbr{\upsilon\rbr{x}}^\top \nabla_{\xtil'}\mu\rbr{\xtil', \beta} & = \sqrt{1 - \beta} \EE_{\PP(x'|\xtil', x; \beta)}\sbr{x'}.
    \end{align}
    Let us introduce new reparametrization $\zeta\rbr{\xtil'; \beta} = \nabla_{\xtil'}\mu\rbr{\xtil', \beta}$, since we only need $\upsilon\rbr{x}$ for the spectral representation, we can learn $\upsilon$, $\texttt{nn}_\theta$, and $\zeta$ by matching both sides of~\eqref{eq:gauss_tweedie}, 
    \begin{equation}\label{eq:intermedia}
        \min_{\upsilon, \theta, \zeta} \,\,\EE_\beta\EE_{(x, \xtil')}\sbr{\nbr{\xtil' + \beta\sigma^2 \texttt{nn}_\theta\rbr{\upsilon\rbr{x}}^\top \zeta\rbr{\xtil', \beta}  - \sqrt{1 - \beta}\EE_{\PP\rbr{x'|\xtil', x; \beta}}\sbr{ x' }}^2}  
    \end{equation}
    which shares the same optimum of 
    \begin{equation}\label{eq:diff_learn}
         \min_{\upsilon, \theta, \zeta} \,\,\ell_{\text{diff}}\rbr{\upsilon, \theta, \zeta}\defeq \EE_\beta\EE_{(x, \xtil', x')}\sbr{\nbr{\xtil' + \beta\sigma^2 \texttt{nn}_\theta\rbr{\upsilon\rbr{x}}^\top \zeta\rbr{\xtil', \beta}  - \sqrt{1 - \beta} x'}^2}. 
    \end{equation}
    With the identity projector, we recover the diffusion loss for spectral representation learning~\citep{shribak2024diffusion}. Since the score is linear in $\texttt{nn}_\theta\rbr{\upsilon\rbr{x}}$ and only involves the Jacobian of $\mu$, which is parametrized directly by $\zeta$, the learning does not require second-order gradients.

    \item {\bf Denoising Auto-Encoders~(DAE)~\citep{vincent2008extracting,vincent2011connection}.} The DAE corrupts the input as $\xtil = x + \sigma\epsilon$ with $\epsilon\sim\Ncal\rbr{0, \Ib}$. We derive it by parametrizing the distribution of the corrupted input with a latent-variable EBM, in which the latent variable $h$ plays the role of the representation,
    \begin{equation}\label{eq:dae_harmonium}
        \PP\rbr{\xtil, h}\propto \exp\rbr{-E\rbr{\xtil, h}}, \quad E\rbr{\xtil, h} = U\rbr{\xtil} - h^\top g\rbr{\xtil} + \sum_{j=1}^{d} V_j\rbr{h_j},
    \end{equation}
    where $V_j$ is a fixed base measure for the $j$-th component. Since the energy is linear in $h$ up to the componentwise terms, which is the structure of exponential-family harmoniums~\citep{welling2004exponential}, the posterior, the representation, and the free energy are available in closed form,
    \begin{equation}\label{eq:dae_harmonium_post}
    \begin{aligned}
        &\PP\rbr{h|\xtil} = \prod_{j=1}^d \exp\rbr{h_j g_j\rbr{\xtil} - V_j\rbr{h_j} - A_j\rbr{g_j\rbr{\xtil}}}, \\
        &\upsilon_j\rbr{\xtil} = \EE\sbr{h_j|\xtil} = A_j'\rbr{g_j\rbr{\xtil}}, \qquad F\rbr{\xtil} = U\rbr{\xtil} - \sum_{j=1}^d A_j\rbr{g_j\rbr{\xtil}},
    \end{aligned}
    \end{equation}
    with $A_j\rbr{\eta} = \log\int\exp\rbr{\eta h_j - V_j\rbr{h_j}}dh_j$, and the score of $\PP\rbr{\xtil}\propto\exp\rbr{-F\rbr{\xtil}}$ is linear in the representation, $\nabla_{\xtil}\log\PP\rbr{\xtil} = -\nabla_{\xtil}U\rbr{\xtil} + \nabla_{\xtil}g\rbr{\xtil}^\top\upsilon\rbr{\xtil}$. Plugging it into the denoising score matching and multiplying by $\sigma^4$ gives
    \begin{equation}\label{eq:dae_recon}
    \begin{aligned}
        &\sigma^4\EE_{x, \xtil}\sbr{\nbr{\nabla_{\xtil}\log\PP\rbr{\xtil} - \frac{x - \xtil}{\sigma^2}}^2} = \EE_{x, \xtil}\sbr{\nbr{r\rbr{\xtil} - x}^2}, \\
        &r\rbr{\xtil} = \xtil - \sigma^2\nabla_{\xtil}U\rbr{\xtil} + \sigma^2\nabla_{\xtil}g\rbr{\xtil}^\top\upsilon\rbr{\xtil},
    \end{aligned}
    \end{equation}
    a reconstruction loss whose optimal solution is $\EE\sbr{x|\xtil}$ by Tweedie's identity. The classical DAE is the simplest case: with binary latents, $g\rbr{\xtil} = W\xtil + b$, and $U\rbr{\xtil} = \frac{1}{\sigma^2}\rbr{\frac{1}{2}\nbr{\xtil}^2 - c^\top\xtil}$, the energy~\eqref{eq:dae_harmonium} is the Gaussian-binary restricted Boltzmann machine, and the encoder and the reconstruction become
    \begin{equation}\label{eq:dae_classic}
        \upsilon\rbr{\xtil} = \text{sigmoid}\rbr{W\xtil + b}, \quad r\rbr{\xtil} = c + \sigma^2W^\top\upsilon\rbr{\xtil},
    \end{equation}
    \ie, a sigmoid encoder and a linear decoder whose weights are tied to the encoder~\citep{vincent2008extracting,vincent2011connection}; with Gaussian latents, the same construction gives a linear DAE. 

    The DAE can also be derived as the maximum likelihood estimation of a Gaussian model of the denoising posterior whose mean is the reconstruction~\eqref{eq:dae_classic},
    \begin{equation}\label{eq:dae_ebm}
        \PP\rbr{x|\xtil} = \Ncal\rbr{x;\, r\rbr{\xtil}, \Ib} \propto \exp\rbr{-\frac{1}{2}\nbr{x}^2 + c^\top x}\exp\rbr{\upsilon\rbr{\xtil}^\top \psi\rbr{x}}, \quad \psi\rbr{x} = \sigma^2 W x,
    \end{equation}
    whose negative $\log$-likelihood is the reconstruction loss
    \begin{equation}\label{eq:dae_loss}
        -\EE_{x, \xtil}\sbr{\log\PP\rbr{x|\xtil}} = \frac{1}{2}\EE_{x, \xtil}\sbr{\nbr{r\rbr{\xtil} - x}^2} + \text{const}.
    \end{equation}
    Since the reconstruction is affine in the representation, the encoder enters the energy bilinearly. For general $g$, the reconstruction~\eqref{eq:dae_recon} remains bilinear in the Jacobian $\nabla_{\xtil}g\rbr{\xtil}$ and the representation $\upsilon\rbr{\xtil}$, \ie, a decoder whose matrix depends on the input, which mirrors the Jacobian $\zeta$ in the diffusion spectral representation; while the separable form $\upsilon\rbr{\xtil}^\top\psi\rbr{x}$ in~\eqref{eq:dae_ebm} holds when $g$ is linear. Therefore, the DAE is an energy-based spectral representation of the asymmetric operator $\PP\rbr{x|\xtil}$ in~\secref{subsec:asymmetric_spectral}, with the corrupted input as the condition and an asymmetric potential. Since the target features $\psi\rbr{x}$ are linear, the partition function is Gaussian and tractable, and the model is learned by maximum likelihood. The original DAE also used masking noise, the discrete counterpart adopted by BERT below; with image patches and a nonlinear decoder, it leads to MAE in~\secref{subsec:nonlinear_spectral_learning}.

    The DAE and the diffusion spectral representation~\eqref{eq:diff_learn} are mirror images of each other, as summarized below: the noise is placed on opposite sides, and the features kept as the representation are on opposite sides.
    \begin{center}
    \small
    \resizebox{\linewidth}{!}{\begin{tabular}{lll}
    \toprule
     & Diffusion spectral representation & DAE \\
    \midrule
    Condition & clean $x$ & corrupted $\xtil$ \\
    Target & corrupted $\xtil'$ & clean $x$ \\
    Operator & $\PP\rbr{\xtil'|x; \beta}$ & $\PP\rbr{x|\xtil}$ \\
    Noise levels & all levels & single level \\
    Target features & nonlinear $\mu\rbr{\xtil', \beta}$ & linear $\psi\rbr{x} = \sigma^2 W x$ \\
    Learning & denoising score matching & single-level denoising score matching \\
    Kept & clean-side $\upsilon\rbr{x}$ & corrupted-side $\upsilon\rbr{\xtil}$ \\
    Discarded & corrupted-side Jacobian $\zeta\rbr{\xtil', \beta}$ & decoder Jacobian $\nabla_{\xtil}g\rbr{\xtil}$ ($\psi\rbr{x} = \sigma^2 W x$ for linear $g$) \\
    \bottomrule
    \end{tabular}}
    \end{center}
    Both regress onto the clean variable through Tweedie's identity, but the DAE keeps the features on the corrupted side, which correspond to the discarded $\zeta$ in the diffusion spectral representation. Therefore, the DAE is not the single-noise-level special case of the diffusion spectral representation.

    \item {\bf Masked Language Modeling~(BERT)~\citep{devlin2018bert}.} BERT can be understood as the discrete counterpart of the DAE. Given a token sequence $x = \rbr{x_1, \ldots, x_L}$ with $x_l\in\Vcal$, BERT corrupts it into $\xtil$ by replacing the tokens in a random subset $M$ with \texttt{[MASK]} or random tokens, encodes $\xtil$ by a bidirectional Transformer, and predicts each masked token from the representation $\upsilon_m\rbr{\xtil}$ at position $m\in M$,
    \begin{equation}\label{eq:bert_ebm}
        \PP\rbr{x_m|\xtil} = \frac{\exp\rbr{\upsilon_m\rbr{\xtil}^\top \psi\rbr{x_m}}}{\sum_{w\in\Vcal}\exp\rbr{\upsilon_m\rbr{\xtil}^\top \psi\rbr{w}}},
    \end{equation}
    where $\psi\rbr{w}$ is the output token embedding, which is linear in the one-hot encoding of $w$. Therefore, BERT is the energy-based spectral representation of the asymmetric operator $\PP\rbr{x|\xtil}$ with the corrupted input as the condition, the same as the DAE~\eqref{eq:dae_ebm}, with the Gaussian corruption replaced by masking, and the Gaussian target replaced by the categorical one. Since the vocabulary $\Vcal$ is finite, the partition function is tractable, and the model is learned by the masked language modeling loss, \ie, the cross-entropy over the masked positions,
    \begin{equation}\label{eq:bert_loss}
        \min_{\upsilon, \psi}\,\, -\EE_{x, \xtil}\sbr{\sum_{m\in M}\log \PP\rbr{x_m|\xtil}},
    \end{equation}
    which is the discrete diffusion loss at a single masking ratio, and plays the role of the denoising score matching in the DAE. Moreover, BERT is the fixed-ratio special case of masked discrete diffusion~\citep{austin2021structured,shi2024simplified}, whose transition mixes the absorbing state \texttt{[MASK]} with the uniform transition, matching the replacement by \texttt{[MASK]} and random tokens in BERT.

    \item {\bf Autoregressive Next-Token Prediction~(GPT)~\citep{radford2018improving, radford2019language, brown2020language}.} Next-token prediction can be understood as the causal counterpart of BERT: instead of masking a random subset, it masks all the future tokens. Pre-training large language models via next-token prediction thus also falls directly under asymmetric energy-based spectral representation learning. Let $x = w_{<t}$ denote the preceding token sequence and $y = w_t \in \Vcal$ denote the next token from a finite vocabulary $\Vcal$. Absorbing the marginal log-prior $\log \PP\rbr{y}$ into the token embedding $\psi\rbr{y}$ by appending a constant coordinate to $\upsilon\rbr{x}$ yields the standard softmax language model
    \begin{equation}\label{eq:gpt_mle}
        p_\theta\rbr{y|x} = \frac{\exp\rbr{\inner{\upsilon\rbr{x}}{\psi\rbr{y}}}}{\sum_{y'\in \Vcal}\exp\rbr{\inner{\upsilon\rbr{x}}{\psi\rbr{y'}}}},
    \end{equation}
    where $\upsilon\rbr{x}$ is the causal Transformer representation of context $x$ and $\psi\rbr{y}$ is the output token embedding. The finiteness of $\Vcal$ makes the partition function $Z\rbr{x} = \sum_{y'\in\Vcal}\exp\rbr{\inner{\upsilon\rbr{x}}{\psi\rbr{y'}}}$ tractable to compute exactly. Maximizing the autoregressive log-likelihood $\EE_{\PP\rbr{x, y}}\sbr{\log p_\theta\rbr{y|x}}$ therefore performs exact conditional MLE for the asymmetric energy-based model, extracting a spectral representation of context-continuation pointwise mutual information with unbiased mini-batch gradients.

    Moreover, since the same Transformer is shared by all contexts, we can split the context into a prompt $p$ and the subsequent tokens, \ie, $x = \rbr{p, \bar x}$. For each prompt $p$,~\eqref{eq:gpt_mle} gives $p_\theta\rbr{y|\bar x, p}\propto\exp\rbr{\inner{\upsilon\rbr{p, \bar x}}{\psi\rbr{y}}}$, \ie, the energy-based spectral representation of the operator $\PP\rbr{y|\bar x, p}$ conditioned on $p$. Therefore, the same model with different prompts represents the decompositions of different PMIs, similar to the condition $\rbr{x, m}$ in JEPA~(\appref{app:jepa_fixed_point}). Since the output token embedding $\psi$ is shared by all $p$, these decompositions are not independent: they form a joint decomposition with the shared target features $\psi$, and only the condition features $\upsilon\rbr{p, \cdot}$ vary with $p$.
\end{itemize}

\section{Nonlinear Spectral Representation}\label{sec:nonlinear_spectral}

In~\secref{sec:spec_repr}, we consider the \emph{linear sufficient} representation from the factorization perspective. With the energy-based spectral representation discussed in~\secref{sec:energy_spectral}, we observe that the nonlinear sufficient representation has a corresponding linear sufficient representation but with \emph{infinite dimensions}.  
Both the linear and nonlinear representations are sufficient, in the sense that the downstream expectation can be fully expressed based on the representation. However, the major difference lies in the downstream formulations for representing $\PP\rbr{y|x}$ and $\EE\sbr{y|x}$ as we revealed in corresponding section.

These observations imply that although both linear spectral representation and energy-based spectral representation are sufficient for downstream task, there exists a tradeoff between the downstream task parametrization and dimensionality of the sufficient representation. Concretely, with linear factorization, linear models are enough for the downstream tasks, but requires low-rank assumption on $\varphi\rbr{x}$ to represent $\PP\rbr{x'|x}$; while for nonlinear factorization, a compact representation $\upsilon\rbr{x}$ will induce infinite-dimensional linear spectral representation $\varphi_\omega\rbr{x}$ without low-rank assumption, however, it requires flexible models, \eg, neural networks and kernel machines, for downstream tasks.

The denoising spectral representation in~\secref{sec:denoising_spectral} reveals two structures beyond the symmetric energy-based form~\eqref{eq:factor_ebm}, both induced by the corruption: \emph{i)}, the cumulant lift in the diffusion spectral representation~\eqref{eq:corrupted_cgf} leads to a projector upon the representation~\eqref{eq:corrupted_nonlinear}; and \emph{ii)}, the DAE~\eqref{eq:dae_ebm}, BERT~\eqref{eq:bert_ebm}, and GPT exploit asymmetric potentials, with different features on the condition and the target sides. In this section, we generalize these structures, and further beyond, to the spectral representation of $\PP\rbr{x'|x}$ without corruption, which also justifies the usage of ``projector" in most of the representation learning algorithms.

\begin{itemize}[leftmargin=*]
    \item {\bf Projector.} Similar to the cumulant lift, where the energy-based spectral representation is linear in the lifted features of $\upsilon\rbr{x}$, the energy-based spectral representation is also linear in the random features in the kernel view. Following this equivalent linear sufficient representation, we can simply replace the $\exp\rbr{\cdot}$ with arbitrary positive-definite kernel, \ie,  
\begin{equation}\label{eq:nonlinear_repr_kernel}
    {\PP\rbr{x'|x}} = {\PP\rbr{x'}}\mathsf{k}\rbr{\upsilon\rbr{x}, \upsilon\rbr{x'}}, 
\end{equation}  
which generalizes the energy-based spectral representation~\eqref{eq:factor_ebm} and~\eqref{eq:normalized_ebm}, \eg, t-SNE~\citep{maaten2008visualizing} with the Cauchy kernel in~\secref{sec:spectral_history}. 
The parametrization~\eqref{eq:nonlinear_repr_kernel} admits a spectral decomposition $\varphi\rbr{x} = f\rbr{\upsilon\rbr{x}}$. Although $f\rbr{\cdot}$ is unknown, it still implies the nonlinear sufficient representation $\upsilon\rbr{x}$ with a corresponding linear spectral representation $f\rbr{\upsilon\rbr{x}}$. So we can parametrize the nonlinear representation $\upsilon\rbr{x}$ with the projector $f\rbr{\cdot} = \texttt{nn}_\theta\rbr{\cdot}$, leading to
\begin{equation}\label{eq:nonlinear_repr_projector}
    {\PP\rbr{x'|x}} = {\PP\rbr{x'}}\inner{\texttt{nn}_\theta\rbr{\upsilon\rbr{x}}} {\texttt{nn}_\theta\rbr{\upsilon\rbr{x'}}}, 
\end{equation}
which will be learned together by exploiting the objectives we discussed in~\secref{sec:spec_repr} and~\ref{sec:energy_spectral}. This is the shared projection head in SimCLR~\citep{chen2020simple}, MoCo~\citep{he2020momentum}, Barlow-Twins~\citep{zbontar2021barlow}, and VICReg~\citep{bardes2021vicreg}, where the loss is computed upon $\texttt{nn}_\theta\rbr{\upsilon\rbr{x}}$, while the representation $\upsilon\rbr{x}$ before the projector is used for downstream tasks. Different from the linear spectral representation in~\secref{sec:spec_repr}, where $\varphi\rbr{x}$ itself is used with a linear predictor, here $\varphi\rbr{x} = \texttt{nn}_\theta\rbr{\upsilon\rbr{x}}$ is the linear sufficient representation, and $\upsilon\rbr{x}$ is also a sufficient representation, since $\varphi\rbr{x}$ is a function of $\upsilon\rbr{x}$, but in a nonlinear way, \ie, $\EE\sbr{y|x} = w^\top\texttt{nn}_\theta\rbr{\upsilon\rbr{x}}$ requires a nonlinear predictor upon $\upsilon\rbr{x}$. Theoretically, $\varphi\rbr{x} = \texttt{nn}_\theta\rbr{\upsilon\rbr{x}}$ is still linearly sufficient, so a linear head upon the projector output is enough. In practice, however, one usually uses $\upsilon\rbr{x}$ with a new nonlinear head for the downstream task: $\upsilon\rbr{x}$ is compact in information, which avoids the potentially infinite-dimensional linear sufficient representation $\varphi\rbr{x}$, and the new head focuses on the particular downstream task, while the pre-trained projector $\texttt{nn}_\theta$ is learned to span the full space required by $\PP\rbr{x'|x}$, which is more than necessary for a particular task.

    \item {\bf Asymmetric Projector.} Similar to the asymmetric potentials in the DAE, BERT, and GPT, where the condition and the target sides have different features, we can break the symmetry in~\eqref{eq:nonlinear_repr_projector} by applying the projector to one side only, leading to
    \begin{equation}\label{eq:asym_nonlinear_repr_nn}
        {\PP\rbr{x'| x}} = {\PP\rbr{x'}}\inner{\upsilon\rbr{x'}} {\texttt{nn}_\theta\rbr{\upsilon\rbr{x}}}. 
    \end{equation}  
    This echoes the DAE~\eqref{eq:dae_ebm}, where the condition side is a deep encoder while the target side is linear, and the diffusion spectral representation~\eqref{eq:corrupted_nonlinear}, where the projector is applied to the condition side while the target side is $\mu\rbr{\xtil', \beta}$. 
    The asymmetric projector has been exploited in several existing algorithms, \eg, SimSiam~\citep{chen2021exploring}, BYOL~\citep{grill2020bootstrap}, and JEPA~\citep{Assran_2023_CVPR, bardes2024revisiting} in~\secref{subsec:nonlinear_spectral_learning}, and MAE in~\secref{subsec:nonlinear_spectral_learning} and the diffusion spectral representation in~\secref{sec:denoising_spectral} share the same form with the target-side features $\mu$. In fact, the asymmetric projector~\eqref{eq:asym_nonlinear_repr_nn} can be viewed as a specialization of multi-modal spectral representation parametrization, which will be explained in~\secref{sec:mm_spectral}.
\end{itemize}

\paragraph{Remark (Neural Product):} Instead of the product of neural networks in~\eqref{eq:nonlinear_repr_projector}, one may further consider a neural network on the pair of representations, \ie, 
\begin{equation}\label{eq:nonlinear_repr_nn}
        {\PP\rbr{x'| x}} = {\PP\rbr{x'}}\texttt{nn}_\theta\rbr{\upsilon\rbr{x'}, \upsilon\rbr{x}}, 
    \end{equation}
which includes the neural network on the product, $\texttt{nn}_\theta\rbr{\inner{\upsilon\rbr{x'}}{\upsilon\rbr{x}}}$, as a special case. In the denoising setting, this corresponds to $\PP(\xtil'|x; \beta) \propto \PP(\xtil'; \beta)\exp\rbr{\texttt{nn}_\theta\rbr{\upsilon\rbr{x}, \mu\rbr{\xtil', \beta}}}$, whose score $\nabla_{\xtil'}\mu\rbr{\xtil', \beta}^\top \partial_\mu\texttt{nn}_\theta\rbr{\upsilon\rbr{x}, \mu\rbr{\xtil', \beta}}$ involves both the value and the Jacobian of $\mu$, which cannot be replaced by an independent $\zeta$. Therefore, the denoising score matching requires the gradient of the Jacobian, \ie, second-order gradients, which is more expensive. Moreover, it loses the linear structure for downstream predictors, as discussed below. To the best of our knowledge, there is no existing representation learning algorithm with this parametrization, which we leave as an open direction.

\paragraph{Remark (Flexibility vs. Predictor Parametrization)}
As a payoff to the flexibility of nonlinear generalization in~\eqref{eq:asym_nonlinear_repr_nn} and~\eqref{eq:nonlinear_repr_nn}, we may lose the ability to get an explicit closed-form parametrization of $\PP\rbr{y|x}$ or $\EE\sbr{y|x}$. Therefore, we must use neural network parametrization upon $\upsilon\rbr{x}$ for downstream task predictors. However, the flexibility brings better modeling for $\PP\rbr{x'|x}$, leading to better generation ability, which might be more important and desirable.

\subsection{Nonlinear Spectral Representation Learning}\label{subsec:nonlinear_spectral_learning}

With the nonlinear spectral representation parametrization proposed, we can recover several existing representation learning algorithm, therefore, provide more understanding.

\begin{itemize}[leftmargin=*]

    \item {\bf Masked Autoencoders~(MAE)~\citep{he2022masked}.} MAE also fits the nonlinear spectral representation with an encoder-decoder structure. Given an image $x$, MAE masks a large portion of patches, encodes the visible patches $\xtil$ by an encoder $\upsilon\rbr{\xtil}$, and reconstructs the masked pixels by a decoder $\texttt{nn}_\theta$ with a linear output layer. Note that $\xtil$ is a perturbation of $x$, rather than the observed data. In fact, this is also the case in other SSL methods, where both views are perturbations, \ie, augmentations, of the clean image, which plays the role of the core set $z$ in~\secref{subsec:symmetric_spectral}. MAE targets the asymmetric operator $\PP\rbr{x|\xtil}$ in~\secref{subsec:asymmetric_spectral}, and its parametrization is the asymmetric version of~\eqref{eq:nonlinear_repr_projector}, \ie, 
\begin{equation}\label{eq:mae_param}
    \PP\rbr{x|\xtil} = \PP\rbr{x}\inner{\mu\rbr{x}}{\texttt{nn}_\theta\rbr{\upsilon\rbr{\xtil}}} \Rightarrow \EE\sbr{x|\xtil} = W\texttt{nn}_\theta\rbr{\upsilon\rbr{\xtil}}, \quad W = \int x\PP\rbr{x}\mu\rbr{x}^\top dx,
\end{equation}
with the encoder $\upsilon$ and the decoder $\texttt{nn}_\theta$. Fitting $\EE\sbr{x|\xtil}$ by the square loss, \ie, 
\begin{equation}\label{eq:mae_obj}
    \min_{\upsilon, \theta, W}\,\, \EE_{\PP\rbr{x, \xtil}}\sbr{\nbr{W\texttt{nn}_\theta\rbr{\upsilon\rbr{\xtil}} - x}_2^2},
\end{equation}
recovers the MAE objective, where the loss is only computed on the masked patches. Compared with JEPA~\eqref{eq:jepa_obj} discussed below, which corresponds to~\eqref{eq:asym_nonlinear_repr_nn} with the target in representation space, MAE keeps the target in pixel space. 

This view explains the downstream behaviors of MAE. For downstream tasks, only the encoder $\upsilon$ is kept and the decoder $\texttt{nn}_\theta$ is discarded, which is exactly the case in Remark~(Flexibility vs.\ Predictor Parametrization), where a nonlinear predictor upon $\upsilon$ is required. Since the decoder $\texttt{nn}_\theta$ aggregates information across tokens through attention, a linear head upon the pooled tokens cannot recover it, while a nonlinear head with attention can, which explains that the representation of MAE is less linearly separable, but fine-tuning only the last Transformer block significantly improves the accuracy~\citep{he2022masked}.
One may consider using the decoder output $\texttt{nn}_\theta\rbr{\upsilon\rbr{\xtil}}$ with a linear head instead. By~\eqref{eq:mae_param}, it is linearly sufficient for any $\EE\sbr{y|x}$, if it spans the full singular subspace of $\PP\rbr{x|\xtil}$. However, the square loss~\eqref{eq:mae_obj} only constrains the conditional mean $W\texttt{nn}_\theta\rbr{\upsilon\rbr{\xtil}} = \EE\sbr{x|\xtil}$, therefore, the decoder output is not enforced to span the full singular subspace, and is specialized to pixel reconstruction. This is consistent with the observation in~\citet{he2022masked} that a deeper decoder improves the linear probing of the encoder.

    \begin{figure}[t]
    \centering
    \includegraphics[width=\textwidth]{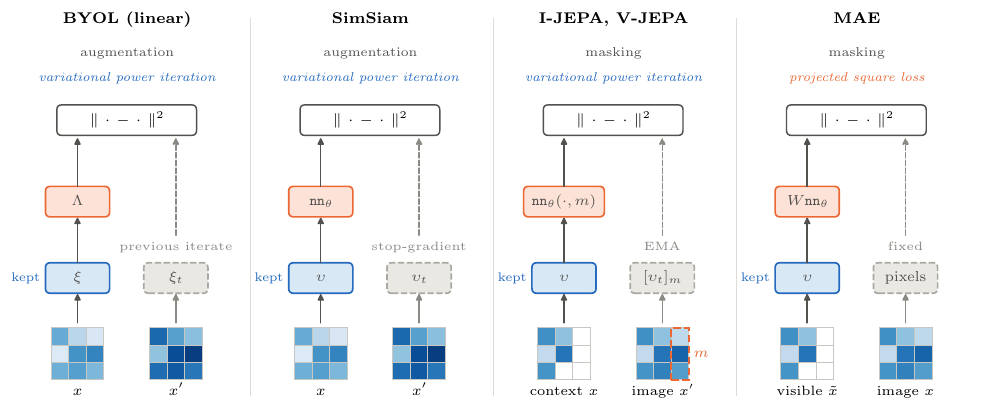}
    \caption{\textbf{Asymmetric projector in BYOL, SimSiam, JEPA, and MAE.} The projector (orange) is applied to the condition side only, and the encoder (blue) is kept as the representation. BYOL, SimSiam, and JEPA regress onto a bootstrapped target in the representation space (dashed, no gradient), \ie, the previous iterate, the stop-gradient copy, or the EMA encoder, which implements the variational power iteration; MAE regresses onto the fixed pixels, which is the projected square loss~\eqref{eq:projected_repr}.}
    \label{fig:asym}
    \end{figure}

    \item {\bf Simple Siamese Representation Learning~(SimSiam)~\citep{chen2021exploring}.} SimSiam~\citep{chen2021exploring} can be understood as nonlinear extension to BYOL~\citep{grill2020bootstrap}. Concretely, as we derived in~\eqref{eq:byol_lambda} and~\eqref{eq:byol_xi}, for $t$-th update, BYOL is iterating 
    \begin{equation}\label{eq:byol_obj}
        \min_{\xi, \Lambda}\,\,\EE_{\PP\rbr{x', x}}\sbr{\nbr{{\Lambda} \xi\rbr{x} - \xi_t\rbr{x'}}_2^2}, 
    \end{equation}
    to implement power iteration for the spectral decomposition 
    \begin{equation}\label{eq:temp_param}
        \PP\rbr{x'|x} = \PP\rbr{x'}\inner{\xi(x')}{\Lambda\xi\rbr{x}},
    \end{equation}
    with $\Lambda$ and $\xi$ defined in~\eqref{eq:orth_cond_I}.
    In the nonlinear extension~\eqref{eq:asym_nonlinear_repr_nn}, we replace $\Lambda\xi\rbr{x}$ in~\eqref{eq:temp_param} with $\texttt{nn}_\theta\rbr{\upsilon\rbr{x}}$. Plug this into~\eqref{eq:byol_obj}, we obtain the $t$-th iteration optimization
    \begin{multline}\label{eq:simsiam_intermediate}
        \min_{\upsilon, \text{nn}_\theta}\,\,\EE_{\PP\rbr{x', x}}\sbr{\nbr{ \texttt{nn}_\theta\rbr{\upsilon\rbr{x}}- \upsilon_t\rbr{x'}}_2^2} \\
        = \EE\sbr{\nbr{\texttt{nn}_\theta\rbr{\upsilon\rbr{x}}}_2^2} + \EE\sbr{\nbr{\upsilon_t\rbr{x'}}_2^2} - 2\EE_{\PP\rbr{x', x}}\sbr{\texttt{nn}_\theta\rbr{\upsilon\rbr{x}}^\top\upsilon_t\rbr{x'}}.
    \end{multline}
    The minimizer of~\eqref{eq:simsiam_intermediate} satisfies $\texttt{nn}_\theta\rbr{\upsilon_{t+1}\rbr{x}} = \EE_{\PP\rbr{x'|x}}\sbr{\upsilon_t\rbr{x'}}$, therefore, the iteration converges to the fixed point
    \begin{equation}\label{eq:nonlinear_fixpoint}
        \texttt{nn}_\theta\rbr{\upsilon\rbr{x}} = \EE_{\PP\rbr{x'|x}}\sbr{\upsilon\rbr{x'}},
    \end{equation}
    which is the nonlinear counterpart of the eigen fixed point~\eqref{eq:eigen_fixpoint}, \ie, $\EE_{\PP\rbr{x'|x}}\sbr{\xi\rbr{x'}} = \Lambda\xi\rbr{x}$, with $\Lambda\xi\rbr{x}$ replaced by $\texttt{nn}_\theta\rbr{\upsilon\rbr{x}}$.
    Additionally, with normalization condition, \ie, $\nbr{\texttt{nn}_\theta\rbr{\upsilon\rbr{x}}}_2^2 = 1$ and $\nbr{\upsilon_t\rbr{x'}}_2^2 = 1$, we obtain an equivalent objective as
    \begin{equation}
        \min_{\upsilon, \text{nn}_\theta}\,\, -\EE_{\PP\rbr{x', x}}\sbr{\texttt{nn}_\theta\rbr{\upsilon\rbr{x}}^\top\upsilon_t\rbr{x'}}.
    \end{equation}
    In practice, the normalization condition is implemented by self-normalization, we achieve the SimSiam objective, \ie, 
    \begin{equation}
        \min_{\upsilon, \text{nn}_\theta}\,\, -\EE_{\PP\rbr{x', x}}\sbr{\rbr{\frac{\texttt{nn}_\theta\rbr{\upsilon\rbr{x}}}{\nbr{\texttt{nn}_\theta\rbr{\upsilon\rbr{x}}}_2}}^\top\frac{\upsilon_t\rbr{x'}}{\nbr{\upsilon_t\rbr{x'}}_2}}.
    \end{equation}
    With the EMA target instead of the stop-gradient copy for $\upsilon_t$, the same derivation recovers the original BYOL~\citep{grill2020bootstrap} with the MLP predictor and the normalization.

    \item {\bf Joint-Embedding Predictive Architecture~(JEPA)~\citep{Assran_2023_CVPR, bardes2024revisiting}.} Image and video JEPAs~(I-JEPA~\citep{Assran_2023_CVPR} and V-JEPA~\citep{bardes2024revisiting}) likewise instantiate the asymmetric nonlinear spectral parametrization~\eqref{eq:asym_nonlinear_repr_nn} via the $L_2$ variational power iteration~\eqref{eq:simsiam_intermediate}. Specifically, given a visible context block $x$ of an image or video $x'$, and the spatial-temporal location $m$ of the masked target blocks, JEPA conditions the asymmetric predictor on $m$ and matches the representation of the exponential moving average~(EMA) target encoder at $m$, \ie, $\sbr{\upsilon_t\rbr{x'}}_m$, where the target encoder processes the entire $x'$ and the masking is applied to its output:
    \begin{equation}\label{eq:jepa_obj}
        \min_{\upsilon, \theta}\,\, \EE_{\PP\rbr{x, x', m}}\sbr{\nbr{\texttt{nn}_\theta\rbr{\upsilon\rbr{x}, m} - \sbr{\upsilon_t\rbr{x'}}_m}_2^2}.
    \end{equation}
    Correspondingly, the fixed point of~\eqref{eq:jepa_obj} is
    \begin{equation}\label{eq:jepa_fixpoint}
        \texttt{nn}_\theta\rbr{\upsilon\rbr{x}, m} = \EE_{\PP\rbr{x'|x, m}}\sbr{\sbr{\upsilon\rbr{x'}}_m},
    \end{equation}
    \ie,~\eqref{eq:nonlinear_fixpoint} with the operator $\PP\rbr{x'|x, m}$, the distribution of the entire input given the visible context and the target locations.
    V-JEPA~\citep{bardes2024revisiting} replaces the $L_2$ regression with the $L_1$ regression for training stability.
    Compared with pixel-level masked autoencoding~(MAE) discussed above, which reconstructs raw targets $x'$, JEPA performs the fixed-point iteration directly in representation space---combining BYOL's EMA target $\upsilon_t$ with SimSiam's nonlinear predictor $\texttt{nn}_\theta$ to discard unpredictable low-level observation noise while preserving predictive spectral subspaces.

    Although MAE, SimSiam, and JEPA share the same asymmetric projector parametrization, \ie, the projector $\texttt{nn}_\theta$ is applied to the condition side only, as in~\eqref{eq:mae_param} and~\eqref{eq:asym_nonlinear_repr_nn}, their learning algorithms come from different sources, as illustrated in~\figref{fig:asym} and summarized below. MAE regresses onto a fixed target in the pixel space, which is the projected representation learning~\eqref{eq:projected_repr} in~\secref{subsec:spectral_learning}; while SimSiam and JEPA regress onto a bootstrapped target in the representation space, which is the variational power iteration of BYOL~\eqref{eq:byol_lambda} and~\eqref{eq:byol_xi} in~\secref{subsec:spectral_ssl}.
    \begin{center}
    \small
    \resizebox{\linewidth}{!}{\begin{tabular}{lllll}
    \toprule
     & Parametrization & Views & Target & Learning algorithm \\
    \midrule
    MAE & Asymmetric projector & masking & pixels $x$, fixed & Projected representation learning~\eqref{eq:projected_repr} \\
    SimSiam & Asymmetric projector & augmentation & $\upsilon_t\rbr{x'}$, stop-gradient & Variational power iteration~\eqref{eq:byol_xi} \\
    JEPA & Asymmetric projector & masking & $\sbr{\upsilon_t\rbr{x'}}_m$, EMA & Variational power iteration~\eqref{eq:byol_xi} \\
    \bottomrule
    \end{tabular}}
    \end{center}
\end{itemize}

\subsection{More Nonlinear Spectral Representation Learning}\label{subsec:more_nonlinear_learning}

As we can see, SimSiam exploits the variational fixed-point iteration for nonlinear spectral representation for~\eqref{eq:asym_nonlinear_repr_nn}, which can also be applied for~\eqref{eq:nonlinear_repr_kernel} and~\eqref{eq:nonlinear_repr_nn}.

We can extend the other objectives in~\secref{subsec:spectral_learning} for nonlinear spectral representation learning. For demonstration, we instantiate one method for nonlinear parametrization, but we emphasize that other algorithms can potentially also be applied. 
\begin{itemize}[leftmargin=*]
    \item {\bf Unnormalized Distribution Fitting.} The classification-based NCE and ranking-based NCE can be directly applied for nonlinear spectral representation~\eqref{eq:nonlinear_repr_kernel}. By plugging the kernel-based nonlinear parametrization~\eqref{eq:nonlinear_repr_kernel} into~\eqref{eq:nce_binary_posterior} and ~\eqref{eq:nce_multi_posterior}, we obtain the corresponding NCE objectives, \ie, 
    \begin{align}\label{eq:nce}
        &\max_{\upsilon, \mathsf{k}}\,\, \EE_{\PP\rbr{x, x'}}\sbr{ \log \frac{\mathsf{k}\rbr{\upsilon\rbr{x}, \upsilon\rbr{x'}}}{1 + \mathsf{k}\rbr{\upsilon\rbr{x}, \upsilon\rbr{x'}}}} -  \EE_{\PP\rbr{x}\PP\rbr{x'}}\sbr{ \log \rbr{1 + \mathsf{k}\rbr{\upsilon\rbr{x}, \upsilon\rbr{x'}}}},\\
        &\max_{\upsilon, \mathsf{k}}\,\,\EE_{x,\, x'_1\sim\PP\rbr{x'|x},\, x'_{2:k}\sim\PP\rbr{x'}}\sbr{\log \mathsf{k}\rbr{\upsilon\rbr{x}, \upsilon\rbr{x'_1}} - \log \sum_{j=1}^{k}\mathsf{k}\rbr{\upsilon\rbr{x}, \upsilon\rbr{x'_j}}}.
    \end{align}
    The algorithm is also applicable to the parametrization~\eqref{eq:asym_nonlinear_repr_nn} and~\eqref{eq:nonlinear_repr_nn}. 

    \item {\bf Density Ratio Fitting.} Recall the spectral representation learning via $f$-divergence in~\secref{subsec:spectral_learning}, we can plug the nonlinear spectral representation~\eqref{eq:nonlinear_repr_kernel} to~\eqref{eq:density_ratio_repr}, which leads to the learning objective,
    \begin{align}
        \max_{\upsilon, \mathsf{k}} \,\, \EE_{\PP\rbr{x', x}}\sbr{f'\rbr{\mathsf{k}\rbr{\upsilon\rbr{x}, \upsilon\rbr{x'}}}} 
        - \EE_{\PP\rbr{x}\PP\rbr{x'}}\sbr{f^*\rbr{f'\rbr{\mathsf{k}\rbr{\upsilon\rbr{x}, \upsilon\rbr{x'}}}}}. 
    \end{align}
    The algorithm is also applicable to the parametrization~\eqref{eq:asym_nonlinear_repr_nn} and~\eqref{eq:nonlinear_repr_nn}. 

\end{itemize}

\section{Latent-Variable Spectral Representation}\label{sec:lv_spectral}

Besides the linear factorization, energy-based, denoising, and nonlinear parametrizations for spectral representation in~\secref{sec:spec_repr},~\ref{sec:energy_spectral},~\ref{sec:denoising_spectral}, and~\ref{sec:nonlinear_spectral}, respectively, in this section, we reveal the latent-variable model as an alternative implicit spectral representation parametrization. 

Recall the justification of the sufficiency of spectral representation for representation $\EE\sbr{y|x}$ for any $\PP\rbr{y|x}$, \ie, 
\begin{equation}\label{eq:lv_spectral}
    \EE\sbr{y|x} = \int y\PP\rbr{y|x}dy = \int\underbrace{\rbr{\int y\PP\rbr{y|z}dy}}_{w\rbr{z}}\PP\rbr{z|x}dz =
    \inner{w(\cdot)}{\PP\rbr{\cdot|x}}_{L_2},
\end{equation}
where the second equation comes from law of total probability. It implies that $\PP\rbr{z|x}$ can also be the sufficient representation. Meanwhile, $\PP\rbr{z|x}$ also provides us a spectral decomposition of $\PP\rbr{x'|x}$, \ie, 
\begin{align}
    \PP\rbr{x'|x} &= \int \PP\rbr{x'|z}\PP\rbr{z|x}dz = \int \frac{\PP\rbr{z|x'}\PP\rbr{x'}}{\PP\rbr{z}}\PP\rbr{z|x}dz \\
    &= \PP\rbr{x'}\inner{\sqrt{\Lambda\rbr{z}}\PP\rbr{z|x'}}{\sqrt{\Lambda\rbr{z}}\PP\rbr{z|x}}_{L_2}, \label{eq:multi_lv}
\end{align}
where $\Lambda\rbr{z} = \frac{1}{\PP\rbr{z}}$ is a diagonal operator, therefore, $\sqrt{\Lambda\rbr{z}}\PP\rbr{z|x}$ spans the same function space of $\PP\rbr{z|x}$. 

\paragraph{Remark (Tractable Predictor):} We demonstrate the $\PP\rbr{z|x}$ as the sufficient representation for linearly representing $\EE\sbr{y|x}$ in~\eqref{eq:lv_spectral}. The representation in~\eqref{eq:lv_spectral} requires an integration. For discrete latent variable $z\in \cbr{1, \ldots, k}$, this can be exactly calculated as 
\[
    \EE\sbr{y|x} = \inner{w(\cdot)}{\PP\rbr{\cdot|x}}_{L_2} = \sum_{i=1}^k w_i \PP\rbr{z_i|x}. 
\]
However, for general continuous latent variable distribution, the integration might not be tractable in general cases. Therefore, we may need to consider the approximator. One idea is following Monte-Carlo approximation. With samples $\cbr{z_i}_{i=1}^n\sim \PP\rbr{z|x}$, we have 
\begin{equation}
     \EE\sbr{y|x} = \inner{w(\cdot)}{\PP\rbr{\cdot|x}}_{L_2} \approx \frac{1}{n}\sum_{i=1}^n w\rbr{z_i}.  
\end{equation}

Particularly, if we have softmax parametrization for $\PP\rbr{z|x}$, \ie, 
\begin{equation}\label{eq:linear_latent}
        \PP\rbr{z|x} = \frac{\exp\rbr{\upsilon(x)^\top W_z + b_z}}{\sum_{z'=1}^k\exp\rbr{\upsilon(x)^\top W_{z'} + b_{z'}}}. 
    \end{equation}
which is an EBM. Based on the argument in~\secref{sec:energy_spectral}, the classifier for $\PP\rbr{y|x}$ is the softmax with a projector~\eqref{eq:ebm_classifier_lift}, whose first-order truncation is the linear softmax classifier
\[
    \PP\rbr{y|x} = \frac{\exp\rbr{\upsilon\rbr{x}^\top W_y + b_y}}{\sum_{y'}\exp\rbr{\upsilon\rbr{x}^\top W_{y'} + b_{y'}}}, 
\]
which is exact when the label is the latent variable itself. 

This implies for classification problem, we can consider softmax for $\PP\rbr{y|x}$; while for regression problem, we can consider the Monte-Carlo approximation from EBM~\eqref{eq:ebm_regressor}. 

\paragraph{Remark (Expressiveness):}
In fact, the latent variable representation $\PP\rbr{z|x}$ is a special case of linear sufficient spectral representation with additional distribution constraint, \ie, $\PP\rbr{z|x}\ge 0$ and $\int \PP\rbr{z|x}dz = 1$. This connection can be revealed more clearly if the latent variable is discrete. Concretely, if $z\in \cbr{1,\ldots, d}$, then, $\PP\rbr{\cdot|x}\in \RR^d$ is a $d$-dimensional vector in simplex. Such constrained simplex representation limits the model expressiveness, as demonstrated in~\citep[Proposition 2]{agarwal2020flambe}: there exists $\PP\rbr{x'|x}$ has finite rank $d^2$, but requires at least $2^{\Omega\rbr{d}}$ finite latent variable $z$~\citep{yannakakis1988expressing,rothvoss2017matching}, showing the original spectral representation can be more compact with fewer dimensions.  

However, on the other hand, the density constraint also brings the benefits in terms of modeling and learning that we can exploit latent variable model for representation, which can be infinite-dimensional with continuous latent variable $z$ and be fitted through probabilistic model learning, which we will discuss below.

\subsection{Latent-Variable Spectral Representation Learning}\label{subsec:lv_spectral_learning}

By revealing the spectral representation interpretation of latent-variable model, we can recast a variety of existing representation learning into our spectral representation framework. 

\begin{itemize}[leftmargin=*]

    \item {\bf DeepCluster~\citep{caron2018deep}.} 
    We here discuss the connection of DeepCluster~\citep{caron2018deep} as latent-variable spectral representation learning. Specifically, we consider the softmax parametrization of $\PP\rbr{z|x}$ in~\eqref{eq:linear_latent}.  
    Then, it can be equivalently written as
    \begin{equation}\label{eq:gauss_posterior}
        \PP\rbr{z|x} \propto \exp\rbr{- \frac{1}{2\sigma^2}\nbr{Cz - \upsilon\rbr{x}}_2^2},
    \end{equation}
    with $W_z = \frac{Cz}{\sigma^2}, \,\, b_z = -\frac{1}{2\sigma^2}z^\top C^\top Cz$, $z$ as one-hot vector, \ie, $z\in \cbr{0, 1}^k$, $z^\top \one =1$, and $C\in \RR^{d\times k}$. 
    
    Given samples $x\sim \Dcal$, it is natural to consider the maximum likelihood to the latent variable model, \ie, 
    \begin{align}
        \max_{\PP\rbr{z|x}}\,\, \EE_{\Dcal}\sbr{\log \int \PP\rbr{x|z}\PP\rbr{z}dz} =& \max_{ \PP\rbr{z|x}}\max_{q\rbr{z|x}} \EE_{\Dcal}\sbr{{\EE_{q(z|x)}\sbr{\log\PP\rbr{x, z} } + H\rbr{q\rbr{z|x}}}} \label{eq:lv_repr}\\
        =&\max_{\PP\rbr{z|x}}\max_{q\rbr{z|x}} \EE_{\Dcal}\sbr{{\EE_{q(z|x)}\sbr{\log\PP\rbr{z|x} \PP\rbr{x} } + H\rbr{q\rbr{z|x}}}}, \label{eq:lv_repr_I}
    \end{align}
    where the first equation comes from the Jensen's inequality, and the second equation comes from definition of joint probability. 
    
    Applying EM algorithm for optimizing~\eqref{eq:lv_repr}, we recover the DeepCluster~\citep{caron2018deep}, \ie, 
    \begin{itemize}
        \item {\bf E-step.} The latent variable distribution $q(z_i|x_i)$ for each data $x_i$ is obtained by optimizing~\eqref{eq:lv_repr} with fixed $\PP\rbr{z|x}$ and nonparametric $\PP\rbr{x}$. Note that $q$ should be restricted to the hard assignment, \ie, one-hot, since otherwise the optimal $q = \PP\rbr{z|x}$ makes the bound tight for any $\PP\rbr{z|x}$. With the hard assignment, the E-step gives $z_i = \argmax_z\PP\rbr{z|x_i}$, which assigns $\upsilon\rbr{x_i}$ to the nearest center $Cz$ with $\rbr{W, b}$ tied to $C$ in~\eqref{eq:gauss_posterior}. In DeepCluster~\citep{caron2018deep}, this is approximated $k$-means, which can be understood as pushing zero variance in the mixture of Gaussians in $q\rbr{z|x}$~\citep{bishop2006pattern}, leading to
        \begin{equation}
            \min_{C} \frac{1}{n}\sum_{i=1}^n\min_{z_i\in \cbr{0, 1}^k, z_i^\top \one =1} \nbr{Cz_i - \upsilon\rbr{x_i}}_2^2. 
        \end{equation}
        
        \item {\bf M-step.} Plug the learned $\cbr{z_i}_{i=1}^n$ and the posterior parametrization of $\PP\rbr{z|x}$ in~\eqref{eq:gauss_posterior} and nonparametric $\PP\rbr{x}$ into~\eqref{eq:lv_repr_I} for representation learning, we obtain
        \begin{equation}
            \max_{\upsilon} \widehat\EE_{x_i\sim \Dcal}\sbr{\upsilon\rbr{x_i}^\top W_{z_i} + b_{z_i} - \log \sum_{z=1}^k\exp\rbr{\upsilon\rbr{x_i}^\top W_{z} + b_{z}}}.
        \end{equation}
    \end{itemize}
    We emphasize that although we reveal the connection between $C$ and $\rbr{W, b}$ in~\eqref{eq:gauss_posterior}, such relationship has not been exploited in DeepCluster, and the $C$ and $W$ are two independent variables. Moreover, with the hard assignment, the objective admits a trivial solution that assigns all the samples to one cluster, which explains the reassignment of empty clusters and the uniform sampling over pseudo-labels in DeepCluster, and motivates the equipartition constraint in SeLa below.

    \item {\bf Self-Labeling~(SeLa)~\citep{asano2019self}.} The Self-Labelling~(SeLa)~\citep{asano2019self} also seeks latent-variable spectral representation, but with different parametrization for $q\rbr{z|x}$, and thus different learning updates. Specifically, with the nonparametric parametrization of $q\rbr{\cdot|x}\in \Delta_k$, \ie, the $k$-dimensional probability simplex, where $z\in \cbr{1, \ldots, k}$ is a finite latent variable, for each sample $x\sim\Dcal$, \ie, $q\rbr{\cdot|x}$ is a $k$-dimensional non-negative vector with constraint $q\rbr{\cdot|x}^\top \one = 1$, we have
    \[
    q\rbr{z, x} = q\rbr{z|x}\PP\rbr{x} = q\rbr{z|x}\frac{1}{\abr{\Dcal}}.
    \]
    Meanwhile, we consider uniform joint distribution as prior
    $p\rbr{z, x} = \frac{1}{k\cdot\abr{\Dcal}}$, we have
    \[
    KL\rbr{q\rbr{z, x} || p\rbr{z, x}} = \widehat\EE_{x_i\sim \Dcal}\EE_{q\rbr{z|x_i}}\sbr{\log q\rbr{z|x} - \log \frac{1}{k}}.
    \]
    With these parametrization, the variational maximum likelihood of latent variable model~\eqref{eq:lv_repr_I} can be equivalently written as
    \begin{equation}\label{eq:sela_I}
         \max_{\upsilon}\max_{q\rbr{z, x}} \inner{q\rbr{z, x}}{\log \PP\rbr{z|x}} - KL\rbr{q\rbr{z, x}||p\rbr{z, x}}.
    \end{equation}
    Obviously,~\eqref{eq:sela_I} is exactly the objective of SeLa~\citep{asano2019self} if we exploit softmax posterior $\PP\rbr{z|x}$ in~\eqref{eq:linear_latent}.  
    
    Particularly, both $q\rbr{z|x}$ and $\PP\rbr{z|x}$ are determined by the model itself, and~\eqref{eq:sela_I} admits trivial solutions without further constraints: with soft $q$, the optimal $q = \PP\rbr{z|x}$ makes the objective constant for any $\PP\rbr{z|x}$, providing no learning signal for the representation; with hard $q$, the objective is maximized by predicting one cluster for all the samples regardless of the input, which collapses the representation. To avoid the trivial solution, in SeLa, the posterior $q\rbr{z|x}$ is further constrained as
    \begin{equation}
        \sum_{x\in\Dcal} q\rbr{z, x} = \frac{1}{k}, \forall z\in \cbr{1, \ldots, k}.  
    \end{equation}
    Therefore, we conclude that the original SeLa is optimizing~\eqref{eq:sela_I} with softmax parametrization of $\PP\rbr{z|x}$ and nonparametric $q\rbr{z|x}$ with additional constraints. For the optimization, SeLa follows the same \textbf{M-step} as DeepCluster~\citep{caron2018deep}, but different \textbf{E-step} for $q\rbr{z, x}$ in~\eqref{eq:sela_I} as an optimal transport problem. 

    \item {\bf DINO~\citep{caron2021emerging} and SwAV~~\citep{caron2020unsupervised}.}
    The DIstillation with NO label~(DINO)~\citep{caron2021emerging, oquab2023dinov2, simeoni2025dinov3} and Swapping Assignments between multiple Views~(SwAV)~\citep{caron2020unsupervised} can both be recast as latent-variable spectral representation. Recall the spectral property of $\PP\rbr{z|x}$,  
    \begin{equation}\label{eq:lv_stationary}
        \int \PP\rbr{z|x'}\PP\rbr{x'|x} dx' = \PP\rbr{z|x},
    \end{equation}
    which follows from the law of total probability, when $z$ is conditionally independent of $x$ given $x'$, \ie, $\PP\rbr{z|x}$ is invariant to the augmentation. In other words, $\PP\rbr{z|\cdot}$ is an eigenfunction of the augmentation operator $\rbr{Tf}\rbr{x} = \int f\rbr{x'}\PP\rbr{x'|x}dx'$ with eigenvalue $1$.
    Then, we can learn the $\PP\rbr{z|x}$, so the representation, by matching the LHS and RHS of~\eqref{eq:lv_stationary} via a variety of divergences over distributions, \eg, $f$-divergence, 
    \begin{equation}\label{eq:f_stationary}
        \min_{\PP\rbr{z|x}}\,\, \EE_{x, x'}\EE_{\PP\rbr{z|x'}}\sbr{f\rbr{\frac{\PP\rbr{z|x}}{\int \PP\rbr{z|x'}\PP\rbr{x'|x} dx'}}}.
    \end{equation}
    However, the learning objective~\eqref{eq:f_stationary} is difficult to be optimized, due to the intractable integration $\int \PP\rbr{z|x'}\PP\rbr{x'|x} dx'$ inside of a nonlinear $f\rbr{\cdot}$. 
    One can exploit power iteration trick, where we fix $\PP\rbr{z|x'}$ in LHS of~\eqref{eq:lv_stationary}, and update RHS through distribution matching~\eqref{eq:f_stationary}. Concretely, at $t$-th iteration, we update $\PP\rbr{z|x}$ through 
    \begin{equation}\label{eq:lv_power}
        \int \PP_t\rbr{z|x'}\PP\rbr{x'|x} dx' = \PP_{t+1}\rbr{z|x}.
    \end{equation}
     We take the $KL$-divergence as an example with $f\rbr{u} = u\log u$ for matching~\eqref{eq:lv_power}. Then, the $t$-th iteration update for $\PP_{t+1}\rbr{z|x}$ can be written as a variational optimization, \ie,
    \begin{equation}\label{eq:power_stationary_I}
        \max_{\PP\rbr{z|x}}\,\, \EE_{x, x'}\EE_{\PP_t\rbr{z|x'}}\sbr{\log \PP\rbr{z|x}}.
    \end{equation}
    Similarly, we can also exploit
    \begin{equation}\label{eq:power_stationary_II}
        \max_{\PP\rbr{z|x'}}\,\, \EE_{x, x'}\EE_{\PP_t\rbr{z|x}}\sbr{\log \PP\rbr{z|x'}}.
    \end{equation}
    Plug the softmax parametrized multinomial distribution of $\PP\rbr{z|x}$ in~\eqref{eq:linear_latent} to the power iteration matching of stationary latent variable model~\eqref{eq:power_stationary_I} and~\eqref{eq:power_stationary_II}, we obtain 
    \begin{multline}\label{eq:dino}
    \max_{\upsilon} \widehat\EE_{x, x'\sim \Dcal}\EE_{\PP_{t}\rbr{z|x'}}\sbr{\upsilon\rbr{x}^\top W_{z} + b_{z} - \log \sum_{z'=1}^k\exp\rbr{\upsilon\rbr{x}^\top W_{z'} + b_{z'}}} \\
    + \widehat\EE_{x, x'\sim \Dcal}\EE_{\PP_{t}\rbr{z|x}}\sbr{\upsilon\rbr{x'}^\top W_{z} + b_{z} - \log \sum_{z'=1}^k\exp\rbr{\upsilon\rbr{x'}^\top W_{z'} + b_{z'}}}.
    \end{multline}
    recovering the exact objective of DINO~\citep{caron2021emerging} and SwAV~\citep{caron2020unsupervised}. 

    The major difference between DINO and SwAV lies in the implementation of $\PP_t\rbr{z|x}$ in~\eqref{eq:dino}: In DINO, $\PP_t\rbr{z|x}$ is simply exploiting the model in previous iteration; while in SwAV exploits the optimal transport solution to~\eqref{eq:sela_I}, the same as \textbf{E-step} in SeLa~\citep{asano2019self}.

    Note that the constant $\PP\rbr{z|x} = \PP\rbr{z}$ also satisfies~\eqref{eq:lv_stationary}, as the trivial eigenfunction of $T$ with eigenvalue $1$, and the power iteration~\eqref{eq:lv_power} converges to it, leading to the collapse. DINO avoids it by centering the teacher output, which removes the constant direction, and sharpening, which prevents the uniform output; while SwAV avoids it by the equipartition constraint, the same as SeLa. This echoes the removal of the trivial constant singular pair in the classic methods in~\secref{sec:spectral_history}.

\end{itemize}

\subsection{More Latent-Variable Spectral Representation Learning}\label{subsec:more_lv_spectral}

We can also exploit the objectives in~\secref{subsec:spectral_learning} for latent-variable spectral representation, which we omit here. We mainly focus on exploiting the distribution property of latent-variable model for more learning algorithms. 

\begin{itemize}[leftmargin=*]
    \item {\bf Variational ELBO for Multi-View. } Recall in~\eqref{eq:multi_lv} that we have 
    \[
    \PP\rbr{x'|x} = \PP\rbr{x'}\int \frac{\PP\rbr{z|x'}\PP\rbr{z|x}}{\PP\rbr{z}}dz.
    \]
    Given samples $\rbr{x, x'}\sim \Dcal$, we consider the maximum likelihood, \ie, 
    \begin{align}\label{eq:lv_mv_rep}
        &\max_{\PP\rbr{z|x'}, \PP\rbr{z|x}} \EE_{x, x'\sim\Dcal}\sbr{\log {\int \frac{\PP\rbr{z|x'}\PP\rbr{z|x}}{\PP\rbr{z}}dz} + \log \PP\rbr{x'} } \\
        &\propto \max_{\PP\rbr{z|x'}, \PP\rbr{z|x}, \PP\rbr{z}}\max_{q\rbr{z|x, x'}} \EE_{x, x'\sim\Dcal}\EE_{q\rbr{z|x, x'}}\sbr{\log { \frac{\PP\rbr{z|x'}\PP\rbr{z|x}}{\PP\rbr{z}}} -\log q\rbr{z|x, x'} }.
    \end{align}
    With different parametrization and learning of $q\rbr{z|x, x'}$, we can obtain different latent-variable spectral representation learning by optimizing~\eqref{eq:lv_mv_rep}. For example, if we exploit nonparametric $q\rbr{z|x, x'}$, we obtain an optimal transport optimization for cluster assignment. With the cluster fixed, we can update $\PP\rbr{z|x'}, \PP\rbr{z|x}, \PP\rbr{z}$ by stochastic gradient descent. 

    \item {\bf Contrastive Divergence for Restricted Boltzmann Machine.} With softmax parametrization of $\PP\rbr{z|x} $ is actually the Gaussian-Softmax restricted Boltzmann machine, 
    \begin{align}\label{eq:bm_param}
        \PP\rbr{z|x} &= \frac{\exp\rbr{\upsilon(x)^\top W_z + b_z}}{\sum_{z'=1}^k\exp\rbr{\upsilon(x)^\top W_{z'} + b_{z'}}},\\
        \PP\rbr{\upsilon(x)|z} &= \Ncal\rbr{Cz, \sigma^2\Ib}, 
    \end{align}
    where $W_z = \frac{Cz}{\sigma^2}, \,\, b_z = -\frac{1}{2\sigma^2}z^\top C^\top Cz$, $z$ as one-hot vector, \ie, $z\in \cbr{0, 1}^k$, $z^\top \one =1$, and $C\in \RR^{d\times k}$. 
    The parametrization also induces the distributions
    \begin{align}
        \PP\rbr{x, z} = \frac{\exp\rbr{\upsilon(x)^\top W_z + b_z}}{\int \sum_{z}\exp\rbr{\upsilon\rbr{x}^\top W_z + b_z}dx},
    \end{align}
    with the marginal distribution
    \begin{align}
        \PP\rbr{x} = \frac{\sum_z\exp\rbr{\upsilon(x)^\top W_z + b_z}}{\int \sum_{z}\exp\rbr{\upsilon\rbr{x}^\top W_z + b_z}dx}, 
    \end{align}
    which is essentially a mixture of Gaussians w.r.t. $\upsilon\rbr{x}$. 
    Then, we have the MLE as
    \begin{equation}
        \max_{W, \upsilon}\,\, \ell\rbr{W, \upsilon} = \EE_{x}\sbr{\log \sum_z \exp\rbr{\upsilon\rbr{x}^\top W_z + b_z} - \log  \int \sum_{z}\exp\rbr{\upsilon\rbr{x}^\top W_z + b_z}dx },
    \end{equation}
    with the gradients
    \begin{align}\label{eq:bm_grad}
        \nabla_{\upsilon}\ell\rbr{W, \upsilon} &= \EE_{x}\EE_{\PP\rbr{z|x}}\sbr{\nabla\upsilon(x)^\top Wz } - \EE_{\PP\rbr{z, x}}\sbr{\nabla\upsilon(x)^\top Wz},\\
        \nabla_{W}\ell\rbr{W, \upsilon} &= \EE_{x}\EE_{\PP\rbr{z|x}}\sbr{\nabla_W \upsilon(x)^\top Wz} - \EE_{\PP\rbr{z, x}}\sbr{\nabla_W\upsilon(x)^\top Wz}.
    \end{align}
    Due to the restricted Boltzmann machine structure, we can sample $(x, z)\sim \PP\rbr{z, x}$ by blockwise Gibbs sampling, \ie, 
    \begin{itemize}
        \item {\bf Sample $z$.} $\PP\rbr{z|x}$ is a multinomial distribution, therefore, sampling $z\sim \PP\rbr{z|x}$ is relative simple.   
        \item {\bf Sample ${x}$.} Sampling $x$ is relatively difficult as we have $\PP\rbr{x|z} = \frac{\exp\rbr{\upsilon\rbr{x}^\top Wz}}{\int \exp\rbr{\upsilon\rbr{x}^\top Wz} dx}$ as a general EBM, therefore, we can exploit the Langevin sampling strategy, \ie, 
        \[
            x_{t+1} = x_t + \eta_t \nabla_x \upsilon\rbr{x}^\top Wz + \sqrt{2\eta_t} \epsilon, \quad \text{with} \quad \epsilon \sim \Ncal\rbr{0, I}.
        \]
        
        Alternatively, we propose another approximate sampling strategy by exploiting the fact that 
        \[
        \PP\rbr{\upsilon(x)|z} = \Ncal\rbr{Cz, \sigma^2I}. 
        \]
        Specifically, we first sample $y \sim  \Ncal\rbr{Cz, \sigma^2I}$. Then, we extract 
        \[
        x = \argmin_{x}\nbr{y - \upsilon\rbr{x}}^2.
        \]
        This strategy is approximate, since it ignores the change of density induced by $\upsilon$.
    \end{itemize}
    With the sampling strategy, we can approximate $\ell\rbr{W, \upsilon}$ in~\eqref{eq:bm_grad}, which is an instantiation of contrastive divergence~(CD) algorithm~\citep{hinton2002training} for representation learning.
    \item {\bf Adversarial Training. } We acknowledge the difficulty in directly optimizing~\eqref{eq:f_stationary} due to the intractable integration inside of nonlinear function $f$. However, we can exploit the Fenchel duality trick~\citep{nguyen2010estimating, dai2017learning, nachum2020reinforcement} to~\eqref{eq:f_stationary}, which leads to the adversarial spectral representation learning,  
        \begin{align}\label{eq:dual_stationary}
            \min_{\PP\rbr{z|x}}\,\, &\EE_{x, x'}\EE_{\PP\rbr{z|x'}}\sbr{f\rbr{\frac{\PP\rbr{z|x}}{\int \PP\rbr{z|x'}\PP\rbr{x'|x} dx'}}}\\
           =  \min_{\PP\rbr{z|x}} \,\, &\EE_{x, x'}\EE_{\PP\rbr{z|x'}}\sbr{\max_{g_{x, z}} \frac{\PP\rbr{z|x}}{\int \PP\rbr{z|x'}\PP\rbr{x'|x} dx'}\cdot  g_{x, z} - f^*\rbr{g_{x, z}}} \\
           = \min_{\PP\rbr{z|x}}\max_{g\rbr{x, z}}\,\, &\EE_{x\sim\Dcal}\EE_{\PP\rbr{z|x}}\sbr{g\rbr{x, z}} - \EE_{x\sim\Dcal}\EE_{\PP\rbr{z|x'}\PP\rbr{x'|x}}\sbr{f^*\rbr{g\rbr{x, z}}}.
        \end{align}
        Particularly, for continuous $z$, we can exploit the reparametrization trick for gradient estimation; while for discrete $z$, \eg, softmax parametrization~\eqref{eq:linear_latent}, we can enumerate for the expectation and gradient calculation. 
        Although~\eqref{eq:dual_stationary} is a valid estimator for $\PP\rbr{z|x}$, it is a $\min$-$\max$ saddle point optimization, which might need more extra strategies for learning stability in practice. 
\end{itemize}

\subsection{Spectral Identification of Latent Variable Models}\label{subsec:lv_spectral_mom}

The decomposition~\eqref{eq:multi_lv} is exactly the multi-view structure exploited by the spectral methods for learning latent variable models~\citep{hsu2012spectral,anandkumar2012method,anandkumar2014tensor,song2010hilbert,song2014nonparametric}, \ie, the views $x$ and $x'$ are conditionally independent given $z$. Concretely, for discrete $z\in\cbr{1, \ldots, k}$ and a feature map $h\rbr{\cdot}\in\RR^m$, \eg, one-hot encoding or kernel features, the second-order moment
\begin{equation}\label{eq:lv_m2}
    M_2 = \EE_{\PP\rbr{x, x'}}\sbr{h\rbr{x}h\rbr{x'}^\top} = \sum_{z=1}^k \PP\rbr{z}\mu_z\mu_z'^\top, \quad \mu_z = \EE\sbr{h\rbr{x}|z},\,\, \mu'_z = \EE\sbr{h\rbr{x'}|z},
\end{equation}
is of rank $k$ when $\PP\rbr{z} > 0$ and $\cbr{\mu_z}$, $\cbr{\mu'_z}$ are linearly independent, and its SVD recovers $\text{span}\cbr{\mu_z}_{z=1}^k$, which is the spectral decomposition of the PMI~\eqref{eq:multi_lv} projected onto $h$. However, the factorization of $M_2$ is only identifiable up to an invertible transformation, therefore, the parameters $\cbr{\mu_z, \PP\rbr{z}}$ cannot be recovered from $M_2$ alone. The spectral methods resolve this ambiguity with a third view $x''$, \ie, the third-order moment
\begin{equation}\label{eq:lv_m3}
    M_3 = \EE\sbr{h\rbr{x}\otimes h\rbr{x'}\otimes h\rbr{x''}} = \sum_{z=1}^k \PP\rbr{z}\mu_z\otimes\mu'_z\otimes\mu''_z,
\end{equation}
whose decomposition is unique up to the permutation of $z$ when $\cbr{\mu_z}$ are linearly independent. With universal kernels, this only requires the conditional distributions $\PP\rbr{x|z}$ to be linearly independent, which allows the number of components $k$ to exceed the dimension of $x$~\citep{song2014nonparametric}. After whitening by $M_2$, it reduces to an orthogonal tensor decomposition, solved by the tensor power method~\citep{anandkumar2014tensor}, which is the third-order counterpart of the power iteration in BYOL and DINO. For hidden Markov models,~\citet{hsu2012spectral} instead learn an observable representation of the operators from the moments of one, two and three consecutive observations, which is sufficient for computing the joint and conditional probabilities without recovering the parameters. The kernel embeddings~\citep{song2010hilbert,song2014nonparametric} further extend these methods from discrete observations with one-hot features to continuous observations with nonparametric components, corresponding to the infinite-dimensional features in~\secref{sec:energy_spectral}.

This comparison clarifies the difference between representation learning and parameter estimation. For the downstream prediction $\EE\sbr{y|x} = \inner{w\rbr{\cdot}}{\PP\rbr{\cdot|x}}_{L_2}$ in~\eqref{eq:lv_spectral}, any invertible transformation of $\PP\rbr{z|x}$ is absorbed into the linear predictor $w$. Therefore, the span from the second-order moment, \ie, the PMI of the pairs $\rbr{x, x'}$, is already sufficient, and the third view and the higher-order moments for identifiability are unnecessary, which explains why SSL only requires pairs of views. On the other hand, the spectral methods are non-iterative with global optimality and consistency guarantees, in contrast to EM in DeepCluster and SeLa in~\secref{subsec:lv_spectral_learning}. We emphasize that the global optimality comes from the simple parametrization, rather than the algorithm itself: with the fixed features $h$, the moments~\eqref{eq:lv_m2} and~\eqref{eq:lv_m3} are linear in the parameters, and the learning reduces to the (tensor) eigen-decomposition, which can be solved globally, either by SVD or by power iteration. DINO and SwAV in~\secref{subsec:lv_spectral_learning} also exploit the spectral structure, \ie, $\PP\rbr{z|\cdot}$ as the eigenfunction of the augmentation operator via the power iteration~\eqref{eq:lv_power}, but with the neural parametrization of $\PP\rbr{z|x}$, the optimization becomes nonconvex, and the global optimality is lost. The price of the simple parametrization is the limited expressiveness of the fixed features, the sensitivity of the whitening and pseudo-inverse to finite samples, and the requirement of a known $k$.

\section{Multi-Modal Spectral Representation}\label{sec:mm_spectral}

In this section, we broaden our spectral representation view for multi-modal representation. Different from~\secref{sec:spec_repr},~\ref{sec:energy_spectral},~\ref{sec:nonlinear_spectral}, and~\ref{sec:lv_spectral}, which consider different parametrizations for the symmetric target $\PP\rbr{x'|x}$, multi-modal representation learning targets the asymmetric operator $\PP\rbr{y|x}$ introduced in~\secref{subsec:asymmetric_spectral}; and different from~\secref{sec:denoising_spectral}, where the asymmetry comes from the corruption, the asymmetry here comes from different modalities, and each parametrization in~\secref{sec:spec_repr}--\ref{sec:lv_spectral} can be applied to it. Specifically, we consider the parametrization for singular value decomposition to pointwise mutual information, \ie,  
\begin{align}\label{eq:asym_svd}
    \frac{\PP\rbr{x, y}}{{\PP\rbr{x}}{\PP\rbr{y}}} = \inner{\varphi\rbr{x}}{\mu\rbr{y}},
\end{align}
which implies
\begin{align}
    \PP\rbr{x|y} = \PP\rbr{x}\inner{\varphi\rbr{x}}{\mu\rbr{y}},\quad \text{and}\quad
    \PP\rbr{y|x} = \PP\rbr{y}\inner{\varphi\rbr{x}}{\mu\rbr{y}}.     
\end{align}
As we explained in~\secref{sec:sufficient_spectral} and~\secref{subsec:regression_classification}, the representation is sufficient for representing regressor $\EE\sbr{y|x}$ in~\eqref{eq:cond_mean} and classifier. 

Similarly, we can also extend the linear multi-modal representation for energy-based (\eqref{eq:factor_ebm} and~\eqref{eq:normalized_ebm}), latent-variable (\eqref{eq:multi_lv}), and nonlinear spectral representation (\eqref{eq:nonlinear_repr_kernel}, and~\eqref{eq:nonlinear_repr_nn}), and modify the corresponding learning methods for the parametrization, respectively. We mainly focus on the energy-based multi-modal representation as a demonstration. Concretely, we consider the EBM
\begin{align}\label{eq:ebm_mm}
    \frac{\PP\rbr{x, y}}{{\PP\rbr{x}}{\PP\rbr{y}}} \propto \exp\rbr{\inner{\phi\rbr{x}}{\nu\rbr{y}}},
\end{align}
which induces the following parametrizations of the conditional and joint distributions, respectively, 
\begin{align}\label{eq:asym_cond}
    \PP\rbr{x|y} = \frac{1}{Z\rbr{y}}\PP\rbr{x}\exp\rbr{\inner{\phi\rbr{x}}{\nu\rbr{y}}},& \quad \PP\rbr{y|x} = \frac{1}{Z\rbr{x}}\PP\rbr{y}\exp\rbr{\inner{\phi\rbr{x}}{\nu\rbr{y}}},\\
    \PP\rbr{x, y} =& \,\frac{1}{Z}\,\PP\rbr{x}\PP\rbr{y}\exp\rbr{\inner{\phi\rbr{x}}{\nu\rbr{y}}},
\end{align}
with 
\begin{align}
    Z\rbr{y} = \int \PP\rbr{x}\exp\rbr{\inner{\phi\rbr{x}}{\nu\rbr{y}}} dx,& \quad  Z\rbr{x} = \int \PP\rbr{y}\exp\rbr{\inner{\phi\rbr{x}}{\nu\rbr{y}}} dy,\\
    Z=& \int \PP\rbr{x}\PP\rbr{y}\exp\rbr{\inner{\phi\rbr{x}}{\nu\rbr{y}}} dxdy.
\end{align}

\subsection{Energy-based Multi-Modal Spectral Representation Learning}\label{subsec:mm_ebm_ssl}

With the energy-based parametrization~\eqref{eq:ebm_mm}, we can recast several representative multi-modal representation learning from spectral view.

\begin{itemize}[leftmargin=*]
    \item {\bf CLIP~\citep{radford2021learning} and SigLIP~\citep{zhai2023sigmoid}.} We apply ranking-based NCE for estimating $\PP\rbr{y|x}$ and $\PP\rbr{x|y}$ in~\eqref{eq:asym_cond}, leading to 
    \begin{multline}\label{eq:clip}
        \min_{\phi, \nu}\,\, -\EE_{x,\, y_1\sim\PP\rbr{y|x},\, y_{2:k}\sim\PP\rbr{y}}\sbr{\phi\rbr{x}^\top \nu\rbr{y_1} - \log\sum_{j=1}^{k}\exp\rbr{\phi\rbr{x}^\top \nu\rbr{y_j}}}\\
        -\EE_{y,\, x_1\sim\PP\rbr{x|y},\, x_{2:k}\sim\PP\rbr{x}}\sbr{\phi\rbr{x_1}^\top \nu\rbr{y} - \log\sum_{j=1}^{k}\exp\rbr{\phi\rbr{x_j}^\top \nu\rbr{y}}},
    \end{multline}
    which is exactly the objective of Contrastive Language-Image Pre-training~(CLIP)~\citep{radford2021learning}. 

    We apply classification-based NCE for estimating $\PP\rbr{y|x}$ in~\eqref{eq:asym_cond}, leading to 
    \begin{multline}\label{eq:siglip}
        \min_{\phi, \nu}\,\, -\EE_{\PP\rbr{x, y}}\sbr{\log\frac{1}{1 + \exp\rbr{-\phi\rbr{x}^\top \nu\rbr{y}}}} - \EE_{\PP\rbr{x}\PP\rbr{y}}\sbr{\log\frac{1}{1 + \exp\rbr{\phi\rbr{x}^\top \nu\rbr{y}}}}.
    \end{multline}
    which is exactly the objective of Sigmoid Language-Image Pre-training (SigLIP)~\citep{zhai2023sigmoid}. Similarly, we can also apply  classification-based NCE for estimating $\PP\rbr{x|y}$ and $\PP\rbr{x, y}$ in~\eqref{eq:asym_cond}, which leads to the same objective to~\eqref{eq:siglip}. 

    As we can see from the difference between~\eqref{eq:clip} and~\eqref{eq:siglip}, the SigLIP is compatible with stochastic gradient with unbiased gradient estimator, while CLIP exploits the ranking-based NCE with the in-batch negative samples, whose statistical efficiency depends on the number of negative samples, therefore, a large batch size is preferred in practice, although the solution is consistent with any batch size~\citep{ma2018noise}. However, empirical study shows that CLIP works better than SigLIP, which is aligned with the theoretical justification in~\citep{ma2018noise} about the consistency condition: the solution from classification-based NCE is consistent with the model with normalization condition that $\int \PP\rbr{x}\PP\rbr{y}\exp\rbr{\phi\rbr{x}^\top \nu\rbr{y}}dx dy = 1$, $\int \PP\rbr{y}\exp\rbr{\phi\rbr{x}^\top \nu\rbr{y}} dy = 1$, and $\int \PP\rbr{x}\exp\rbr{\phi\rbr{x}^\top \nu\rbr{y}}dx = 1$. Therefore, the SigLIP empirically introduces an additional temperature parameter, \ie, $\exp\rbr{\frac{\phi\rbr{x}^\top \nu\rbr{y}}{\tau}}$, requiring heavy tuning.
    
    \item {\bf SogCLR~\citep{yuan2022provable}, AmorLIP~\citep{sun2025amorlip}, and NeuCLIP~\citep{wei2025neuclip}.} To bypass the implicit assumption in SigLIP while also reduce the large-batch size issue in CLIP, there have been a few attempts, in which SogCLR~\citep{yuan2022provable}, AmorLIP~\citep{sun2025amorlip}, and NeuCLIP~\citep{wei2025neuclip} are the representative. These algorithms can all be viewed as approximations of maximum $\log$-likelihood estimation of~\eqref{eq:asym_cond}. 
    
    The MLE of $\PP\rbr{x|y}$ and $\PP\rbr{y|x}$ is 
    \begin{equation}
        \max_{\phi, \nu}\,\, 2\EE_{\PP\rbr{x, y}}\sbr{\phi\rbr{x}^\top \nu\rbr{y}} - \EE_{\PP\rbr{x}}\sbr{\log Z\rbr{x}}- \EE_{\PP\rbr{y}}\sbr{\log Z\rbr{y}}, 
    \end{equation}
    with the same gradient as the objective
    \begin{equation}\label{eq:mle_I}
        \max_{\phi, \nu}\,\, 2\EE_{\PP\rbr{x, y}}\sbr{\phi\rbr{x}^\top \nu\rbr{y}} - \EE_{\PP\rbr{x}\PP\rbr{y}}\sbr{\frac{\exp\rbr{\phi\rbr{x}^\top \nu\rbr{y}}}{\texttt{stopgrad}\rbr{Z\rbr{x}}}}- \EE_{\PP\rbr{x}\PP\rbr{y}}\sbr{\frac{\exp\rbr{\phi\rbr{x}^\top \nu\rbr{y}}}{\texttt{stopgrad}\rbr{Z\rbr{y}}}}. 
    \end{equation}
    The major difficulty in estimating the gradients to~\eqref{eq:mle_I} is the intractability of $Z\rbr{x}$ and $Z\rbr{y}$.  These algorithms are exploiting different ways for the partition function approximation. 
    Particularly, in SogCLR~\citep{yuan2022provable}, nonparametric function is exploited for partition functions: for each sample $x$ and $y$, the corresponding $Z_{x}$ and $Z_{y}$ are scalars and estimated by moving averaging, \ie, 
    \begin{align}
    Z_{x}^t &= \rbr{1 - \eta}Z^{t-1}_{x} + \frac{\eta}{\abr{B}} \sum_{y\in B}\exp\rbr{\phi\rbr{x}^\top \nu\rbr{y}},\\
    Z_{y}^t &= \rbr{1 - \eta}Z^{t-1}_{y} + \frac{\eta}{\abr{B}} \sum_{x\in B}\exp\rbr{\phi\rbr{x}^\top \nu\rbr{y}}.    
    \end{align}
    The nonparametric parametrization of $Z\rbr{x}$ and $Z\rbr{y}$ induces easy way for updating, but $O\rbr{n}$ memory cost and additional overhead in function computing through retrieval, which is unaffordable for large datasets. AmorLIP~\citep{sun2025amorlip} bypasses this by exploiting the property that $Z\rbr{x}$ and $Z\rbr{y}$ are also representable by $\phi\rbr{x}$ and $\nu\rbr{y}$ as we proved in~\eqref{eq:z_repr}. Then,~\citet{sun2025amorlip} introduces several variational objectives to fit a simple model upon the learned representation $\phi\rbr{x}$ and $\nu\rbr{y}$ for approximating $Z\rbr{x}$ and $Z\rbr{y}$, respectively. Following the idea, NeuCLIP~\citep{wei2025neuclip} introduces another objective for fitting partition functions by exploiting Fenchel dual trick~\citep{dai2017learning,dai2019exponential, nachum2020reinforcement}.
\end{itemize}

\subsection{More Multi-Modal Spectral Representation Learning}\label{subsec:more_ebm_ssl}

In fact, besides NCE and variational MLE for energy-based multi-modal spectral representation, the objectives and algorithms we introduced in~\secref{sec:spec_repr}--\ref{sec:lv_spectral}, \eg, fixed-point iteration, density ratio estimation, variational ELBO and adversarial training, and so on, can be tailored for multi-modal spectral representation learning with asymmetric parametrization~\eqref{eq:asym_svd} and~\eqref{eq:ebm_mm}. The derivation of these variants are similar and we will not repeat. Here we elaborate the diffusion loss for learning multi-modal energy-based spectral representation~\eqref{eq:ebm_mm}. 

Following the derivation in~\secref{subsec:denoising_spectral_learning}, where we connect the diffusion and energy-based model~\citep{shribak2024diffusion}, we augment our energy-based parametrization~\eqref{eq:ebm_mm} with noise level, \ie, 
\begin{align}\label{eq:ebm_diffusion}
   {\PP\rbr{\xtil, \ytil; \beta_1, \beta_2}} \propto &\int \PP\rbr{\xtil|x, \beta_1}{{\PP\rbr{x}}\PP\rbr{\ytil|y, \beta_2}{\PP\rbr{y}}}\exp\rbr{\inner{\phi\rbr{x}}{\nu\rbr{y}}}dx dy\\
   = & \int \PP\rbr{\xtil|x, \beta_1}{{\PP\rbr{x}}\PP\rbr{\ytil|y, \beta_2}{\PP\rbr{y}}} \inner{\varphi_\omega\rbr{x}}{\mu_\omega\rbr{y}}_{\Ncal\rbr{\omega}} dxdy\\
   = & \inner{\underbrace{\int\PP\rbr{\xtil|x, \beta_1}\PP\rbr{x}\varphi_\omega\rbr{x}dx}_{\varphi_\omega\rbr{\xtil; \beta_1}}}{\underbrace{\int \PP\rbr{\ytil|y, \beta_2}{\PP\rbr{y}}\mu_\omega\rbr{y} dy}_{\mu_\omega\rbr{\ytil; \beta_2}} }_{\Ncal\rbr{\omega}},
\end{align}
which is exactly linear in the random features. Similar to the diffusion spectral representation in~\secref{sec:denoising_spectral}, we parametrize it as $\exp\rbr{\inner{\tilde\phi\rbr{\xtil; \beta_1}}{\tilde\nu\rbr{\ytil; \beta_2}}}$, which is the first-order truncation of the cumulant lift.
Then, we consider the diffusion loss to learn $\tilde\phi\rbr{\xtil; \beta_1}$ and $\tilde\nu\rbr{\ytil; \beta_2}$ based on the Tweedie's Identity for $\PP\rbr{\xtil|\ytil; \beta_1, \beta_2}$ and $\PP\rbr{\ytil|\xtil; \beta_1, \beta_2}$, which leads to the diffusion loss
\begin{align}\label{eq:diffusion_mm}
    \min_{\tilde\phi, \tilde\nu}\,\, &\EE_{x, y\sim \PP\rbr{x, y}, \xtil, \ytil, \beta_1, \beta_2}\sbr{\nbr{\xtil + \beta_1\sigma^2 \nabla_{\xtil}\tilde\phi\rbr{\xtil; \beta_1}^\top \tilde\nu\rbr{\ytil; \beta_2} - \sqrt{1 - \beta_1} x}^2}\\
    &+ \EE_{x, y\sim \PP\rbr{x, y}, \xtil, \ytil, \beta_1, \beta_2}\sbr{\nbr{\ytil + \beta_2\sigma^2 \tilde\phi\rbr{\xtil; \beta_1}^\top \nabla_{\ytil}\tilde\nu\rbr{\ytil; \beta_2} - \sqrt{1 - \beta_2} y}^2}. 
\end{align}

\paragraph{Remark (Discrete Diffusion and Masked Language Modeling):} In the diffusion loss~\eqref{eq:diffusion_mm}, we implicitly assume the gradient exists, which is not satisfied for discrete variables, \eg, the text $x$. Following BERT in~\secref{subsec:denoising_spectral_learning}, we replace the first term in~\eqref{eq:diffusion_mm} with the masked diffusion loss~\citep{shi2024simplified}, \ie, the cross-entropy over the masked tokens as in~\eqref{eq:bert_loss}, with the masked text $\xtil$ and the corrupted $\ytil$ as the condition, averaged over the masking ratios. BERT is its unimodal special case with a fixed masking ratio. Other discrete diffusion losses can likewise be exploited~\citep{austin2021structured,sun2022score}.

\section{Discussion}\label{sec:objective_discussion}

\subsection{Learning Objectives and Batch Size}\label{subsec:objective_batch}

\begin{figure}[t]
\centering
\includegraphics[width=\textwidth]{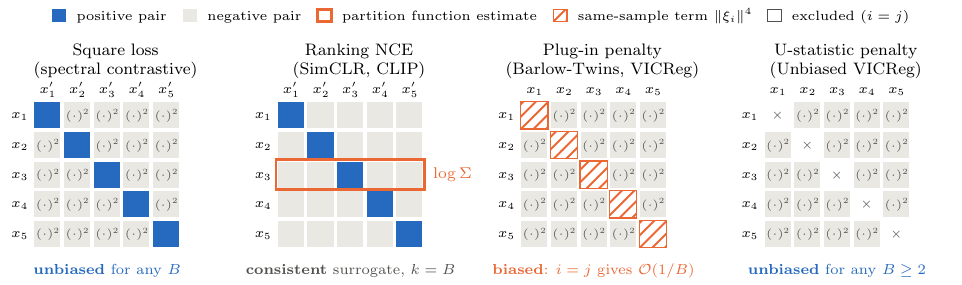}
\caption{\textbf{Mini-batch estimators of the objectives.} Each panel shows the pairs in a batch of size $B$ used by the objective. The square loss~\eqref{eq:spectral_contrastive} averages the positive pairs and the squared scores of the negative pairs, both linear in the expectations, and is unbiased. The ranking NCE~\eqref{eq:nce_multiclass} estimates the partition function by summing each row inside the $\log$, which optimizes the surrogate $J_k$ with $k=B$, consistent for any $B$. The plug-in penalty in Barlow-Twins and VICReg expands as $\frac{1}{B^2}\sum_{i,j}\rbr{\xi\rbr{x_i}^\top\xi\rbr{x_j}}^2$, whose same-sample terms $i=j$ involve the fourth moments and induce the $\Ocal\rbr{1/B}$ bias~\eqref{eq:vicreg_bias}, in addition to the batch normalization in Barlow-Twins; the U-statistic~\eqref{eq:vicreg_u} excludes them and is unbiased.}
\label{fig:batch}
\end{figure}

The objectives introduced in~\secref{subsec:spectral_learning} and instantiated in~\secref{sec:spec_repr}--\ref{sec:mm_spectral} aim for the same spectral representation, but differ significantly in optimization, particularly in the role of the mini-batch size. We distinguish three categories.
\begin{itemize}[leftmargin=*]
    \item \emph{Unbiased objectives.} If the objective is linear in expectations over the data distribution, or a polynomial of them estimated by U-statistics, \ie, with distinct samples in the batch---\eg, the square loss~\eqref{eq:spectral_contrastive}, binary NCE~\eqref{eq:nce_binary}, density ratio fitting~\eqref{eq:density_ratio_repr}, variational power iteration in BYOL and MINC~(\secref{subsec:spectral_ssl}), the denoising losses in~\secref{sec:denoising_spectral}, exact MLE, and the Unbiased VICReg~\eqref{eq:vicreg_u}---the mini-batch stochastic gradient is unbiased for any batch size, and a larger batch size only reduces gradient variance. 
    \item \emph{Consistent surrogates.} In the ranking-based NCE~\eqref{eq:nce_multiclass}, the batch size (or queue size) determines the number of negative samples $k-1$, and thus determines \emph{which} surrogate objective $J_k$ is optimized. However, for any fixed $k \ge 2$, the mini-batch gradient is unbiased w.r.t.\ $J_k$, and under a well-specified model its maximizer is statistically consistent~\citep{ma2018noise}; increasing $k$ improves asymptotic statistical efficiency towards MLE ($k\rightarrow\infty$) rather than removing an estimation bias. This explains why MoCo~\citep{he2020momentum} succeeds with a small batch size paired with a large negative queue: what matters is $k$, not the batch size itself. Moreover, under model misspecification (\eg, a finite-rank parametrization when the true PMI has higher rank), distinct $k$ generically lead to distinct optima, which converge to the MLE ($KL$) projection at the rate $\Ocal\rbr{1/k}$ as $k\rightarrow\infty$, as shown in~\appref{app:nce_misspec}; therefore, a larger $k$ also changes the learned representation, which is complementary to the efficiency analysis in~\citet{ma2018noise}.
    \item \emph{Biased objectives.} If a nonlinear function of the expectations is estimated by plugging in the batch statistics, \eg, the penalized orthogonality in Barlow-Twins~\eqref{eq:barow_twins_II} and VICReg~\eqref{eq:vicreg_I} and~\eqref{eq:vicreg_II}, the mini-batch gradient is biased even w.r.t.\ the population objective itself. Specifically, the expected mini-batch loss equals the population objective plus an $\Ocal\rbr{1/B}$ term depending on the fourth moments of the representation, as shown in~\appref{app:bt_bias} and~\eqref{eq:vicreg_bias}, rendering small-batch solutions \emph{inconsistent} and necessitating large batch sizes. The bias is inherent for the whitening, \ie, the batch normalization, in Barlow-Twins and the square root and the hinge in VICReg~\eqref{eq:vicreg_I}; while for the quadratic penalty in~\eqref{eq:vicreg_II}, the bias only comes from the plug-in estimator, and is removed by the Unbiased VICReg~\eqref{eq:vicreg_u}.
\end{itemize}
Therefore, the contrastive vs.\ non-contrastive distinction does not characterize the batch-size requirement: contrastive ranking-based NCE is consistent with any batch size (and square contrastive loss is completely unbiased), whereas non-contrastive Barlow-Twins and VICReg are biased and inconsistent with small batch sizes. From an optimization standpoint, unbiased objectives such as~\eqref{eq:spectral_contrastive}, the variational power iteration, or the Unbiased VICReg~\eqref{eq:vicreg_u} are preferable under small batches, while ranking-based NCE benefits statistically from as many negative samples as affordable. We illustrate the mini-batch estimators in~\figref{fig:batch} and summarize the objectives in~\tabref{tab:objectives}.

\begin{table}[h]
\centering
\caption{Learning objectives for spectral representation and the role of the batch size.}
\label{tab:objectives}
\small
\resizebox{\linewidth}{!}{\begin{tabular}{llll}
\toprule
Objective & Mini-batch gradient & Role of batch size & Existing instances \\
\midrule
Square loss~\eqref{eq:spectral_contrastive} & Unbiased & Variance & Spectral contrastive \\
Denoising loss (Gaussian / masked) & Unbiased & Variance & Diffusion, DAE, BERT \\
Binary NCE~\eqref{eq:nce_binary} & Unbiased & Variance & word2vec, SigLIP \\
Density ratio~\eqref{eq:density_ratio_repr} & Unbiased & Variance & \citet{lu2024f} \\
Projected square loss~\eqref{eq:projected_repr} & Unbiased & Variance & MAE \\
Variational power iteration & Unbiased & Variance & BYOL, MINC, SimSiam, JEPA, DINO \\
MLE / Variational EM (exact $Z$) & Unbiased & Variance & GPT, DeepCluster \\
Approximate MLE (estimated $Z$) & Approximately unbiased & Variance & SogCLR, AmorLIP, NeuCLIP \\
Ranking NCE~\eqref{eq:nce_multiclass} & Unbiased for fixed $k$ & Statistical efficiency & SimCLR, MoCo, CLIP \\
Penalized orthogonality (plug-in) & Biased & Bias (inconsistent) & Barlow-Twins, VICReg \\
Penalized orthogonality (U-statistic)~\eqref{eq:vicreg_u} & Unbiased & Variance & Unbiased VICReg (new) \\
\bottomrule
\end{tabular}}
\end{table}

\subsection{Parametrization and Downstream Predictors}\label{subsec:downstream_predictor}

Although all the spectral representations above are sufficient, the parametrization determines the form of the downstream predictor upon the learned representation, as summarized in~\tabref{tab:predictor}. For the linear spectral representation, the linear head is enough for regression and classification, as shown in~\secref{subsec:regression_classification}. For the energy-based spectral representation, the linear sufficient representation $\varphi_\omega\rbr{x}$ is infinite-dimensional, and the predictor upon $\upsilon\rbr{x}$ is in the attention form~\eqref{eq:ebm_regressor}, or the softmax with a projector~\eqref{eq:ebm_classifier_lift}, whose first-order truncation is the linear softmax classifier. For the nonlinear spectral representation with a projector, $\texttt{nn}_\theta\rbr{\upsilon\rbr{x}}$ is linearly sufficient, while $\upsilon\rbr{x}$ requires a nonlinear head, as discussed in Remark~(Flexibility vs.\ Predictor Parametrization). When the projector aggregates information across tokens, \eg, the decoder in MAE, the nonlinear head should also aggregate across tokens, which explains the gap between the linear probing and the partial fine-tuning of MAE. For the latent-variable spectral representation, the predictor is the posterior-weighted average upon $\PP\rbr{z|x}$, which is exact for discrete latent variables. These observations suggest that the downstream predictor should be designed according to the parametrization used in the representation learning.

\begin{table}[h]
\centering
\caption{Parametrizations of spectral representation and the corresponding downstream predictors.}
\label{tab:predictor}
\small
\resizebox{\linewidth}{!}{\begin{tabular}{llll}
\toprule
Parametrization & Linear sufficient representation & Predictor upon the learned representation & Section \\
\midrule
Linear & $\varphi\rbr{x}$ & linear head & \ref{subsec:regression_classification} \\
Energy-based & $\varphi_\omega\rbr{x}$ (infinite-dimensional) & attention; softmax with projector (linear softmax as first-order truncation) & \ref{sec:energy_spectral} \\
Nonlinear (projector) & $\texttt{nn}_\theta\rbr{\upsilon\rbr{x}}$ & linear head upon $\texttt{nn}_\theta\rbr{\upsilon\rbr{x}}$; nonlinear head upon $\upsilon\rbr{x}$ & \ref{sec:nonlinear_spectral} \\
Asymmetric projector across tokens (MAE) & decoder output & nonlinear head with attention upon $\upsilon\rbr{x}$ & \ref{subsec:nonlinear_spectral_learning} \\
Latent-variable & $\PP\rbr{z|x}$ & posterior-weighted average $\inner{w}{\PP\rbr{\cdot|x}}$; softmax for classification & \ref{sec:lv_spectral} \\
\bottomrule
\end{tabular}}
\end{table}

\section{Spectral Representation in History}\label{sec:spectral_history}

In history, there have been a rich literature in exploiting the spectral representation. Below we establish the connections of the existing SSL algorithms to corresponding dimension reduction/embedding algorithms in statistics and manifold learning from spectral representation view. These classic methods fit into the same framework as the modern SSL methods, differing only in the target, the parametrization, and the learning objective, as summarized in~\tabref{tab:history}.

\begin{table}[h]
\centering
\caption{Classic dimension reduction and embedding methods in the spectral framework.}
\label{tab:history}
\small
\resizebox{\linewidth}{!}{\begin{tabular}{llll}
\toprule
Method & Target & Parametrization & Learning objective \\
\midrule
SNE & Symmetric & Energy-based, nonparametric & Ranking NCE \\
t-SNE & Symmetric & Nonlinear (Cauchy kernel), nonparametric & Jointly normalized NCE \\
NCA & Symmetric (label-defined) & Energy-based, linear & Marginal likelihood / Ranking NCE \\
Laplacian Embedding, Normalized cut & Symmetric & Linear, nonparametric & Variational SVD \\
LPP & Symmetric & Linear, linear map & Variational SVD \\
LLE & Symmetric & Linear, nonparametric & Fixed-point residual with shift $\Lambda = \Ib$ \\
MDS, Isomap, MVU, kernel PCA & Symmetric & Linear, nonparametric & Variational SVD \\
PCA & Symmetric & Linear, linear map & Variational SVD \\
CCA, kernel CCA, CA & Asymmetric & Linear & Variational SVD \\
PLS & Asymmetric & Linear & SVD of cross-covariance \\
\bottomrule
\end{tabular}}
\end{table}

\begin{itemize}[leftmargin=*]

    \item {\bf Stochastic Neighbor Embedding and Neighborhood Component Analysis.} We remark that the stochastic neighbor embedding~(SNE)~\citep{hinton2002stochastic} and neighborhood component analysis~(NCA)~\citep{goldberger2004neighbourhood} can be recovered from NCE~\eqref{eq:rank_nce_ebm} with linear or nonparametric form of $\upsilon\rbr{x}$ in the energy-based spectral representation. 
    
    Specifically, we consider the Gaussian kernel form of the energy-based model, which differs from~\eqref{eq:factor_ebm} only by the norm terms as shown in~\eqref{eq:quadratic_ebm}. Plugging $\upsilon\rbr{x} = Wx$, we have 
    \begin{align}
        \PP\rbr{x'|x} \propto \PP\rbr{x'}\exp\rbr{-\nbr{Wx - Wx'}^2},
    \end{align}
    with corresponding ranking NCE as
    \begin{align}\label{eq:nca_nce}
         \min_{W}\,\, \EE_{\PP\rbr{x, x'}}\sbr{\nbr{Wx - Wx'}^2} + \EE_{\PP\rbr{x}}\sbr{\log\sum_{i=1}^{k}\exp\rbr{-\nbr{Wx - Wx'_i}^2}}. 
    \end{align}
    In NCA, the neighborhood is defined by the class labels, \ie, $\PP\rbr{x'|x}$ is uniform over the samples $C\rbr{x}$ in the same class as $x$, and all the other samples in $\Dcal$ are used in the denominator. Denote $p\rbr{x'|x} = \frac{\exp\rbr{-\nbr{Wx - Wx'}^2}}{\sum_{x_i\in\Dcal}\exp\rbr{-\nbr{Wx - Wx_i}^2}}$. The original $\log$ loss of NCA treats the neighbor $x'$ as a latent variable and maximizes the marginal likelihood of the label agreement, which is lower bounded by~\eqref{eq:nca_nce} through Jensen's inequality, \ie, 
    \begin{align}
        \log \sum_{x'\in C\rbr{x}} p\rbr{x'|x} &= \log\abr{C\rbr{x}} + \log\frac{1}{\abr{C\rbr{x}}}\sum_{x'\in C\rbr{x}} p\rbr{x'|x} \nonumber\\ &\ge \log\abr{C\rbr{x}} + \frac{1}{\abr{C\rbr{x}}}\sum_{x'\in C\rbr{x}}\log p\rbr{x'|x}. 
    \end{align}
    Therefore,~\eqref{eq:nca_nce} is the evidence lower bound of NCA with the uniform variational posterior over $C\rbr{x}$, and the original NCA is recovered by maximizing the marginal likelihood exactly. 
    
    If we consider the nonparametric form $\upsilon\rbr{x_i} = y_i$ for each sample $x_i$ in $\Dcal$, set $\PP\rbr{x_j|x_i} = p_{j|i}\propto \exp\rbr{-\nbr{x_i - x_j}^2/2\sigma_i^2}$ in the input space, and use all the samples in $\Dcal$ in the denominator, the ranking NCE becomes $\sum_i KL\rbr{p_{\cdot|i}||q_{\cdot|i}}$ up to a constant with $q_{j|i}\propto\exp\rbr{-\nbr{y_i - y_j}^2}$, which recovers SNE~\citep{hinton2002stochastic}. Similarly, t-SNE~\citep{maaten2008visualizing} replaces the Gaussian kernel with the Cauchy kernel $\rbr{1 + \nbr{y - y'}^2}^{-1}$, which is a positive-definite kernel, and thus corresponds to the nonlinear spectral representation~\eqref{eq:nonlinear_repr_kernel}. Moreover, t-SNE normalizes over all the pairs rather than for each $x$, \ie, it parametrizes the joint distribution $\PP\rbr{x, x'}\propto\PP\rbr{x}\PP\rbr{x'}\rbr{1 + \nbr{y - y'}^2}^{-1}$ with symmetrized $p_{ij}$.

    \item {\bf Locality Preserving Projections and Laplacian Embedding.} Recall the variational characterization~\eqref{eq:svd_variational} of the spectral decomposition under the orthonormal condition~\eqref{eq:orthonomal_cond}, \ie,
    \begin{equation}\label{eq:le_variational}
        \max_{\xi}\,\, \EE_{\PP\rbr{x, x'}}\sbr{\xi\rbr{x}^\top \xi\rbr{x'}}, \quad \st\,\, \EE_{\PP\rbr{x}}\sbr{\xi\rbr{x}\xi\rbr{x}^\top} = \Ib. 
    \end{equation}
    Locality Preserving Projections and Laplacian Embedding implement~\eqref{eq:le_variational} with linear or nonparametric form of $\xi\rbr{x}$, by estimating the joint distribution with a neighborhood graph. Concretely, given dataset $\Dcal$, we exploit $W_{ij} = \begin{cases} \exp\rbr{-\frac{\nbr{x_i - x_j}^2}{t}}       & \quad \text{if } \nbr{x_i - x_j}\le \epsilon \\
    0  & \quad \text{otherwise }\end{cases}$ and $D = \diag\rbr{W\one}$, and estimate 
    \begin{equation}
        \widehat\PP\rbr{x_i, x_j} = \frac{W_{ij}}{\one^\top W\one}, \quad \widehat\PP\rbr{x_i} = \frac{D_{ii}}{\one^\top W\one}, 
    \end{equation}
    \ie, the degree plays the role of the density. With $\xi\rbr{x_i} = y_i$ and $Y = \sbr{y_1, \ldots, y_n}^\top$, plugging the estimates into~\eqref{eq:le_variational} leads to
    \begin{equation}\label{eq:le}
        \max_{Y}\,\, \tr\rbr{Y^\top W Y}, \quad \st\,\, Y^\top D Y = \rbr{\one^\top W\one}\Ib, 
    \end{equation}
    whose solution satisfies the generalized eigen-decomposition $WY = DY\Lambda$, or equivalently, $\rbr{D - W}Y = DY\rbr{\Ib - \Lambda}$. This recovers the Laplacian Embedding~\citep{belkin2001laplacian}, where the trivial constant eigenvector is discarded. Normalized cut~\citep{shi2000normalized} and spectral clustering decompose $D^{-\frac{1}{2}}WD^{-\frac{1}{2}}$, which is exactly the estimate of $T$ in~\eqref{eq:T}, and is equivalent to~\eqref{eq:le} with $Y = D^{-\frac{1}{2}}U$. 

    If we exploit the linear parametrization $\xi\rbr{x} = Vx\in\RR^{d\times 1}$, \ie, $Y = XV^\top$, in~\eqref{eq:le}, we obtain
    \begin{align}\label{eq:lpp_I}
        \max_{V}\,\, \tr\rbr{ V X^\top W XV^\top}, \quad \st\,\, VX^\top D X V^\top = \rbr{\one^\top W\one}\Ib,
    \end{align}
    which recovers the Locality Preserving Projections~\citep{he2003locality}. 

    \item {\bf Locally Linear Embedding.} Locally linear embedding~(LLE)~\citep{roweis2000nonlinear} can also be recast as the spectral decomposition of an estimated transition operator. LLE first learns the reconstruction weights $W\in\RR^{n\times n}$ by $\min_W \sum_i \nbr{x_i - \sum_j W_{ij}x_j}^2$, with $\sum_j W_{ij} = 1$ and $W_{ij} = 0$ for non-neighbors, and then obtains the embedding by 
    \begin{equation}\label{eq:lle}
        \min_{Y}\,\, \sum_{i=1}^n\nbr{y_i - \sum_j W_{ij}y_j}^2 = \tr\rbr{Y^\top \rbr{\Ib - W}^\top\rbr{\Ib - W} Y}, \quad \st\,\, Y^\top Y = n\Ib, \quad \one^\top Y = 0. 
    \end{equation}
    Since each row of $W$ sums to $1$, we can view $W_{ij}$ as the estimated transition $\widehat\PP\rbr{x_j|x_i}$ in the neighborhood. If $W$ is symmetric with eigen-decomposition $W = \sum_k \gamma_k u_k u_k^\top$, we have $\rbr{\Ib - W}^\top\rbr{\Ib - W} = \sum_k \rbr{1 - \gamma_k}^2 u_k u_k^\top$, which shares the same eigenvectors with $W$. For a stochastic matrix, $\gamma_k\le 1$, therefore, $\rbr{1 - \gamma_k}^2$ is monotonically decreasing in $\gamma_k$, and the bottom-$d$ eigenvectors of $\rbr{\Ib - W}^\top\rbr{\Ib - W}$ are exactly the top-$d$ eigenvectors of $W$, \ie, the spectral representation. The constant eigenvector with $\gamma = 1$, \ie, $W\one = \one$, is excluded by the centering constraint. 
    
    In general, the spectral representation is invariant under a monotone spectral transformation $f\rbr{\cdot}$ of the operator, \ie, $f\rbr{T}$ shares the eigenfunctions with $T$, up to the ordering. Laplacian Embedding corresponds to $f\rbr{\gamma} = \gamma$, LLE corresponds to $f\rbr{\gamma} = \rbr{1 - \gamma}^2$, and diffusion maps~\citep{coifman2006diffusion} correspond to $f\rbr{\gamma} = \gamma^t$. 
    
    From the fixed-point view~\eqref{eq:eigen_fixpoint}, \ie, $\EE\sbr{\varphi\rbr{x'}|x} = \Lambda\varphi\rbr{x}$, the LLE objective~\eqref{eq:lle} is the squared residual of the fixed-point condition with $\Lambda = \Ib$. We emphasize that $\Lambda = \Ib$ is not an assumption that the eigenvalues equal $1$, but a shift: LLE finds the eigenvalues of $W$ closest to $1$, which coincide with the top eigenvalues since $1$ is the upper bound of the spectrum; any shift $\sigma\ge \gamma_{\max}$ leads to the same solution. Indeed, $\Lambda$ cannot be optimized jointly in the residual alone: with $\Lambda$ free, any $d$-dimensional invariant subspace of $W$ achieves zero residual, so the objective does not select the top-$d$ subspace. Adding the reward $-c\tr\rbr{\Lambda}$ with any $c>0$ resolves this, since the residual is nonnegative and $\tr\rbr{Y^\top W Y}/n$ is maximized by the top-$d$ eigenvectors by Ky Fan's theorem, which also achieve zero residual. Alternatively, freezing one side as the previous iterate leads to the power iteration in BYOL~\eqref{eq:byol_lambda} and~\eqref{eq:byol_xi}, where $\Lambda$ is learned as the predictor. Therefore, LLE can be viewed as the classic counterpart of BYOL: both exploit the fixed-point residual, while LLE fixes $\Lambda$ as the shift $\Ib$, and BYOL learns $\Lambda$ with the stop-gradient target. Moreover, LLE only exploits the ordering of the eigenvalues, without scaling the embedding by the eigenvalues as in diffusion maps. 
    
    The equivalence is exact under two conditions. First, $W$ is symmetric, which is not satisfied in general; for asymmetric $W$, the eigenvectors of $\rbr{\Ib - W}^\top\rbr{\Ib - W}$ are the right singular vectors of $\Ib - W$, which only approximate the spectral decomposition of $W$, corresponding to the asymmetric target in~\secref{subsec:asymmetric_spectral}. Second, $W$ is nonnegative; the least-square reconstruction weights may be negative, in which case $W$ is not a probability, the spectral radius may exceed $1$, and the eigenvalues closest to $1$ no longer coincide with the top eigenvalues.

    \item {\bf Multidimensional Scaling, PCA and Isomap.} The principal component analysis~(PCA)~\citep{pearson1901liii}, multidimensional scaling~(MDS)~\citep{kruskal1964multidimensional}, and Isomap~\citep{tenenbaum2000global} can be recast to the same optimization with different estimation of $\frac{\PP\rbr{x, x'}}{\PP\rbr{x}\PP\rbr{x'}}$. 

    Concretely, we start with the variational characteristic property of spectral decomposition~\eqref{eq:svd_variational}, \ie, 
    \begin{align}\label{eq:eig_decomp}
        \varphi(x) = \argmax_{\EE_{\PP\rbr{x}}\sbr{\xi\rbr{x}\xi\rbr{x}^\top} = \Ib}\,\, \underbrace{\int \frac{\PP\rbr{x, x'}}{{\PP\rbr{x}}{\PP\rbr{x'}}}   {\PP\rbr{x}\xi\rbr{x}^\top}\PP\rbr{x'}\xi\rbr{x'}dx dx'}_{
         \EE_{\PP\rbr{x}\PP\rbr{x'}}\sbr{\frac{\PP\rbr{x, x'}}{{\PP\rbr{x}}{\PP\rbr{x'}}} \xi\rbr{x}^\top\xi\rbr{x'}} 
         }.
    \end{align}
    Since all the $d$-rank basis are equivalent in terms of representation ability, in~\eqref{eq:eig_decomp}, we do not enforce for a particular basis. It will be instantiated in different algorithms below. 
    
    MDS~\citep{kruskal1964multidimensional} can be instantiated from~\eqref{eq:eig_decomp} as a U-statistic upon sample $\Dcal$ with nonparametric $\xi\rbr{x}$ with additional constraints. Concretely, given $\Dcal = \cbr{x_i}_{i=1}^n$, we replace $\frac{\PP\rbr{x, x'}}{{\PP\rbr{x}}{\PP\rbr{x'}}}$ by the similarity $k\rbr{x, x'} = -\nbr{x - x'}^2$, and consider the nonparametric $\xi\rbr{x_i} = z_i$ with centering constraint, \ie, $\sum_{i=1} z_i = 0$. To ensure the centering constraint, we consider the reparametrization $Z = HY \in \RR^{n\times d}$, where $H = \Ib - \frac{1}{n}\one\one^\top$. Obviously, it is easy to verify that 
    \begin{equation}
        \sum_{i=1}^n z_i = \one^\top Z = \one^\top Y - \frac{1}{n}\rbr{\one^\top\one}\one^\top Y = \one^\top Y  - \one^\top Y  = 0.
    \end{equation}
    Meanwhile, with the parametrization $Z = \rbr{\Ib - \frac{1}{n}\one\one^\top} Y$, full-rank constraint can also be easily implemented by enforcing $Y^\top Y = n\Ib$. We emphasize that the centering is necessary: since the pointwise mutual information is nonnegative, $k\rbr{x, x'}$ is only a similarity replacing it, which is conditionally positive definite rather than positive definite, \ie, $c^\top K c\ge 0$ only for $\one^\top c = 0$. The centering restricts to this subspace, where $HKH$ is positive semi-definite, and equals the Gram matrix of the centered data up to a constant for Euclidean distances. Meanwhile, the centering also removes the constant direction. With this parametrization of $\xi\rbr{x}$,    
    we obtain the optimization,
    \begin{equation}\label{eq:mds_opt}
        \max_{Y^\top Y = {n}I_d}\sum_{i, j=1}^n -\nbr{x_i - x_j}^2 z_i^\top z_j =\tr\rbr{Y^\top HKHY},
    \end{equation}
    which is exactly MDS. 
    
    Isomap can be naturally extended from MDS with different kernel estimation for $\frac{\PP\rbr{x, x'}}{{\PP\rbr{x}}{\PP\rbr{x'}}}$ and parametrization of $\xi\rbr{x}$. Particularly, in Isomap~\citep{tenenbaum2000global}, the Euclidean kernel $k\rbr{x, x'}$ is replaced by geodesic distance, which is computed based on shortest paths over given dataset. Since the geodesic distance is generally not Euclidean, $HKH$ may have negative eigenvalues, and only the top-$d$ positive eigenvalues are used. Similarly, maximum variance unfolding~(MVU)~\citep{weinberger2006unsupervised} learns the kernel by semidefinite programming, and then applies~\eqref{eq:mds_opt}. 

    Obviously, the optimization~\eqref{eq:mds_opt} is exactly the same as kernel PCA~\citep{scholkopf1998nonlinear}, which implies the MDS is a special case of kernel PCA with special kernel~\citep{
    williams2000connection}. Particularly, with linear kernel and linear parametrization $Y = XW$ in~\eqref{eq:mds_opt}, we recover the classic PCA. 
    For simplicity, we assume $X$ is centralized, \ie, $HX = X$. Plug the linear kernel into~\eqref{eq:mds_opt}, the KKT condition implies, after multiplying $X^\top$ from the left, 
    \begin{align}
        XX^\top Y = Y\Lambda \Rightarrow \cancel{X^\top X}X^\top XW = \cancel{X^\top X}W\Lambda.
    \end{align}
    With the constraint $W^\top X^\top X W = nI$ and KKT condition, we achieve the solution that $W = \sqrt{n}U\Lambda^{-\frac{1}{2}}$, where $U$ and $\Lambda$ denotes the eigenvector and eigenvalue, respectively, \ie, $X^\top X = U \Lambda U^\top$. 

    \item {\bf Canonical Correlation Analysis (CCA).} The canonical correlation analysis~(CCA)~\citep{harold1936relations} can be recovered from~\eqref{eq:asym_svd} for spectral representation in multi-view setting. We start with kernel CCA and reduce to linear CCA as special case. Following the definition of CCA~\citep{bach2002kernel,fukumizu2007statistical,michaeli2015nonparametric}, which is finding the transformation to maximize the correlation, the objective of CCA can be written as
    \begin{align}\label{eq:func_cca}
        \max_{\varphi(\cdot), \mu\rbr{\cdot}}&\,\, \text{cov}\sbr{\varphi, \mu}\defeq \EE_{\PP\rbr{x, y}}\sbr{\varphi\rbr{x}^\top \mu\rbr{y}} = \EE_{\PP\rbr{x}\PP\rbr{y}}\sbr{\frac{\PP\rbr{x, y}}{\PP\rbr{x}\PP\rbr{y}}\varphi\rbr{x}^\top \mu\rbr{y}}\\
        \st&\,\, \EE_{\PP\rbr{x}}\sbr{\varphi\rbr{x}\varphi\rbr{x}^\top} = \Ib, \quad \EE_{\PP\rbr{y}}\sbr{\mu\rbr{y}\mu\rbr{y}^\top} = \Ib, \quad \EE_{\PP\rbr{x}}\sbr{\varphi\rbr{x}} = 0, \quad \EE_{\PP\rbr{y}}\sbr{\mu\rbr{y}} = 0,
    \end{align}
    which is exactly the variational characteristic property of singular value decomposition of~\eqref{eq:asym_svd}, where the centering constraints exclude the trivial constant singular pair. In the empirical version of kernel CCA, regularization is further needed to avoid the degenerate solution~\citep{bach2002kernel}. Plug the linear parametrization, $\varphi(x) = Wx$ and $\mu\rbr{y} = Vy$, we recover classic CCA~\citep{harold1936relations}. With the one-hot features of discrete variables, CCA reduces to the correspondence analysis~\citep{greenacre1984theory}, which applies SVD to the standardized residuals of the contingency table, \ie, the asymmetric counterpart of $T$ in~\eqref{eq:T} with the trivial constant singular pair removed. Without the whitening constraints, the SVD of the cross-covariance leads to the partial least squares~\citep{wold1985partial}. 
    
\end{itemize}

\paragraph{Remark (Trivial Constant Singular Pair):} Since $\int \frac{\PP\rbr{x, x'}}{\PP\rbr{x}\PP\rbr{x'}}\PP\rbr{x'}dx' = 1$, the top singular pair of the pointwise mutual information is always the constant function with singular value $1$, which carries no information. The classic methods remove it in different ways, \ie, the double centering in MDS, the zero-mean constraint in CCA, and discarding the constant eigenvector in Laplacian Embedding and LLE, which are the same operation from the spectral view.

\section{Application of Spectral Representation}\label{sec:application_spectral}
As we discussed in~\secref{sec:sufficient_spectral}, the learned representation is sufficient for downstream regression and classification tasks. In this section, we discuss the usage of the sufficient representation for broader decision-making and inference tasks, including causal inference with unobserved confounders~\citep{sun2024spectral}, control~\citep{ren2025stochastic}, reinforcement learning~\citep{gao2025spectral}, and animal behavior analysis~\citep{wang2025learning}.

As we reveal, for nonlinear sufficient representation, there exists a corresponding linear sufficient representation. Therefore, for simplicity, we take the sufficient linear representation as an example. The same argument can be generalized for applying energy-based and latent variable spectral representation for downstream tasks.

\subsection{Causal Inference with Hidden Confounders}
The spectral representation has been exploited in~\citep{sun2024spectral} for function identification in causal inference with hidden confounder, including instrumental variable~(IV) regression~(\figref{fig:IV_Graph} and~\figref{fig:IV_Graph_obs_confounder}), and proxy causal inference~(\figref{fig:PCL_Graph}). 

We take the IV regression as the example for illustration. We define the conditional expectation operator $E: L_2 \rbr{P_x} \to L_2 \rbr{P_z}$ as $E f = \EE\sbr{f\rbr{x}| z}$ for any $f \in L_2 \rbr{P_x}$. We omit the rigorous assumptions, but only reveal the spectral representation structure.

We follow the notations in~\figref{fig:IV_Graph}. Specifically, we denote the factorization 
\[
\PP\rbr{x|z} = \PP\rbr{x}\inner{\varphi\rbr{x}}{\mu\rbr{z}}. 
\]
Then,~\citet{dai2017learning} recast the IV problem as a saddle-point problem by exploiting Fenchel duality, \ie, 
\begin{align}\label{eq:iv_opt}
    \min_{f}\,\, &\EE_{z}\sbr{\rbr{\EE_{x, y|z}\sbr{y - f\rbr{x}}}^2} + \lambda\Omega\rbr{f}\\
    \Rightarrow\min_{f}\max_g \,\ &\EE_{x, z, y}\sbr{\rbr{y - {f\rbr{x}}}g\rbr{z} - \frac{1}{2}g\rbr{z}^2} + \lambda\Omega\rbr{f},
\end{align}
where $\lambda>0$, $\Omega\rbr{\cdot}$ denotes some regularization,

\begin{figure*}[!h]
\vspace{-2mm}
\begin{center}
\begin{subfigure}{.29\linewidth}
  \centering
  \begin{tikzpicture}[every node/.style={circle,thick,draw}]
    \node[fill=lightgray] (Z) at (0.1,0) {Z};
    \node[fill=lightgray] (X) at (1.9,0) {X};
    \node[fill=lightgray] (Y) at (3.8,0) {Y};
    \node[minimum size=20pt] (e) at (2.85,1.73) {$\epsilon$};
        \draw [-stealth, thick](Z) -- (X);
        \draw [-stealth, thick](X) -- (Y);
        \draw [-stealth, thick](e) -- (Y);
        \draw [-stealth, thick](e) -- (X);
  \end{tikzpicture}
  \caption{Instrumental Variable}
  \label{fig:IV_Graph}
\end{subfigure}%
\begin{subfigure}{.4\linewidth}
  \centering
    \begin{tikzpicture}[every node/.style={circle,thick,draw}]
    \node[fill=lightgray] (Z) at (0.2,0) {Z};
    \node[fill=lightgray] (X) at (2,0) {X};
    \node[fill=lightgray] (Y) at (4,0) {Y};
    \node[fill=lightgray] (O) at (2,1.73) {O};
    \node[minimum size=20pt] (e) at (4,1.73) {$\epsilon$};
    \draw [-stealth, thick](Z) -- (X);
    \draw [-stealth, thick](X) -- (Y);
    \draw [-stealth, thick](e) -- (Y);
    \draw [-stealth, thick](e) -- (X);
    \draw [-stealth, thick](O) -- (X);
    \draw [-stealth, thick](O) -- (Y);
    \end{tikzpicture}  
    \caption{IV with Observable Confounder}
    \label{fig:IV_Graph_obs_confounder}
\end{subfigure}
\begin{subfigure}{.29\linewidth}
  \centering
   \begin{tikzpicture}[every node/.style={circle,thick,draw}]
    \node[fill=lightgray] (Z) at (1.5,1.23) {Z};
    \node[fill=lightgray] (W) at (4.5,1.23) {W};
    \node[fill=lightgray] (X) at (2,0) {X};
    \node[fill=lightgray] (Y) at (4,0) {Y};
    \node[minimum size=20pt] (e) at (3,1.73) {$\epsilon$};
        \draw [stealth-stealth, dashed, thick](Z) -- (X);
        \draw [-stealth, thick](X) -- (Y);
        \draw [-stealth, thick](W) -- (Y);
        \draw [-stealth, thick](e) -- (Y);
        \draw [-stealth, thick](e) -- (X);
        \draw [-stealth, thick](e) -- (Z);
        \draw [stealth-stealth, dashed, thick](e) -- (W);
    \end{tikzpicture}  
    \caption{Proxy Causal Inference}
  \label{fig:PCL_Graph}
\end{subfigure}
\caption{Causal graphs considered, with grey node denoting the observable variable. }
\label{fig:causal_graph}
\vspace{-3mm}
\end{center} 
\end{figure*}

\citet{sun2024spectral} reveal the fact that both $f$ and $g$ can be represented by the primal-dual spectral representation of $\PP\rbr{x|z}$. Specifically, we can represent $Ef$ as 
\begin{align}\label{eq:cme}
    E f =  \int f \rbr{x} \PP(x|z) dx = \inner{\mu\rbr{z}}{\underbrace{\int f\rbr{x}\varphi\rbr{x}\PP\rbr{x}dx}_{w_f}}. 
\end{align}
The optimal dual solution to~\eqref{eq:iv_opt} can be represented linearly by $\mu\rbr{z}$, \ie,   
\[
g^*\rbr{z} = \EE\sbr{y - f\rbr{x}|z} = \underbrace{\EE\sbr{y|z}}_{\inner{\mu\rbr{z}}{w_y}} - \inner{\mu\rbr{z}}{w_f} = \inner{\mu\rbr{z}}{w}. 
\]
For the primal function $f\rbr{\cdot}$, we can always separate as $f\rbr{x} = f^{\parallel}(x) + f^{\perp}(x)$, where
\begin{eqnarray}
    f^{\parallel}(\cdot) \in \Kcal\rbr{\varphi}, \quad f^{\perp}\rbr{\cdot} \in \Ncal\rbr{\varphi}, 
\end{eqnarray}
or equivalently, 
\begin{equation}
    f^\parallel(\cdot) = v_{\parallel}^\top \varphi\rbr{\cdot}, \quad f^\perp(\cdot) = v_\perp^\top \varphi^\perp\rbr{\cdot},
\end{equation}
where $\varphi\rbr{\cdot}$ and $\varphi^\perp(\cdot)$ are the basis for $\Kcal(\varphi)$ and $\Ncal\rbr{\varphi}$ with the condition 
$
\inner{\varphi}{\varphi^\perp}_{\PP\rbr{x}} = 0$.
Then, we have 
\begin{align}\label{eq:iv_param}
\EE_{x|z}\sbr{f(x)}=&\int \PP(x|z)f(x)dx =\inner{\mu\rbr{z}}{\int \varphi\rbr{x}f(x)\PP\rbr{x}dx}\\
=& \inner{\mu\rbr{z}}{\int \varphi\rbr{x}\rbr{f^\parallel(x) + f^\perp(x)}\PP\rbr{x}dx}   \\ 
=&\inner{\mu\rbr{z}}{\int \varphi\rbr{x}f^\parallel(x)\PP\rbr{x}dx} + \underbrace{\inner{\mu\rbr{z}}{\int \varphi\rbr{x}f^\perp(x)\PP\rbr{x}dx}}_{=0},
\end{align}
which indicates that the effective parametrization of $f\rbr{x}$ is $f^\parallel\rbr{x}$. Therefore, we can only consider $f^\parallel\rbr{x} = v^\top \varphi\rbr{x}$, \emph{without any loss}. With this understanding, we have the primal-dual optimization for IV as
\begin{equation}\label{eq:linear_primal_dual}
    \min_{v}\max_{w}\,\, \EE_{x, y, z}\sbr{{w^\top \mu\rbr{z}}\rbr{y - v^\top \varphi\rbr{x}} - \frac{1}{2}\rbr{w^\top \mu\rbr{z}}^2 } + \lambda \Omega\rbr{v^\top \varphi\rbr{x}}.
\end{equation}

For other causal inference problems, \eg, IV with observed confounders and PCL, we can also reveal the spectral representation structure. Please refer to~\citep{sun2024spectral} for details.

\subsection{Control and Reinforcement Learning}

The spectral representation has already been applied in control~\citep{ren2025stochastic}, imitation learning~\citep{ma2025offline}, off-policy evaluation~\citep{hu2024primal} and reinforcement learning~(RL)~\citep{zhang2022making, ren2022spectral, ren2022latent,zhang2023provable,shribak2024diffusion, ma2024skill, nabati2025spectral} for flexible modeling, efficient planning and exploration. 

Concretely, given a Markov Decision Process (MDP) specified by a tuple $\Mcal=\langle \Scal, \Acal, \PP, r , \gamma, d_0\rangle$, where $\Scal$ is the state space, $\Acal$ is the action space, $\PP:\Scal\times\Acal\rightarrow \Delta(\Scal)$ specifies the transition probability of states, $r :\Scal\times\Acal\rightarrow \RR$ is the instantaneous reward function, $\gamma\in[0, 1)$ is the discounting factor and $d_0\in\Delta(\Scal)$ is the initial state distribution. 
In both control and RL, we are seeking for the optimal policy $\pi\rbr{\cdot|s}:\Scal\to\Delta(\Acal)$ that maximize the expected cumulative return, defined as:
\begin{equation}
    \begin{aligned}
        \Jcal(\pi,\Mcal):=\EE_{a_t\sim \pi(\cdot|s_t),s_{t+1}\sim \PP(\cdot|s_t, a_t)}\left[\sum_{t=0}^\infty\gamma^tr (s_t, a_t)\Big|s_0\sim d_0\right].
    \end{aligned}
\end{equation}
The major differences between control and RL mainly lies in the information provided: in control setting, the full information of $\Mcal$ is given, while in RL setting, we can only access to $\Mcal$ by sampling. 

We further define the state value functions $V^\pi: \Scal \to \RR$ and state-action value functions $Q^\pi: \Scal\times\Acal\to \RR$ of a given policy $\pi$ as:
\begin{equation}
    \begin{aligned}
        V^\pi(s)&:=\EE_{a_t\sim \pi(\cdot|s_t), s_{t+1}\sim \PP(\cdot|s_t, a_t)}\left[\sum_{t=0}^\infty\gamma^t r (s_t, a_t)\Big| s_0=s\right],\\
        Q^\pi(s, a)&:=\EE_{a_t\sim \pi(\cdot|s_t), s_{t+1}\sim \PP(\cdot|s_t, a_t)}\left[\sum_{t=0}^\infty\gamma^t r (s_t, a_t)\Big| s_0=s, a_0=a\right].
    \end{aligned}
\end{equation}
Equivalently, the task can be conveniently defined as finding $\pi^*=\argmax_{\pi}\ \EE_{s\sim d_0}\left[V^\pi(s)\right]=\argmax_{\pi}\ \EE_{s\sim d_0, a\sim \pi}\left[Q^\pi(s, a)\right]$.
Based on the definition of $V^\pi$ and $Q^\pi$, we have the following Bellman recursion holds:
\begin{equation}\label{eq:bellman_equation}
    \begin{aligned}
        Q^\pi(s, a)&=r(s, a)+\gamma \EE_{s_{t+1}\sim \PP(\cdot|s_t, a_t)}\left[V^\pi(s_{t+1})\right]\\
        &=r(s, a)+\gamma \EE_{s_{t+1}\sim \PP(\cdot|s_t, a_t), a_{t+1}\sim\pi(\cdot|s_{t+1})}\left[Q^\pi(s_{t+1}, a_{t+1})\right]. 
    \end{aligned}
\end{equation}

We consider the spectral decomposition to $\PP\rbr{s'|s, a}$, \ie,
\begin{equation}\label{eq:spectral_mdp}
    \PP(s'|s, a) =\langle\varphi(s, a), \mu(s')\rangle,\quad \text{and}\quad
    r(s, a)=\langle \varphi(s, a), \theta_r\rangle.
\end{equation}
As argued in~\citep{ren2025stochastic}, the second condition that representing $r\rbr{s, a}$ linearly by $\varphi\rbr{s, a}$ is easy to implement by augmenting original spectral representation for $\PP\rbr{s'|s, a}$ by $r\rbr{s, a}$. Then, for any policy $\pi$, we can represent the $Q^\pi$ linearly by $\varphi\rbr{s, a}$, 
\begin{equation}
    \begin{aligned}
        Q^\pi(s, a) &= r(s, a)+\gamma \int_\Scal \PP(s'|s, a)V^\pi(s')\mathrm{d}s' =\inner{\varphi(s, a)}{\theta_r} +\gamma \inner{\varphi(s, a)}{\int_\Scal \mu(s')V^\pi(s')\mathrm{d}s'}\\
        &=\bigg\langle\varphi(s, a),\underbrace{\theta_r+\gamma \int_\Scal \mu(s')V^\pi(s')\mathrm{d}s'}_{\boldsymbol{\eta}^\pi}\bigg\rangle,
    \end{aligned}
\end{equation}
which implies efficient temporal difference learning for fitting $Q^\pi$ by least square optimization~\citep{ren2025stochastic}, and tractable uncertainty estimation for exploration~\citep{zhang2022making, ren2022free}. 

The linear structure of $Q^\pi$ also facilitates the inverse problem, \ie, inferring the reward and the behavior primitives from demonstrations, which has been exploited for animal behavior analysis~\citep{wang2025learning}. Specifically, with the maximum-entropy policy $\pi\rbr{a|s}\propto\exp\rbr{\inner{\varphi\rbr{s, a}}{\eta}}$, the spectral representation $\varphi\rbr{s, a}$ learned from the transition data serves as a set of task-agnostic motor primitives, and the weights $\eta$ are fitted by maximum likelihood on the demonstrated state-action pairs. For continuous dynamics, the energy-based spectral representation in~\secref{sec:energy_spectral} is learned by NCE, with a projector upon it as in~\secref{sec:nonlinear_spectral}. Moreover, the time-varying weights $\eta\rbr{t}$ with a smoothness prior model the behavior as continuously evolving mixtures of the primitives, from which the time-varying rewards can be recovered. 

We only briefly introduce here about exploiting spectral representation for MDP planning. Please refer to~\citep{gao2025spectral} for more details about (partially observable) MDP learning, planning and exploration with comprehensive comparison. 

\section{Conclusion}\label{sec:conclusion}

After this long journey, we have systematically revealed the unified key view behind representation learning algorithms, from the classic component analysis, to manifold learning, to modern self-supervised learning: 
\begin{center}
    \emph{All of these algorithms are aiming for extracting spectral representation for pointwise mutual information. }
\end{center}
Here the pointwise mutual information is between paired variables, \eg, augmented views, corrupted and clean data, or different modalities. This view not only clarifies the long-standing confusion about the sufficiency of representation and organizes the existing algorithms in a clear way for new researchers, but also provides tools for algorithms analysis and design, which inspires a few novel algorithms as we presented. In particular, the denoising view shows that the corruption preserves the linear spectral representation, and naturally leads to the lifted and asymmetric parametrizations, with the DAE and the diffusion spectral representation as mirror images of each other. The classic methods are also connected to the modern ones, \eg, LLE is the classic counterpart of BYOL, and the double centering in MDS, the zero-mean constraint in CCA, and discarding the constant eigenvector in Laplacian Embedding are the same operation that removes the trivial constant singular pair. Moreover, the spectral representation extends beyond regression and classification to causal inference, control and reinforcement learning, and animal behavior analysis. 

However, we emphasize that the empirical performance gaps among different algorithms suggest that the {\bf parametrization} and {\bf optimization} for spectral representation matters. For example, the direct spectral decomposition parametrization~\eqref{eq:reparam_repr} induces downstream linear predictor, but with finite-dimension assumption, therefore, limited expressiveness; while the energy-based spectral representation~\eqref{eq:spectral_ebm} eliminates the finite-dimension assumption and becomes more flexible, but nonlinear downstream predictor~\eqref{eq:ebm_repr}. Similarly, the representation of MAE in~\secref{subsec:nonlinear_spectral_learning} is read out through the discarded decoder, which explains its weak linear probing but strong attentive probing performance. Meanwhile, the optimization objectives and algorithms for spectral representation seeking are also important, while the traditional understanding through contrastive vs non-contrastive is incomplete: the square contrastive loss and the variational power iteration in BYOL and MINC are unbiased with any batch size; the major criticism to contrastive SSL is the large batch size, however, the ranking-based NCE is consistent with any number of negative samples, and the batch size only affects its statistical efficiency under a well-specified model. In contrast, as we specified in~\secref{subsec:spectral_ssl} and~\ref{sec:objective_discussion}, the non-contrastive VICReg and Barlow-Twins induce biased gradients, leading to inconsistent solutions with small batch sizes and performance degeneration. 

Ablating the effect of parametrization and optimization and finding the optimal empirical configuration is out of the scope of this paper, and we leave it as future work.

\section*{Acknowledgements}
{This paper is supported by NSF AI institute: 2112085, NSF ECCS-2401390, NSF ECCS2401391, ONR N000142512173, and NSF IIS2403240.
We thank Jincheng Mei, Daniel Guo in Google DeepMind and Chenxiao Gao, Haotian Sun in Georgia Tech for helpful thoughts and discussions.
}

\vskip 0.2in
\bibliographystyle{plainnat}
\bibliography{ref}

\clearpage
\begin{appendices}
\section{Proofs}\label{app:proofs}

\subsection{Weighted Barlow-Twins Objective}\label{app:bt_lambda}

\begin{theorem}\label{thm:bt_lambda}
Assume $\PP\rbr{x, x'} = \PP\rbr{x', x}$ is symmetric and the associated pointwise mutual information operator on $L_2\rbr{\PP\rbr{x}}$ is compact with top eigenvalues $\gamma_1 \ge \ldots \ge \gamma_d \ge \gamma_{d+1}$ ($\gamma_d \ge 0$, which holds automatically when the operator is positive semidefinite as in~\eqref{eq:sufficient_repr}). Under the orthonormal constraint $\EE_{\PP\rbr{x}}\sbr{\xi\rbr{x}\xi\rbr{x}^\top} = \Ib$, for any $\lambda \ge 0$, the minimum of the weighted Barlow-Twins objective~\eqref{eq:barow_twins_II},
\begin{equation*}
    L_\lambda\rbr{C} = \sum_{i=1}^d\rbr{1 - C_{ii}}^2 + \lambda \sum_{i\neq j} C_{ij}^2, \quad C = \EE_{\PP\rbr{x, x'}}\sbr{\xi\rbr{x}\xi\rbr{x'}^\top},
\end{equation*}
is attained in the top-$d$ eigen-space of the pointwise mutual information operator, and uniquely spans it whenever $\gamma_d > \gamma_{d+1}$.
\end{theorem}

\begin{proof}
\textbf{Preparation.} Under $\EE_{\PP\rbr{x}}\sbr{\xi\rbr{x}\xi\rbr{x}^\top} = \Ib$, any feasible $\xi$ forms an $L_2\rbr{\PP\rbr{x}}$-orthonormal basis of a $d$-dimensional subspace $S$, and $C$ is the symmetric matrix of the self-adjoint pointwise mutual information operator restricted to $S$, expressed in the basis $\xi$. Therefore, the real eigenvalues $\mu_1\ge\ldots\ge\mu_d$ of $C$ only depend on $S$, and changing the orthonormal basis within $S$ rotates $C$ into $R^\top C R$ with an orthogonal matrix $R\in O\rbr{d}$, preserving its eigenvalues. We exploit two facts: \emph{i)} by the Courant-Fischer / Poincar\'e separation theorem, $\mu_k\le\gamma_k$ for all $k = 1, \ldots, d$ and any $d$-dimensional subspace $S$, with equality for all $k$ when $S$ is the top-$d$ eigen-space (and strictly $\mu_k < \gamma_k$ for at least one $k$ on any other subspace when $\gamma_d > \gamma_{d+1}$); and \emph{ii)} $\gamma_k\le 1$ by~\eqref{eq:pmi_eigen_bound}.

\textbf{Key identity.} Since the off-diagonal entries of $\Ib - C$ are $-C_{ij}$, we have $\nbr{\Ib - C}_F^2 = \sum_i\rbr{1 - C_{ii}}^2 + \sum_{i\neq j}C_{ij}^2$, and thus
\begin{equation*}
    L_\lambda\rbr{C} = \rbr{1 - \lambda}\sum_{i=1}^d \rbr{1 - C_{ii}}^2 + \lambda\nbr{\Ib - C}_F^2,
\end{equation*}
\ie, a weighted combination of the diagonal part and the overall distance between $C$ and $\Ib$.

\textbf{Key construction.} Consider any competitor, \ie, a subspace $S'$ with an orthonormal basis in it, and write $C' = Q\,\diag\rbr{\mu'}Q^\top$ with an orthogonal matrix $Q\in O\rbr{d}$. Since any rotation can be realized within the top-$d$ eigen-space, we construct a candidate in the top-$d$ eigen-space using the same rotation $Q$,
\begin{equation*}
    C = Q\,\diag\rbr{\gamma}Q^\top,
\end{equation*}
which keeps the orientation of $C'$ and only replaces the eigenvalues $\mu'_k$ with the larger ones $\gamma_k\le 1$, and is therefore closer to $\Ib$:
\begin{equation*}
    \Ib - C = Q\,\diag\rbr{1 - \gamma}Q^\top, \quad \Ib - C' = Q\,\diag\rbr{1 - \mu'}Q^\top, \quad 0\le 1 - \gamma_k \le 1 - \mu'_k.
\end{equation*}

\textbf{Case $\lambda\in\sbr{0, 1}$.} Both weights $1-\lambda$ and $\lambda$ are nonnegative, so it suffices to show neither term increases. For the overall term, $\nbr{\Ib - C}_F^2 = \sum_k\rbr{1 - \gamma_k}^2 \le \sum_k\rbr{1 - \mu'_k}^2 = \nbr{\Ib - C'}_F^2$. For the diagonal term, $1 - C_{ii} = \sum_k Q_{ik}^2\rbr{1 - \gamma_k}$ is a convex combination of the nonnegative gaps $1 - \gamma_k$ with weights $\sum_k Q_{ik}^2 = 1$, so $0 \le 1 - C_{ii} \le 1 - C'_{ii}$, which implies $\rbr{1 - C_{ii}}^2\le\rbr{1 - C'_{ii}}^2$ and hence $L_\lambda\rbr{C}\le L_\lambda\rbr{C'}$. Moreover, if $\mu'_k < \gamma_k$ for some $k$, then both $\nbr{\Ib - C}_F^2 < \nbr{\Ib - C'}_F^2$ and (since $\sum_i Q_{ik}^2 = 1$ implies $Q_{ik}^2 > 0$ for at least one $i$) $0 \le 1 - C_{ii} < 1 - C'_{ii}$ for that $i$, making $L_\lambda\rbr{C} < L_\lambda\rbr{C'}$ strict for every $\lambda\in\sbr{0, 1}$.

\textbf{Case $\lambda > 1$.} The weight $1 - \lambda$ of the diagonal term is negative, so we instead compare against the eigen-basis within the top-$d$ eigen-space, \ie, $C = \diag\rbr{\gamma}$. Its off-diagonal entries vanish, giving $L_\lambda\rbr{C} = \sum_k\rbr{1 - \gamma_k}^2$. For the competitor,
\begin{equation*}
    L_\lambda\rbr{C'} = \nbr{\Ib - C'}_F^2 + \rbr{\lambda - 1}\sum_{i\neq j}C_{ij}'^2 \ge \nbr{\Ib - C'}_F^2 = \sum_k\rbr{1 - \mu'_k}^2 \ge \sum_k\rbr{1 - \gamma_k}^2 = L_\lambda\rbr{C},
\end{equation*}
where the second inequality is again strict whenever $\mu'_k < \gamma_k$ for some $k$.

\textbf{Conclusion.} For any competitor, there is a candidate in the top-$d$ eigen-space that is no worse, so the minimum is attained in the top-$d$ eigen-space; when $\gamma_d > \gamma_{d+1}$, any other subspace has $\mu'_k < \gamma_k$ for some $k$, making the inequalities strict so every minimizer spans the top-$d$ eigen-space uniquely. Within the top-$d$ eigen-space, a sharp transition occurs at $\lambda = 1$:
\begin{itemize}[leftmargin=*]
    \item For $\lambda > 1$, minimizing $L_\lambda$ requires $\sum_{i\neq j}C_{ij}^2 = 0$, so the minimizer is the eigen-basis (up to sign flips, permutations, and rotations within repeated eigenvalues).
    \item For $\lambda = 1$, $L_1\rbr{C} = \nbr{\Ib - C}_F^2$ is orthogonally invariant, so \emph{any} orthonormal basis of the top-$d$ eigen-space is a minimizer.
    \item For $\lambda\in\sbr{0, 1}$, since $\nbr{\Ib - C}_F^2$ and $\tr\rbr{C} = \sum_{k=1}^d \gamma_k$ are invariant to rotations $C = Q\,\diag\rbr{\gamma}Q^\top$ within the top-$d$ eigen-space, we have $L_\lambda\rbr{C} = \text{const} + \rbr{1 - \lambda}\sum_{i=1}^d C_{ii}^2$. By Cauchy-Schwarz and the Schur-Horn theorem, for $\lambda\in [0, 1)$ this is minimized if and only if $Q$ rotates the eigen-basis to equalize all diagonal entries, $C_{11} = \ldots = C_{dd} = \frac{1}{d}\sum_{k=1}^d \gamma_k$. For example, with $\gamma = \rbr{1, 0}$, the eigen-basis gives $L_\lambda = 1$, whereas rotating by $45^\circ$ gives $C_{11} = C_{22} = \frac{1}{2}$ and $L_\lambda = \frac{1+\lambda}{2} < 1$ for all $\lambda\in [0, 1)$.
\end{itemize}
Finally, note that $\lambda = 0$ avoids dimensional collapse here because the full whitening constraint $\EE\sbr{\xi\rbr{x}\xi\rbr{x}^\top} = \Ib$ already enforces orthogonality across coordinates; practical Barlow-Twins replaces full whitening with per-coordinate batch normalization, and thus requires $\lambda > 0$ to prevent all coordinates from collapsing onto $\gamma_1$.
\end{proof}

\subsection{Bias of the Barlow-Twins Objective}\label{app:bt_bias}

We derive the bias of the mini-batch Barlow-Twins objective with batch normalization. For clarity, we consider the one-dimensional case; the multi-dimensional case shares the same structure, with the normalization replaced by the whitening $\widehat\Sigma^{-\frac{1}{2}}$, which is a nonlinear function of the batch covariance $\widehat\Sigma$, and the batch normalization only takes its diagonal part.

\textbf{Setup.} For a positive pair $\rbr{x_i, x'_i}$, denote
\begin{equation*}
    a_i = \varphi\rbr{x_i}\varphi\rbr{x'_i}, \quad c_i = \frac{1}{2}\rbr{\varphi\rbr{x_i}^2 + \varphi\rbr{x'_i}^2},
\end{equation*}
with $m = \EE\sbr{a}$ and $v = \EE\sbr{c}$, so the population cross-correlation is $\rho = C = \frac{m}{v}$. With a batch of size $B$, Barlow-Twins estimates it by the ratio of two batch averages,
\begin{equation*}
    \widehat C = \frac{\widehat m}{\widehat v}, \quad \widehat m = \frac{1}{B}\sum_{i=1}^B a_i, \quad \widehat v = \frac{1}{B}\sum_{i=1}^B c_i,
\end{equation*}
where $\widehat m$ and $\widehat v$ are unbiased, but their ratio is not.

\textbf{Delta method.} Expanding the ratio around $\rbr{m, v}$ to the second order, we have the bias and the variance of the ratio estimator,
\begin{align*}
    \EE\sbr{\widehat C} &= \rho + \frac{1}{B}\frac{\rho\operatorname{Var}\rbr{c} - \operatorname{Cov}\rbr{a, c}}{v^2} + \Ocal\rbr{\frac{1}{B^2}}, \\
    \operatorname{Var}\rbr{\widehat C} &= \frac{1}{B}\frac{\operatorname{Var}\rbr{a - \rho c}}{v^2} + \Ocal\rbr{\frac{1}{B^2}}.
\end{align*}
Plugging into $\EE\sbr{\rbr{1 - \widehat C}^2} = \rbr{1 - \EE\sbr{\widehat C}}^2 + \operatorname{Var}\rbr{\widehat C}$, we obtain
\begin{align*}
    \EE_B\sbr{\rbr{1 - \widehat C}^2} &= \rbr{1 - \rho}^2 + \frac{1}{B}R\rbr{\varphi} + \Ocal\rbr{\frac{1}{B^2}}, \\
    R\rbr{\varphi} &= \frac{\operatorname{Var}\rbr{a - \rho c} - 2\rbr{1 - \rho}\rbr{\rho\operatorname{Var}\rbr{c} - \operatorname{Cov}\rbr{a, c}}}{v^2}.
\end{align*}

\textbf{Interpretation.} The bias is of order $\frac{1}{B}$, and $R\rbr{\varphi}$ depends on the fourth moments of $\varphi$ through $\operatorname{Var}\rbr{a}$, $\operatorname{Var}\rbr{c}$, and $\operatorname{Cov}\rbr{a, c}$, rather than being a constant. Therefore, the minimizer of the expected mini-batch loss differs from that of the population loss for a fixed $B$, and only coincides as $B\rightarrow\infty$, which explains the large batch size required by Barlow-Twins. Moreover, $R\rbr{\varphi}$ consists of two parts from different sources: the variance term $\frac{\operatorname{Var}\rbr{a - \rho c}}{v^2}$ comes from the square of the estimate, whose part from the numerator can be removed by U-statistics, \ie, using only distinct pairs in the batch, as for the square contrastive loss; while the term $-\frac{2\rbr{1 - \rho}\rbr{\rho\operatorname{Var}\rbr{c} - \operatorname{Cov}\rbr{a, c}}}{v^2}$ comes from the division by the estimated variance in the batch normalization, which cannot be removed as long as the batch statistics are used for the normalization.

\subsection{Ranking-based NCE under Model Misspecification}\label{app:nce_misspec}

We analyze the population optimum of the ranking-based NCE with $k$ candidates under model misspecification. Consider the score functions $\cbr{s_\theta\rbr{x, x'}: \theta\in\Theta}$, \eg, $s_\theta\rbr{x, x'} = \upsilon\rbr{x}^\top\upsilon\rbr{x'}$, and the population objective
\begin{equation*}
    J_k\rbr{\theta} = \EE_{x,\, x'_1\sim\PP\rbr{x'|x},\, x'_{2:k}\sim\PP\rbr{x'}}\sbr{s_\theta\rbr{x, x'_1} - \log\sum_{j=1}^k\exp\rbr{s_\theta\rbr{x, x'_j}}}.
\end{equation*}
Denote the partition function and the conditional model
\begin{equation*}
    Z_\theta\rbr{x} = \EE_{\PP\rbr{x'}}\sbr{\exp\rbr{s_\theta\rbr{x, x'}}}, \quad q_\theta\rbr{x'|x} = \frac{\PP\rbr{x'}\exp\rbr{s_\theta\rbr{x, x'}}}{Z_\theta\rbr{x}},
\end{equation*}
and the conditional maximum likelihood objective
\begin{align*}
    J_\infty\rbr{\theta} &= \EE_{\PP\rbr{x, x'}}\sbr{s_\theta\rbr{x, x'}} - \EE_{\PP\rbr{x}}\sbr{\log Z_\theta\rbr{x}} \\
    &= -\EE_{\PP\rbr{x}}\sbr{KL\rbr{\PP\rbr{\cdot|x}||q_\theta\rbr{\cdot|x}}} + \text{const}.
\end{align*}

\begin{proposition}\label{prop:nce_misspec}
Assume \emph{i)} the scores are bounded, \ie, $\abr{s_\theta}\le M$ for all $\theta\in\Theta$; and \emph{ii)} in an identifiable parametrization (or modulo gauge symmetries such as orthogonal rotations of $\upsilon$ and $x$-dependent shifts of $s_\theta$ that leave $q_\theta$ invariant), $J_\infty$ has a unique and well-separated maximizer $\theta_\infty$, \ie, $\sup_{d\rbr{\theta, \theta_\infty}\ge\epsilon}J_\infty\rbr{\theta} < J_\infty\rbr{\theta_\infty}$ for any $\epsilon > 0$. Then, 
\begin{equation*}
    \sup_{\theta\in\Theta}\abr{J_k\rbr{\theta} + \log k - J_\infty\rbr{\theta}} \le \frac{C e^{2M}}{\sqrt{k}},
\end{equation*}
for a constant $C$, and any maximizer $\theta_k$ of $J_k$ converges to $\theta_\infty$, \ie, the KL projection of the conditional maximum likelihood estimation, as $k\rightarrow\infty$. Moreover, uniformly over $\theta\in\Theta$,
\begin{align*}
    J_k\rbr{\theta} + \log k &= J_\infty\rbr{\theta} + \frac{c\rbr{\theta}}{k} + \Ocal\rbr{\frac{1}{k^2}}, \\
    c\rbr{\theta} &= -\EE_{\PP\rbr{x}}\sbr{\frac{\EE_{\PP\rbr{x'|x}}\sbr{e^{s_\theta\rbr{x, x'}}} - Z_\theta\rbr{x}}{Z_\theta\rbr{x}} - \frac{\operatorname{Var}_{\PP\rbr{x'}}\rbr{e^{s_\theta\rbr{x, x'}}}}{2Z_\theta\rbr{x}^2}} \\
    &= \frac{1}{2}\EE_{\PP\rbr{x}\PP\rbr{x'}}\sbr{\rbr{\frac{q_\theta\rbr{x'|x}}{\PP\rbr{x'}} - \frac{\PP\rbr{x'|x}}{\PP\rbr{x'}}}^2} - \frac{1}{2}\EE_{\PP\rbr{x}}\sbr{\chi^2\rbr{\PP\rbr{\cdot|x}||\PP\rbr{\cdot}}},
\end{align*}
where the last identity holds whenever the $\chi^2$-divergence is finite. Assuming further that $\theta\mapsto s_\theta\rbr{x, x'}$ is continuously differentiable in a neighborhood of $\theta_\infty\in\operatorname{int}\rbr{\Theta}$ with $\nbr{\nabla_\theta s_\theta}\le M'$, we have $\theta_k\neq\theta_\infty$ for all sufficiently large $k$ when $\nabla c\rbr{\theta_\infty}\neq 0$, and $\nbr{\theta_k - \theta_\infty} = \Ocal\rbr{1/k}$ when $J_\infty$ is locally strongly concave around $\theta_\infty$.
\end{proposition}

\begin{proof}
\textbf{Limit objective.} Splitting the $\log$-sum, we have
\begin{equation*}
    J_k\rbr{\theta} + \log k = \EE\sbr{s_\theta\rbr{x, x'_1}} - \EE\sbr{\log \bar S_k}, \quad \bar S_k = \frac{1}{k}\sum_{j=1}^k\exp\rbr{s_\theta\rbr{x, x'_j}},
\end{equation*}
and $\bar S_k\rightarrow Z_\theta\rbr{x}$ by the law of large numbers, which leads to $J_\infty$. 

\textbf{Uniform convergence.} By the boundedness, $\bar S_k$ and $Z_\theta\rbr{x}$ lie in $\sbr{e^{-M}, e^{M}}$, on which $\log$ is $e^{M}$-Lipschitz. Writing $\Delta_k = \bar S_k - Z_\theta\rbr{x}$ as
\begin{equation*}
    \Delta_k = \frac{1}{k}\rbr{e^{s_\theta\rbr{x, x'_1}} - Z_\theta\rbr{x}} + \frac{1}{k}\sum_{j=2}^k\rbr{e^{s_\theta\rbr{x, x'_j}} - Z_\theta\rbr{x}},
\end{equation*}
the first term is bounded by $\frac{2e^{M}}{k}$, and the second term has conditional variance at most $\frac{e^{2M}}{k}$, so $\EE\abr{\Delta_k}\le \frac{2e^{M}}{k} + \frac{e^{M}}{\sqrt{k}}$, uniformly in $\theta$. Therefore, $\sup_\theta\abr{J_k\rbr{\theta} + \log k - J_\infty\rbr{\theta}}\le e^{M}\rbr{\frac{2e^{M}}{k} + \frac{e^{M}}{\sqrt{k}}}\le\frac{Ce^{2M}}{\sqrt{k}}$.

\textbf{Convergence of the maximizers.} Since $\log k$ does not depend on $\theta$, $\theta_k$ also maximizes $J_k + \log k$. By the uniform convergence, $J_\infty\rbr{\theta_k}\ge J_k\rbr{\theta_k} + \log k - \frac{Ce^{2M}}{\sqrt{k}}\ge J_k\rbr{\theta_\infty} + \log k - \frac{Ce^{2M}}{\sqrt{k}}\ge J_\infty\rbr{\theta_\infty} - \frac{2Ce^{2M}}{\sqrt{k}}$, and the well-separated maximizer implies $\theta_k\rightarrow\theta_\infty$.

\textbf{Asymptotic expansion and rate.} Expanding $\log\bar S_k$ around $Z_\theta\rbr{x}$ to the third order with the fourth-order integral remainder,
\begin{equation*}
    \log\bar S_k - \log Z_\theta\rbr{x} = \frac{\Delta_k}{Z_\theta\rbr{x}} - \frac{\Delta_k^2}{2Z_\theta\rbr{x}^2} + \frac{\Delta_k^3}{3Z_\theta\rbr{x}^3} - R_k\rbr{\theta}, \quad R_k\rbr{\theta} = \Delta_k^4\int_0^1 \frac{\rbr{1-\tau}^3}{\rbr{Z_\theta\rbr{x} + \tau\Delta_k}^4}d\tau,
\end{equation*}
where $Z_\theta\rbr{x} + \tau\Delta_k\ge e^{-M}$. Conditioned on $x$, the summands $Y_j = e^{s_\theta\rbr{x, x'_j}} - Z_\theta\rbr{x}$ for $j=2, \ldots, k$ are independent, bounded, and zero-mean, so the cross terms in the central moments cancel, yielding
\begin{align*}
    \EE\sbr{\Delta_k|x} &= \frac{1}{k}\rbr{\EE_{\PP\rbr{x'|x}}\sbr{e^{s_\theta}} - Z_\theta\rbr{x}}, & \EE\sbr{\Delta_k^2|x} &= \frac{1}{k}\operatorname{Var}_{\PP\rbr{x'}}\rbr{e^{s_\theta}} + \Ocal\rbr{\frac{1}{k^2}}, \\
    \EE\sbr{\Delta_k^3|x} &= \Ocal\rbr{\frac{1}{k^2}}, & \EE\sbr{\Delta_k^4|x} &= \Ocal\rbr{\frac{1}{k^2}},
\end{align*}
uniformly in $\theta$. Taking expectations gives the $\Ocal\rbr{1/k^2}$ expansion with $c\rbr{\theta}$, and substituting $\frac{e^{s_\theta\rbr{x, x'}}}{Z_\theta\rbr{x}} = \frac{q_\theta\rbr{x'|x}}{\PP\rbr{x'}}$ (which has mean $1$ under $\PP\rbr{x'}$) completes the square into the density-ratio form.

Under the differentiability assumption near $\theta_\infty\in\operatorname{int}\rbr{\Theta}$, differentiating the expansion above and noting that $\nabla_\theta Y_j$ ($j\ge 2$) are also conditionally independent, bounded, and zero-mean given $x$---so $\EE\sbr{\abr{\Delta_k}^3\nbr{\nabla_\theta\Delta_k}}\le\rbr{\EE\sbr{\Delta_k^6}}^{1/2}\rbr{\EE\sbr{\nbr{\nabla_\theta\Delta_k}^2}}^{1/2} = \Ocal\rbr{k^{-3/2}}\Ocal\rbr{k^{-1/2}} = \Ocal\rbr{1/k^2}$ and hence $\nbr{\nabla_\theta\EE\sbr{R_k\rbr{\theta}}} = \Ocal\rbr{1/k^2}$---yields uniformly near $\theta_\infty$:
\begin{equation*}
    \nabla J_k\rbr{\theta} = \nabla J_\infty\rbr{\theta} + \frac{1}{k}\nabla c\rbr{\theta} + \Ocal\rbr{\frac{1}{k^2}}.
\end{equation*}
Since $\nabla J_\infty\rbr{\theta_\infty} = 0$, if $\nabla c\rbr{\theta_\infty}\neq 0$, then $\nabla J_k\rbr{\theta_\infty}\neq 0$ for all sufficiently large $k$, so $\theta_k\neq\theta_\infty$. Finally, when $J_\infty$ is locally $\mu$-strongly concave around $\theta_\infty$, for all sufficiently large $k$, $\theta_k$ lies in this neighborhood with $\nabla J_k\rbr{\theta_k} = 0$, and thus
\begin{equation*}
    \mu\nbr{\theta_k - \theta_\infty} \le \nbr{\nabla J_\infty\rbr{\theta_k} - \nabla J_\infty\rbr{\theta_\infty}} = \nbr{\nabla J_\infty\rbr{\theta_k} - \nabla J_k\rbr{\theta_k}} = \nbr{\frac{1}{k}\nabla c\rbr{\theta_k} + \Ocal\rbr{\frac{1}{k^2}}},
\end{equation*}
which gives $\nbr{\theta_k - \theta_\infty} = \Ocal\rbr{1/k}$.
\end{proof}

\paragraph{Remark (Relation to~\citet{ma2018noise}):} Proposition~\ref{prop:nce_misspec} is complementary to the analysis in~\citet{ma2018noise}. \citet{ma2018noise} consider the well-specified model, \ie, the model family contains the true distribution, and analyze the finite-sample estimator as the sample size $n\rightarrow\infty$: the ranking-based NCE is consistent for any fixed $k$, and its asymptotic variance approaches that of MLE as $k\rightarrow\infty$. Therefore, $k$ only affects the variance, \ie, the statistical efficiency. In contrast, Proposition~\ref{prop:nce_misspec} considers the misspecified model, which is the common case with a finite-dimensional representation, and analyzes the population optimum with $n=\infty$: distinct $k$ generically lead to distinct optima, and $k$ affects the bias, \ie, the learned representation itself, which converges to the KL projection of MLE at the rate $\Ocal\rbr{1/k}$. In the well-specified case, the optima of $J_k$ coincide with the true distribution for all $k$, recovering the consistency in~\citet{ma2018noise}. Combining both, a larger $k$ improves the statistical efficiency in the well-specified case, and further moves the learned representation towards the MLE projection in the misspecified case.

\subsection{Projected Representation Learning}\label{app:projected}

\begin{proposition}\label{prop:projected}
Assume $m\ge d$ and $\EE\sbr{\nbr{h\rbr{x'}}_2^2} < \infty$. Denote $g\rbr{x} = \EE\sbr{h\rbr{x'}|x}\in\RR^m$ and $\Sigma_g = \EE\sbr{g\rbr{x}g\rbr{x}^\top}$ with eigenvalues $\sigma_1\ge\cdots\ge\sigma_m\ge 0$ ($\sigma_{m+1}\defeq 0$) and orthonormal eigenvectors $u_1, \ldots, u_m$. Over measurable $\varphi\in L_2\rbr{\PP\rbr{x}; \RR^d}$ and $W\in\RR^{m\times d}$ with $W^\top W = \Ib$,~\eqref{eq:projected_repr} is minimized by $W = \sbr{u_1, \ldots, u_d}R$ with any orthogonal $R\in O\rbr{d}$ and $\varphi\rbr{x} = W^\top g\rbr{x}$ (uniquely so whenever $\sigma_d > \sigma_{d+1}$). Moreover, $\EE\sbr{\varphi\rbr{x}\varphi\rbr{x}^\top} = R^\top\diag\rbr{\sigma_1, \ldots, \sigma_d}R$ (diagonal when $R = \Ib$), and when $\sigma_d > 0$, $\varphi\rbr{x}$ is non-redundant and spans the top-$d$ left singular subspace of the operator $a\mapsto a^\top g\rbr{x}$.
\end{proposition}
\begin{proof}
Since $\EE\sbr{h\rbr{x'} - g\rbr{x}|x} = 0$ and $\EE\sbr{\nbr{h\rbr{x'}}_2^2} < \infty$, the objective decomposes as
\begin{equation*}
    \EE\sbr{\nbr{W\varphi\rbr{x} - h\rbr{x'}}_2^2} = \EE\sbr{\nbr{W\varphi\rbr{x} - g\rbr{x}}_2^2} + \EE\sbr{\nbr{h\rbr{x'} - g\rbr{x}}_2^2},
\end{equation*}
where the second term is a finite constant. For fixed $W$ with $W^\top W = \Ib$, the first term is minimized over $\varphi\in L_2\rbr{\PP\rbr{x}; \RR^d}$ pointwise by $\varphi\rbr{x} = W^\top g\rbr{x}$, the projection of $g\rbr{x}$ onto the column space of $W$, with the minimum
\begin{equation*}
    \EE\sbr{\nbr{\rbr{\Ib - WW^\top}g\rbr{x}}_2^2} = \tr\rbr{\Sigma_g} - \tr\rbr{W^\top\Sigma_g W}.
\end{equation*}
By Ky Fan's maximum principle~\citep{fan1949theorem}, \ie, $\max_{W^\top W = \Ib}\tr\rbr{W^\top\Sigma_g W} = \sum_{k=1}^d\sigma_k$ for symmetric $\Sigma_g$, $\tr\rbr{W^\top\Sigma_g W}$ is maximized under $W^\top W = \Ib$ by $W = \sbr{u_1, \ldots, u_d}R$ for any $R\in O\rbr{d}$, and uniquely so whenever $\sigma_d > \sigma_{d+1}$. This gives $\EE\sbr{\varphi\rbr{x}\varphi\rbr{x}^\top} = W^\top\Sigma_g W = R^\top\diag\rbr{\sigma_1, \ldots, \sigma_d}R$, so the components of $\varphi$ are linearly independent (\ie, non-redundant) in $L_2\rbr{\PP\rbr{x}}$ if and only if $\sigma_d > 0$, and are further uncorrelated when $R = \Ib$. Finally, the operator $K: a\mapsto a^\top g\rbr{x}$ from $\RR^m$ to $L_2\rbr{\PP\rbr{x}}$ satisfies $K^*K = \Sigma_g$, so for $\sigma_k > 0$ its left singular functions are $u_k^\top g\rbr{x}/\sqrt{\sigma_k}$; hence when $\sigma_d > 0$ (\ie, $K$ has rank at least $d$), $\varphi\rbr{x}$ spans the top-$d$ left singular subspace of $K$.
\end{proof}

\subsection{Convergence of the Linear BYOL}\label{app:byol}

Denote the conditional expectation operator $\rbr{\Tcal f}\rbr{x} = \EE\sbr{f\rbr{x'}|x}$ on $L_2\rbr{\PP\rbr{x}}$, so that the fixed-point condition~\eqref{eq:eigen_fixpoint} reads $\Tcal\xi = \Lambda\xi$. Under the symmetric core set~\eqref{eq:sufficient_repr} and assuming compactness, $\Tcal$ is self-adjoint and positive semi-definite, with eigenvalues $1 = \lambda_1\ge\lambda_2\ge\cdots\ge 0$ and orthonormal eigenfunctions $\phi_1, \phi_2, \ldots$. Denote $\Phi_d = \sbr{\phi_1, \ldots, \phi_d}^\top$.

\begin{proposition}\label{prop:byol}
Consider the exact updates~\eqref{eq:byol_lambda} and~\eqref{eq:byol_xi} over $\xi\in L_2\rbr{\PP\rbr{x}; \RR^d}$ (a sufficiently rich function class). Assume $\lambda_d > \lambda_{d+1}$, and the initialization $\xi_0 = M_0\Phi_d + r_0\in L_2\rbr{\PP\rbr{x}; \RR^d}$ with $M_0\in\RR^{d\times d}$ invertible and $r_0\perp\phi_1, \ldots, \phi_d$. Then, for all $t\ge 0$, $\Lambda_{t+1}$ is invertible, $\text{span}\rbr{\xi_t} = \text{span}\rbr{\Tcal^t\xi_0}$, and $\text{span}\rbr{\xi_t}$ converges to the top-$d$ eigenspace $\text{span}\rbr{\Phi_d}$ at the rate $\Ocal\rbr{\rbr{\lambda_{d+1}/\lambda_d}^t}$. Moreover, any basis $M\Phi_d$ with invertible $M\in\RR^{d\times d}$ is a fixed point; the iterate $\xi_t$ itself converges in $L_2\rbr{\PP\rbr{x}}$ to $M_\infty\Phi_d$ for an invertible $M_\infty\in\RR^{d\times d}$, and $\Lambda_{t+1}$ converges to $M_\infty\diag\rbr{\lambda_1, \ldots, \lambda_d}M_\infty^{-1}$ at the rate $\Ocal\rbr{\rbr{\lambda_{d+1}/\lambda_d}^{2t}}$.
\end{proposition}
\begin{proof}
Denote $G_t = \EE\sbr{\xi_t\rbr{x}\xi_t\rbr{x}^\top}$ and $C_t = \EE_{\PP\rbr{x, x'}}\sbr{\xi_t\rbr{x'}\xi_t\rbr{x}^\top} = \EE\sbr{\rbr{\Tcal\xi_t}\rbr{x}\xi_t\rbr{x}^\top}$. The least squares~\eqref{eq:byol_lambda} gives $\Lambda_{t+1} = C_tG_t^{-1}$, and~\eqref{eq:byol_xi} is minimized pointwise by $\Lambda_{t+1}\xi_{t+1}\rbr{x} = \EE\sbr{\xi_t\rbr{x'}|x}$, \ie,
\begin{equation*}
    \xi_{t+1} = \Lambda_{t+1}^{-1}\Tcal\xi_t,
\end{equation*}
whenever $\Lambda_{t+1}$ is invertible. Since $\lambda_d > \lambda_{d+1}\ge 0$, $D = \diag\rbr{\lambda_1, \ldots, \lambda_d}\succ 0$ is invertible. By induction, suppose $\xi_t = M_t\Phi_d + r_t$ with $M_t$ invertible and $r_t\perp\Phi_d$. Then $G_t \succ 0$, and for any $a\neq 0$, $a^\top C_t a = \inner{a^\top\xi_t}{\Tcal a^\top\xi_t} \ge \lambda_d\nbr{M_t^\top a}_2^2 > 0$, so $C_t$ and $\Lambda_{t+1}$ are invertible. Moreover, $\xi_{t+1} = \Lambda_{t+1}^{-1}\rbr{M_tD\Phi_d + \Tcal r_t}$ gives $M_{t+1} = \Lambda_{t+1}^{-1}M_tD$ invertible and $r_{t+1} = \Lambda_{t+1}^{-1}\Tcal r_t\perp\Phi_d$. Since $\Lambda_{t+1}^{-1}$ only changes the basis, $\text{span}\rbr{\xi_{t+1}} = \text{span}\rbr{\Tcal\xi_t}$, \ie, the updates implement subspace iteration without orthogonalization, cf.~\citet[Section~8.2]{golub2013matrix}.

To establish the convergence of both $\xi_t$ and $\Lambda_{t+1}$ on $L_2\rbr{\PP\rbr{x}}$, set $\tilde r_t = M_t^{-1}r_t\perp\Phi_d$. Since $\Tcal$ acts componentwise, $\tilde r_{t+1} = M_{t+1}^{-1}r_{t+1} = D^{-1}M_t^{-1}\Tcal r_t = D^{-1}\Tcal\tilde r_t$, so $\tilde r_t = D^{-t}\Tcal^t\tilde r_0$ and $\nbr{\tilde r_t}_{L_2}\le\rbr{\lambda_{d+1}/\lambda_d}^t\nbr{\tilde r_0}_{L_2}$. Since $M_t^{-1}\xi_t = \Phi_d + \tilde r_t$ spans $\text{span}\rbr{\xi_t}$, the subspace distance to $\text{span}\rbr{\Phi_d}$ decays as $\Ocal\rbr{\rbr{\lambda_{d+1}/\lambda_d}^t}$. Furthermore, writing $E_t = \EE\sbr{\tilde r_t\tilde r_t^\top}\succeq 0$ and $F_t = \EE\sbr{\rbr{\Tcal\tilde r_t}\tilde r_t^\top}\succeq 0$, both of norm $\Ocal\rbr{\rbr{\lambda_{d+1}/\lambda_d}^{2t}}$, we have $G_t = M_t\rbr{\Ib + E_t}M_t^\top$, $C_t = M_t\rbr{D + F_t}M_t^\top$, and
\begin{equation*}
    M_{t+1} = G_t C_t^{-1}M_t D = M_t\rbr{\Ib + E_t}\rbr{D + F_t}^{-1}D = M_t\rbr{\Ib + \Delta_t}, \quad \Delta_t = \rbr{E_t - D^{-1}F_t}\rbr{D + F_t}^{-1}D.
\end{equation*}
Since $\nbr{\Delta_t}_2 = \Ocal\rbr{\rbr{\lambda_{d+1}/\lambda_d}^{2t}}$ is summable and each factor $\Ib + \Delta_t$ is invertible, the product $M_t = M_0\prod_{s=0}^{t-1}\rbr{\Ib + \Delta_s}$ converges to an invertible $M_\infty\in\RR^{d\times d}$. Hence $\xi_t = M_t\rbr{\Phi_d + \tilde r_t}\rightarrow M_\infty\Phi_d$ in $L_2\rbr{\PP\rbr{x}}$ and $\Lambda_{t+1} = M_t\rbr{D + F_t}\rbr{\Ib + E_t}^{-1}M_t^{-1}\rightarrow M_\infty DM_\infty^{-1}$ at the rate $\Ocal\rbr{\rbr{\lambda_{d+1}/\lambda_d}^{2t}}$. Finally, for $\xi = M\Phi_d$ with any invertible $M$, $G = MM^\top$, $C = MDM^\top$, $\Lambda = MDM^{-1}$, and $\Lambda^{-1}\Tcal\xi = \xi$, confirming every basis of $\text{span}\rbr{\Phi_d}$ is a fixed point.
\end{proof}

Therefore, although $\text{span}\rbr{\xi_t} = \text{span}\rbr{\Tcal^t\xi_0}$ for every finite $t$, in the unnormalized power iteration $\Tcal^t\xi_0 = M_0D^t\Phi_d + \Tcal^t r_0$ all components asymptotically align with $\phi_1$ when $\lambda_1 > \lambda_2$ (collapsing the limit basis to rank $1$); in contrast, the learned $\Lambda_{t+1}^{-1}$ rescales the eigen-directions so that $\xi_t\rightarrow M_\infty\Phi_d$ remains full-rank without requiring the explicit orthogonalization~\eqref{eq:orth_op}, consistent with the analysis of linear BYOL dynamics in~\citet{tian2021understanding}.

\subsection{JEPA Fixed Point as Functional SVD}\label{app:jepa_fixed_point}

We show that JEPA~\eqref{eq:jepa_obj} is the asymmetric nonlinear spectral parametrization~\eqref{eq:asym_nonlinear_repr_nn} with the condition $\rbr{x, m}$, that the fixed points~\eqref{eq:nonlinear_fixpoint} and~\eqref{eq:jepa_fixpoint} are satisfied by the functional singular value decomposition~(SVD), and that this decomposition is the minimizer of the corresponding objective with the normalization condition.

\paragraph{Alignment with the framework.} In I-JEPA and V-JEPA, the target locations $m$ are sampled independently of the input, \ie, $m\perp x'$, so that $\PP\rbr{x'|m} = \PP\rbr{x'}$. Moreover, the masking on the target-encoder output is a linear selection, $\sbr{\upsilon\rbr{x'}}_m = S_m\upsilon\rbr{x'}$, with $S_m$ selecting the entries at the locations $m$ and $S_mS_m^\top = \Ib$. Therefore, letting the predictor $g\rbr{x, m} = \texttt{nn}_\theta\rbr{\upsilon\rbr{x}, m}\in\RR^d$ output the representation at all locations, of which JEPA only uses $S_mg\rbr{x, m}$, the parametrization~\eqref{eq:asym_nonlinear_repr_nn} with the condition $\rbr{x, m}$ reads
\begin{equation}\label{eq:jepa_param}
    \PP\rbr{x'|x, m} = \PP\rbr{x'}\inner{\upsilon\rbr{x'}}{\texttt{nn}_\theta\rbr{\upsilon\rbr{x}, m}},
\end{equation}
with the normalization condition $\EE_{\PP\rbr{x'}}\sbr{\upsilon\rbr{x'}\upsilon\rbr{x'}^\top} = \Ib$, which implies $\EE_{\PP\rbr{x'}}\sbr{\sbr{\upsilon\rbr{x'}}_m\sbr{\upsilon\rbr{x'}}_m^\top} = \Ib$ for every $m$.

Denote the conditional expectation operator $\Tcal: L_2\rbr{\PP\rbr{x'}}\rightarrow L_2\rbr{\PP\rbr{x, m}}$,
\begin{equation}\label{eq:Tm_operator}
    \rbr{\Tcal h}\rbr{x, m} = \EE_{\PP\rbr{x'|x, m}}\sbr{h\rbr{x'}} = \int \frac{\PP\rbr{x'|x, m}}{\PP\rbr{x'}}h\rbr{x'}\PP\rbr{x'}dx'.
\end{equation}
Since the condition $\rbr{x, m}$ and the target $x'$ follow different distributions, $\Tcal$ maps between two different spaces and is not self-adjoint. Assuming $\Tcal$ is compact, it admits the SVD of the density ratio, \ie, the pointwise mutual information between $x'$ and $\rbr{x, m}$,
\begin{equation}\label{eq:jepa_svd}
    \frac{\PP\rbr{x'|x, m}}{\PP\rbr{x'}} = \sum_{i\ge 1}\sigma_i\psi_i\rbr{x, m}\phi_i\rbr{x'},
\end{equation}
with $1 = \sigma_1\ge\sigma_2\ge\cdots\ge 0$, and orthonormal $\cbr{\psi_i}$ in $L_2\rbr{\PP\rbr{x, m}}$ and $\cbr{\phi_i}$ in $L_2\rbr{\PP\rbr{x'}}$, so that $\Tcal\phi_i = \sigma_i\psi_i$. Denote $\Phi_d = \sbr{\phi_1, \ldots, \phi_d}^\top$, $\Psi_d = \sbr{\psi_1, \ldots, \psi_d}^\top$, and $\Sigma_d = \diag\rbr{\sigma_1, \ldots, \sigma_d}$.

\begin{proposition}[Fixed point]\label{prop:jepa_fixpoint}
Let $\upsilon = Q\Phi_d$ and $\texttt{nn}_\theta\rbr{\upsilon\rbr{x}, m} = Q\Sigma_d\Psi_d\rbr{x, m}$ for any orthogonal $Q\in\RR^{d\times d}$. Then $\upsilon$ satisfies the normalization condition, $\inner{\upsilon\rbr{x'}}{\texttt{nn}_\theta\rbr{\upsilon\rbr{x}, m}}$ is the rank-$d$ truncation of~\eqref{eq:jepa_svd}, and
\begin{equation*}
    \texttt{nn}_\theta\rbr{\upsilon\rbr{x}, m} = \EE_{\PP\rbr{x'|x, m}}\sbr{\upsilon\rbr{x'}},
\end{equation*}
whose entries at the locations $m$, \ie, multiplying $S_m$ on both sides, give the fixed point~\eqref{eq:jepa_fixpoint}.
\end{proposition}
\begin{proof}
By orthonormality, $\EE_{\PP\rbr{x'}}\sbr{\upsilon\upsilon^\top} = Q\EE\sbr{\Phi_d\Phi_d^\top}Q^\top = \Ib$, and $\inner{\upsilon\rbr{x'}}{\texttt{nn}_\theta\rbr{\upsilon\rbr{x}, m}} = \Phi_d\rbr{x'}^\top Q^\top Q\Sigma_d\Psi_d\rbr{x, m} = \sum_{i=1}^d\sigma_i\psi_i\rbr{x, m}\phi_i\rbr{x'}$. By $\Tcal\phi_i = \sigma_i\psi_i$, $\EE_{\PP\rbr{x'|x, m}}\sbr{\upsilon\rbr{x'}} = Q\Tcal\Phi_d = Q\Sigma_d\Psi_d$.
\end{proof}

\begin{proposition}[Variational characterization]\label{prop:jepa_svd}
Assume $\sigma_d > \sigma_{d+1}$. Consider the objective~\eqref{eq:jepa_obj} with the target $\upsilon_t = \upsilon$ on all locations, over $\upsilon$ and $g\rbr{x, m} = \texttt{nn}_\theta\rbr{\upsilon\rbr{x}, m}$ under the normalization condition,
\begin{equation}\label{eq:jepa_norm_obj}
    \min_{\upsilon, g}\,\,\EE_{\PP\rbr{x, x', m}}\sbr{\nbr{g\rbr{x, m} - \upsilon\rbr{x'}}_2^2}, \quad\text{s.t.}\quad \EE_{\PP\rbr{x'}}\sbr{\upsilon\upsilon^\top} = \Ib.
\end{equation}
Then the minimizers are exactly the pairs in Proposition~\ref{prop:jepa_fixpoint}, with the minimum $d - \sum_{i=1}^d\sigma_i^2$. Therefore, if the minimizers are realizable by the encoder $\upsilon$ and the predictor $\texttt{nn}_\theta$, the solution recovers the top-$d$ singular subspaces of~\eqref{eq:jepa_svd} and satisfies the fixed point~\eqref{eq:jepa_fixpoint}.
\end{proposition}
\begin{proof}
For fixed $\upsilon$, since $\EE\sbr{\rbr{\upsilon\rbr{x'} - \Tcal\upsilon\rbr{x, m}}^\top\rbr{\Tcal\upsilon\rbr{x, m} - g\rbr{x, m}}} = 0$ by the tower property,
\begin{equation*}
    \EE\sbr{\nbr{g - \upsilon}_2^2} = \EE\sbr{\nbr{g - \Tcal\upsilon}_2^2} + \EE\sbr{\nbr{\upsilon - \Tcal\upsilon}_2^2},
\end{equation*}
which is minimized by $g = \Tcal\upsilon$, \ie, the fixed point in Proposition~\ref{prop:jepa_fixpoint}. Plugging it back and applying the tower property again, $\EE\sbr{\nbr{\upsilon - \Tcal\upsilon}_2^2} = \EE\sbr{\nbr{\upsilon}_2^2} - \EE\sbr{\nbr{\Tcal\upsilon}_2^2}$, so~\eqref{eq:jepa_norm_obj} reduces to
\begin{equation*}
    d - \sum_{j=1}^d\inner{\upsilon_j}{\Tcal^*\Tcal\upsilon_j}_{L_2\rbr{\PP\rbr{x'}}}.
\end{equation*}
Since $\Tcal^*\Tcal$ is compact, self-adjoint, and positive semi-definite on $L_2\rbr{\PP\rbr{x'}}$ with eigenvalues $\sigma_i^2$ and eigenfunctions $\phi_i$, by Ky Fan's maximum principle~\citep{fan1949theorem}, for orthonormal $\upsilon_1, \ldots, \upsilon_d$, $\sum_{j=1}^d\inner{\upsilon_j}{\Tcal^*\Tcal\upsilon_j}\le\sum_{i=1}^d\sigma_i^2$, with equality if and only if $\text{span}\rbr{\upsilon} = \text{span}\rbr{\Phi_d}$ under $\sigma_d > \sigma_{d+1}$, \ie, $\upsilon = Q\Phi_d$ with orthogonal $Q$, and then $g = \Tcal\upsilon = Q\Sigma_d\Psi_d$.
\end{proof}

JEPA computes the loss only at the locations $m$, \ie, $\nbr{S_m\rbr{g - \upsilon}}_2^2$. For fixed $\upsilon$, its minimizer still satisfies $S_mg = S_m\Tcal\upsilon$, \ie, the fixed point~\eqref{eq:jepa_fixpoint}, while the reduced objective in $\upsilon$ becomes a location-weighted version of the above.

\paragraph{Symmetric case.} The SimSiam fixed point~\eqref{eq:nonlinear_fixpoint} is the special case without $m$ and with $x$ and $x'$ exchangeable, where $\Tcal$ is self-adjoint and positive semi-definite on $L_2\rbr{\PP\rbr{x}}$ under the symmetric core set~\eqref{eq:sufficient_repr}, so that~\eqref{eq:jepa_svd} reduces to the eigendecomposition with $\psi_i = \phi_i$ and $\sigma_i = \lambda_i$. Then Propositions~\ref{prop:jepa_fixpoint} and~\ref{prop:jepa_svd} give $\upsilon = Q\Phi_d$ and $\texttt{nn}_\theta\rbr{\upsilon\rbr{x}} = Q\Sigma_d\Phi_d\rbr{x} = \EE_{\PP\rbr{x'|x}}\sbr{\upsilon\rbr{x'}}$, \ie,~\eqref{eq:nonlinear_fixpoint}, where the predictor reduces to the linear map $\Lambda = Q\Sigma_dQ^\top$ on the representation, recovering the eigen fixed point~\eqref{eq:eigen_fixpoint}.

\end{appendices}

\end{document}